\documentclass[12pt]{article}

\usepackage{etoolbox}
\usepackage{authblk}
\usepackage[utf8]{inputenc}
\usepackage{lipsum}
\usepackage{authblk}
\usepackage[hyperfootnotes=false]{hyperref}
\usepackage{graphicx}
\graphicspath{{figures/}}
\usepackage{fullpage}
\usepackage{setspace}
\usepackage{algorithm}
\usepackage{algpseudocode}
\newcommand{\papertitle}{RObotic MAnipulation Network (ROMAN) -- \\ Hybrid Hierarchical Learning for Solving Complex Sequential Tasks}
\relax
\title{\papertitle}
\author[1]{\textbf{Eleftherios Triantafyllidis}}
\author[2]{\textbf{Fernando Acero}}
\author[1]{\textbf{Zhaocheng Liu}}
\author[1,2*]{\textbf{Zhibin Li}}
\affil[1]{School of Informatics, University of Edinburgh, UK}
\affil[2]{Department of Computer Science, University College London, UK}
\affil[*]{\small Corresponding author: alex.li@ucl.ac.uk}
\date{}

\usepackage[a4paper, total={6.5in, 9.5in}, footskip=0.5in]{geometry}
\usepackage{amsmath}
\usepackage{etoolbox}

\usepackage{float}
\usepackage{amssymb}
\usepackage{booktabs}
\usepackage{caption}
\usepackage{subcaption}
\usepackage{multirow}
\usepackage{tikz}
\usepackage{textcomp}
\usepackage{collcell}
\usepackage{color, xcolor, colortbl}
\usepackage{arydshln}
\usepackage{enumerate}
\definecolor{Gray}{gray}{0.9}
\definecolor{DarkGray}{gray}{0.8}
\definecolor{DarkerGray}{gray}{0.7}

\newcommand*{\opacity}{40}
\definecolor{high}{HTML}{32D732}
\definecolor{mid}{HTML}{FFBF00}
\definecolor{low}{HTML}{FF0000} 
\newcommand*{\minval}{0.00}
\newcommand*{\midval}{0.50}
\newcommand*{\maxval}{1.00}
\newcommand{\gradient}[1]{
    \ifdim #1 pt > \midval pt
            \pgfmathparse{int(round(100*(#1/(\maxval-\midval))-(\midval*(100/(\maxval-\midval)))))}
            \xdef\tempa{\pgfmathresult}
            \cellcolor{high!\tempa!mid!\opacity} #1
        \else
            \pgfmathparse{int(round(100*(#1/(\midval-\minval))-(\minval*(100/(\midval-\minval)))))}
            \xdef\tempa{\pgfmathresult}
            \cellcolor{mid!\tempa!low!\opacity} #1
        \fi
        }
\renewenvironment{abstract}
 {\small
  \begin{center}
  \bfseries \abstractname\vspace{-.5em}\vspace{0pt}
  \end{center}
  \list{}{
    \setlength{\leftmargin}{1.25cm}%
    \setlength{\rightmargin}{\leftmargin}%
  }%
  \item\relax}
 {\endlist}
 
\begin{document}
\maketitle
\begin{abstract}
\normalsize Solving long sequential tasks poses a significant challenge in embodied artificial intelligence. Enabling a robotic system to perform diverse sequential tasks with a broad range of manipulation skills is an active area of research. In this work, we present a Hybrid Hierarchical Learning framework, the Robotic Manipulation Network (ROMAN), to address the challenge of solving multiple complex tasks over long time horizons in robotic manipulation. ROMAN achieves task versatility and robust failure recovery by integrating behavioural cloning, imitation learning, and reinforcement learning. It consists of a central manipulation network that coordinates an ensemble of various neural networks, each specialising in distinct re-combinable sub-tasks to generate their correct in-sequence actions for solving complex long-horizon manipulation tasks. Experimental results show that by orchestrating and activating these specialised manipulation experts, ROMAN generates correct sequential activations for accomplishing long sequences of sophisticated manipulation tasks and achieving adaptive behaviours beyond demonstrations, while exhibiting robustness to various sensory noises. These results demonstrate the significance and versatility of ROMAN's dynamic adaptability featuring autonomous failure recovery capabilities, and highlight its potential for various autonomous manipulation tasks that demand adaptive motor skills.
\end{abstract}

\newpage
\section*{Introduction}
When we humans interact with our surrounding environment, we perform highly complex in-sequence tasks with seemingly minimal effort, especially when these are repeated in our everyday lives \cite{10.1145/3411763.3443442, Ashe2006-po, Ortenzi2019}. Even though these everyday tasks are so diverse in their nature, by virtue of our highly complex cognition, perception and unmatched motor dexterity among biological organisms, solving complex sequences of manipulation tasks appears anything than a difficult task \cite{9076603, Billardeaat8414}.

Changing this perspective to robots as agents with embodied intelligence, these interactions are currently far from trivial \cite{Tee2022, Billardeaat8414}. Solving complex sequential robotic manipulation tasks that are of long-horizon and further aggravated by not being interrelated remains an ongoing challenge \cite{davchev2021wish, 8843293}. A task as simple as retrieving a glass from a shelf, pouring in some water and placing it onto a specific part of a table may seem trivial to us, but from an embodied intelligence perspective this is still significantly challenging. Essentially, successful manipulation is achieved when both higher-level skills such as (i) grasping, lifting and moving objects are satisfied, (ii) sensory events are predicted and when it comes to sequential tasks, (iii) the higher-level end-goal is known and (iv) the sequences of different skills are conceptualised in our minds and more broadly by our nervous system \cite{Flanagan2006-ei, Ortenzi2019}.

Nevertheless, robots possess the ability to perform repetitive manipulation tasks with incredible amounts of high precision, provided these are confined to a specific task \cite{triantafyllidisrobot, 10.1145/3450626.3459830}. Some of these tasks include picking and placing \cite{8461249, 9076603}, swing-peg-in-hole \cite{8793789, 8793485}, catching in-flight objects \cite{7139529}, insertion \cite{8793485, schoettler2020deep} or solving something as complex as a Rubik's cube task \cite{doi:10.1177/0278364919887447}. However, when it comes to solving a sequence of multiple tasks that are independent in their nature and vary in complexity, significant challenges arise \cite{10.1145/3450626.3459830}.

\begin{figure}
    \begin{center}
    \centering
    \includegraphics[width=\textwidth]{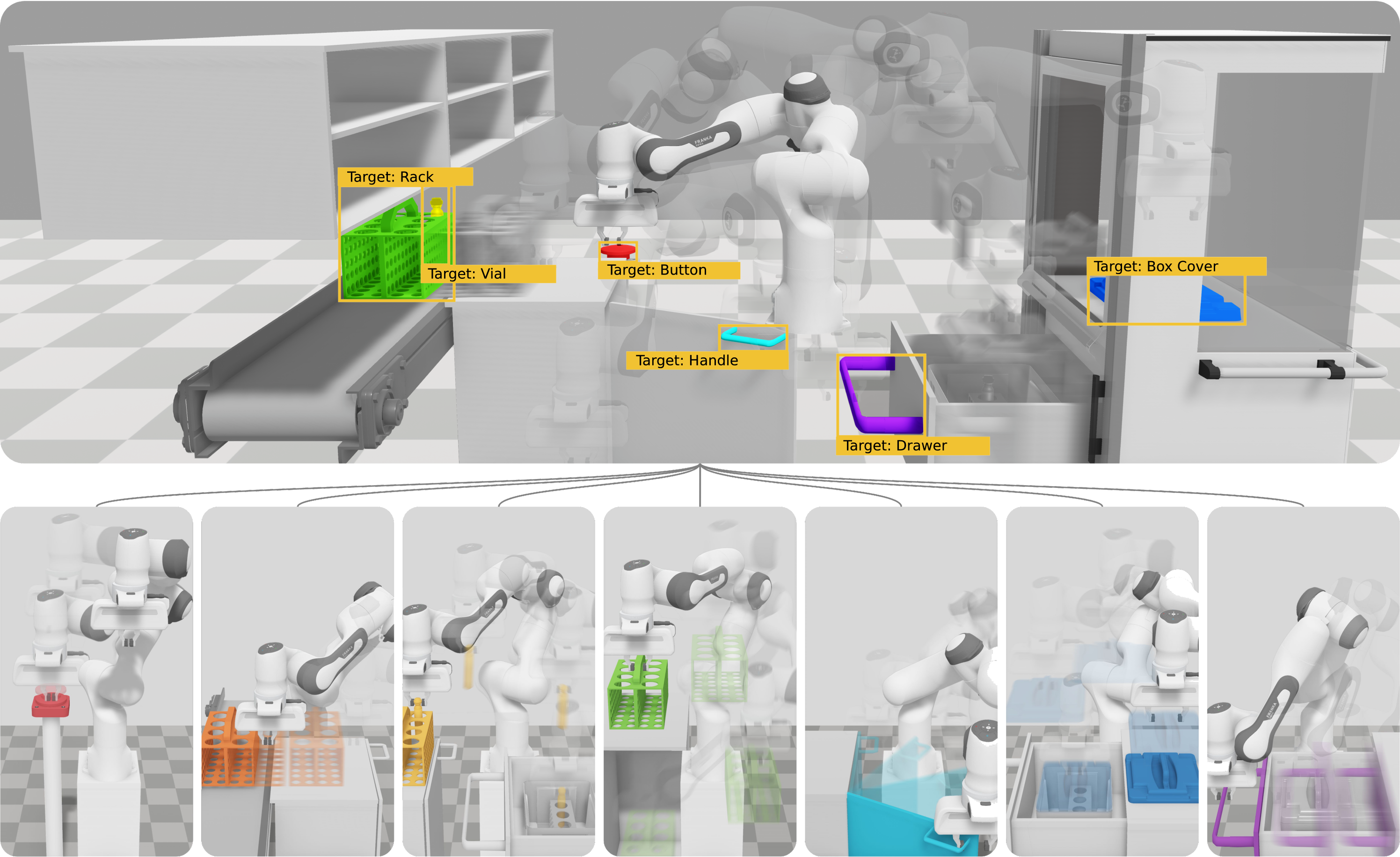}
    \caption{\small \textbf{The capabilities of the hierarchical architecture of the ROMAN framework:} A Hybrid Hierarchical Learning (HHL) framework for hierarchical task learning, with the capability of solving significantly long horizon sequential tasks that require the successful activation and coordination of diverse expert skills, commonly necessary in robotics and physics-based interactions. The derivation of high-level specialised experts in ROMAN, allowed the construction of a gating network, referred to as the Manipulation Network (MN), that is trained for elevated task-level scene understandings, the planning and execution of complex sequential long-time horizon tasks for the successful and timely activation of low-level expert networks. A set of seven in total specialised manipulation skills that are common in daily life were derived that can be recombined to create higher level types of manipulation skills. The specialised skills included in the ROMAN framework are: (i) \textit{Pushing a Button}, (ii) \textit{Pushing}, (iii) \textit{Picking \& Inserting}, (iv) \textit{Picking \& Placing}, (v) \textit{Rotating-Opening}, (vi) \textit{Picking \& Dropping} and (vii) \textit{Pulling-Opening}. Unlike conventional planning methods or state machines, ROMAN exhibits dynamic adaptability in (i) randomised task sequences, (ii) generalisation outside of demonstrated cases as well as (iii) recovery and robustness against local minima. The ability of the gating network (MN) to achieve such versatility and robustness is attributed to: (i) the HHL architecture in ROMAN's core framework as well as (ii) the task decomposition of complex sequences by the various experts in the framework in a high-level manner, allowing, in turn, the central gating network (MN) to be trained on high-level scene understanding and orchestrations of experts. The system architecture is based on the Mixture of Experts (MoE) that is able to successfully adapt to environmental demands, overcome various levels of uncertainties and most importantly learn with minimal human imitation complex sequential manipulation tasks.}
    \label{fig:CoverImage}
    \end{center}
\end{figure}

To overcome these limitations, we developed the novel RObotic MAnipulation Network (ROMAN) for hierarchical task learning. ROMAN is an event-based Hybrid-Hierarchical Learning (HHL) framework, visualised in \autoref{fig:CoverImage}. To the best of our knowledge, this is the first Mixture of Experts (MoE) based hierarchical approach that is capable of solving complex long-horizon manipulation tasks. We evaluated the framework in simulation and validated its robustness during long-horizon sequential tasks against sensory uncertainties. Thereafter, we performed extensive ablation studies of the internal learning procedure, evaluated the effects of different demonstrations and benchmarked the performance of ROMAN compared to monolithic neural networks. Our results demonstrate that by recombining and fusing ROMAN's core experts and skills together, our framework is able to solve significantly complex, long-horizon sequential manipulation tasks, commonly encountered in our everyday lives, with generalising capabilities. 

In the remainder of the paper, we review the related work, present the results of ROMAN from extensive validations in various test scenarios, discuss future work, and elaborate on the technical details of our methodology.

\paragraph{Real-world Impact of Intelligent Robotics} 
Programming a robotic system via analytical models to do a series of pre-programmed tasks can lead to sub-optimal solutions, as analytical models are typically simplifications of real-world dynamics, with a tendency to require expensive online computation, and are frequently unable to account for dynamically changing physical properties. Current advances in Artificial Intelligence (AI) and Machine Learning (ML) offer a promising avenue to advance robot learning and embodied intelligence \cite{8793485, 8461249, 10.1145/3197517.3201366, doi:10.1126/scirobotics.abb2174}.

The common RL algorithms among related work are the Proximal Policy Optimization (PPO) \cite{schulman2017proximal} and Soft Actor-Critic (SAC) \cite{pmlr-v80-haarnoja18b}. Although PPO is on-policy and generally less sample efficient than off-policy algorithms like SAC, PPO is less prone to instabilities and typically requires less hyperparameter tuning than SAC \cite{pmlr-v80-haarnoja18b, schulman2017proximal, 10.5555/3294996.3295141}. For these reasons, we chose PPO as our RL algorithm.

\paragraph{Imitation Learning and Learning from Demonstration}
RL algorithms face challenges in dealing with complex tasks, particularly when rewards are sparse, which exacerbate the exploration-exploitation trade-off \cite{10.1145/3170427.3186500, 10.5555/3454287.3455660, Zaadnoordijk2022}. A primary limitation of RL algorithms is that they do not typically possess any prior knowledge of the task, and in order to learn to solve a new task, they usually start from scratch by generating their own experience \cite{NIPS1996_68d13cf2, Zador2019}, which can range to millions of state transitions, often requiring days of training in simulation due to the absence of prior knowledge \cite{Thor2022, doi:10.1126/scirobotics.abb2174}. 

An alternative approach is the use of Imitation Learning (IL), inspired by the priory knowledge humans possess when learning motor tasks instead of starting from scratch \cite{Goldberg2019}, whereby expert demonstrations are provided with the purpose of enabling the robotic agent to learn to emulate the demonstrated behaviour. This is also known as Learning from Demonstration (LfD), which has shown promising results in complex and dexterous robotic tasks that would have been impossible to pre-program or significantly difficult to learn via conventional RL approaches due to the degree of exploration required as well as the necessity to carefully craft and specify rewards to achieve the desired behaviour \cite{8461249, NIPS1996_68d13cf2, 10.1145/3170427.3186500}.

Most IL and LfD approaches depend on demonstrations from human experts. While some form of demonstrations could instead be substituted via conventional trajectory optimisation \cite{8461249, NIPS2014_6766aa27} or RL approaches \cite{Mnih2015, pmlr-v37-schulman15, 10.5555/3045390.3045594}, these methods generally require carefully designed costs or rewards as well as significant interaction time between the robot and the environment. 

One of the main IL algorithms used in the related work is Behavioural Cloning (BC), which performs supervised learning on the policy from a set of demonstrated state-action transitions, showing promising success in robotic tasks \cite{8461249, 5152385, 4813899, 8843293}. While BC is a promising approach for warm-starting a policy and not having to rely on sampling from scratch such as conventional RL approaches, it has numerous limitations when used in isolation, such as lack of exploration, limited robustness towards new non-encountered states, and dependence on large, near-optimal demonstrations \cite{NIPS2016_cc7e2b87}.

Naively copying expert demonstrations via BC is prone to problematic performance as the agent visits states not encountered in the demonstrations due to covariate shifting errors that compound over time, which drives the need for large amounts of demonstration data \cite{NIPS2016_cc7e2b87, ross2011reduction}, which can also, in turn, lead to operator fatigue resulting in degraded performance over time \cite{9076603, 9492850}. Even from a biological perspective, it appears that the sole and naive dependence on an expert to learn new skills appears to be misguided \cite{Zaadnoordijk2022, Zador2019, Saxe2021}. While perhaps learning from scratch appears to be counter-intuitive, Zaadnoordijk et al. provided a very fitting analogy from a biological perspective whereby trial and error is a crucial part of our early lives: \textit{``Human infants are in many ways a close counterpart to a computational system learning in an unsupervised manner, as infants too must learn useful representations from unlabeled data"} \cite{Zaadnoordijk2022}. Switching back to a learning perspective, this could perhaps mean that learning in its core, should not entirely and solely be dependent upon imitating an ``expert", but rather allow some further exploration beyond ``naively copying others" and in this way still draw inspiration from a (neuro-) biological perspective \cite{Zador2019, Saxe2021}. 

An alternative to overcome some of the limitations of BC is Inverse Reinforcement Learning (IRL), where the reward function in the underlying observed demonstrations is inferred, as to best explain the demonstrations to a near-optimal behaviour \cite{NIPS2016_cc7e2b87, 10.1145/1015330.1015430, 10.5555/3045390.3045397}. One of the popular IRL algorithms is Generative Adversarial Imitation Learning (GAIL), which leverages Generative Adversarial Networks (GANs) \cite{NIPS2016_cc7e2b87}. In this framework, GAIL uses a second Neural Network (NN) known as a discriminator, responsible for distinguishing whether sets of trajectories were generated by the agent or the expert, and the less their divergence, the higher the reward in the imitation learning framework \cite{NIPS2016_cc7e2b87}.

\paragraph{Hierarchical Learning}
While combining deep RL and IL has the potential to endow robots with some of the capabilities exhibited by human cognition, there are still significant challenges to be solved when approaching large-scale problems, with the most notable one being long-horizon dexterous manipulation tasks \cite{9659344, 8843293}. Learning to solve very complex tasks using monolithic neural networks through RL or IL can be challenging due to: (i) long horizon problems, whereby the computational complexity of approximating a policy is very high, (ii) the variability of the task usually entails numerous sub-tasks, as well as (iii) the sample complexities that are associated with complex dexterous robotic tasks \cite{DBLP:conf/icml/0001JADYD18, Behabahani2018, 8843293, davchev2021wish}. Moreover, the successful completion of a long-time horizon task is contingent upon the successful completion of all sub-tasks and in a particular sequence \cite{9659344}. Lastly, even if such tasks are broken down into smaller sub-tasks to solve the problem\cite{Rajeswaran-RSS-18, 9659344, 9659344}, these can still significantly vary in their nature, further aggravating learning due to limited task inter-relation \cite{1707.03374}.

Hierarchical Learning (HL), whether used for RL or IL is a promising mitigation strategy for the above problems and can alleviate some of these complexities \cite{frans2018meta, merel2018hierarchical, Merel2019, doi:10.1126/scirobotics.abb2174}. HL exhibits numerous benefits when it comes to multitasking or generally complex tasks associated with sparse rewards \cite{davchev2021wish}, as it allows the decomposition of tasks, commonly referred to as ``skills" \cite{8843293}. When these hierarchical control policies furthermore implement IL, commonly referred to as HIL, the decomposition of those sub-tasks allows for significantly easier differentiation between the specialised experts and the acquisition of specialised skills by a human in a teacher-student fashion \cite{DBLP:conf/icml/0001JADYD18, 8843293, fox2018parametrized}.

A popular approach is the use of MoEs, whereby multiple task-specific experts are trained and specialised on a given sub-task and managed by a gating network, with successful applications in computer graphics and animation \cite{10.1145/3197517.3201366, 10.5555/3454287.3454618}, as well as robotics \cite{doi:10.1177/0278364912472380, 8843293, doi:10.1126/scirobotics.abb2174}. While HRL is a step closer towards solving complex robotic tasks, it still fundamentally depends on the RL paradigm and hence is adversely affected by sparse rewards, complex planning tasks or the difficulty to use prior knowledge to solve tasks \cite{8843293, 9659344}. Alternatively, HIL \cite{8843293, DBLP:conf/icml/0001JADYD18} and unlike RL or HRL, leverages expert demonstrations, aiding the overall training process as it allows the demonstrator to isolate certain sub-tasks to facilitate solving longer, more complex and in-sequence tasks, as opposed to RL or HRL \cite{fox2018parametrized, 8843293}. 

Currently, in robotic manipulation, methods using MoEs trained with HRL or HIL are limited in the state-of-the-art \cite{Rajeswaran-RSS-18, 9659344}. Based on prior work that introduced ensemble techniques in robot locomotion \cite{doi:10.1126/scirobotics.abb2174} and human-centred teleoperation \cite{9492850}, we are motivated to explore a new approach of IL using human-demonstrated tasks developing a suitable MoE architecture in the domain of robotic manipulation. This approach has the potential to extend beyond the original demonstrations and enable more complex manipulation tasks. While their results were validated against BC, showing higher (90\%+) success rates compared to RL, the tasks studied remained fairly simplistic, assuming non-sequential tasks with short time horizons using a lower DoF manipulator. This effectively limited their evaluation to three experts solving only \textit{Picking and Placing }tasks \cite{9659344}. In contrast, our work can train a single expert capable of solving \textit{Picking and Placing} and when combined with other experts specialised in rather high-level sub-tasks compared to \cite{9659344}, complex and long-horizon sequential tasks commonly seen in robotic manipulation can be solved.

\subsection*{Results}
\label{sec:Results}
This section presents the results of the ROMAN framework. ROMAN is composed of a modular hybrid hierarchical architecture to combine adaptive motor skills for solving complex long sequential manipulation tasks. It features a central gating network referred to as the Manipulation Network (MN), that activates specialised task-level experts in a required sequential combination, resulting in higher levels of manipulation capabilities and improved generalisation to non-demonstrated situations. Moreover, the MN exhibits recovery capabilities by activating multiple expert weights to overcome local minima, which ultimately enhances the robustness for solving long horizon sequential tasks.

From the experiments, our validation shows the robustness of ROMAN's HHL approach against (i) large exteroceptive observational noise, (ii) complex non-interrelated compositional sub-tasks, (iii) long time-horizon sequential tasks, and (iv) cases not encountered during the demonstrated sequences. ROMAN achieves behaviour beyond imitation through the hybrid training procedure which allows, in turn, the dynamic coordination of experts to recover from local minima successfully. These findings highlight the versatility and adaptability of ROMAN, enabling autonomous manipulation with adaptive motor skills.

Foremost, the scalability of the hierarchical learning versus a monolithic neural network approach was evaluated. Consequently, we initially compared ROMAN's preliminary 2D and final 3D hierarchical architecture stage against monolithic neural networks sharing an equivalent hybrid learning procedure. Thereafter, we evaluate ROMAN's final 3D stage composed of seven experts against (i) different levels of exteroceptive uncertainty, (ii) perform extensive ablation studies of the internal hybrid learning procedure, and evaluate (iii) the effects of a different number of demonstrations provided to the framework. All subsequent results from the experiments were conducted with identical network settings (states $s$, actions $a$ and rewards $r$), number of demonstrations and hyperparameter settings to retain consistency and conduct a fair comparison. The architecture of ROMAN is visually depicted in \autoref{figure:NN_MN_StructureOverview}. The state space and settings of each NN incorporated in ROMAN is specified in \autoref{table:DetailedNetworkCharacteristicsStructure}. More details regarding the hyperparameters and dimensions of the networks can be found in the supplementary materials and more specifically in the Supplementary Tables 12, 13, 14 and 15. More information regarding the demonstrations, their characteristics and their acquisition can be found in the Methods section.

\paragraph{\textbf{Definition of Success Rate: }} \label{par:SuccessRate} To properly validate the system, we first define the term success rate. In the medical laboratory environment setting that we chose to validate ROMAN, success is attained when all seven sub-task goals depicted in \autoref{fig:CoverImage} were satisfied. Consequently, to consider a scenario successful, all inter-related sub-tasks needed to be sequentially completed within a time limit.

\subsubsection*{\textbf{Hierarchical Learning and the Limitations of Monolithic Networks in Long Horizon Tasks}}
In ROMAN's preliminary stage, a total of five experts were derived, operating in planar i.e. 2D coordinates, visually shown in \autoref{figure:TSNE-Depiction-ROMAN-Snapshots}.c. Thereafter the ROMAN framework was scaled up to 3D space consisting of a total of seven experts, shown in \autoref{fig:CoverImage} and \autoref{figure:TSNE-Depiction-ROMAN-Snapshots}.d. As such, in this section we compare ROMAN's preliminary and final stage against two monolithic single NNs, all sharing an equivalent employed hybrid learning procedure (shown in detail in \autoref{figure:NN_MN_StructureOverview}.a) for the 2D and 3D case respectively. These baseline evaluations allowed for a direct comparison of a monolithic versus a hierarchical approach, to infer, evaluate and demonstrate the advantages of a hierarchical task decomposition sharing an identical learning procedure. The single NNs shared an identical state space to ROMAN's MN and correspondingly identical actions to ROMAN's experts. Moreover, to conduct and allow for fair comparison, a total of $N=100$ and $N=140$ demonstrations were provided to the single NNs, accounting for ROMAN's 2D and 3D cases composed of five and seven experts pre-trained each with $N=20$ demonstrations respectively.

The results are shown in \autoref{table:2DCase_SingleNN_VS_RMN} and \autoref{table:SingleNN_VS_RMN}, for the 2D and 3D case of the monolithic NN respectively. These results suggest that a single NN is unable to solve and converge to a stable policy that entails the complex nature and long sequential task of the validated manipulation scenario. Even though the same training procedure is employed, it can be inferred that monolithic solutions do not contribute to stability nor high success rates when faced with long horizon and complex sequential tasks and the value of a hierarchical architecture can be underlined. More specifically, while in 2D the single NN attains to some extent high success rates these remain significantly lower than ROMAN's hierarchical architecture especially in increasing time horizons (S3, S4 and S5). Extending the dimensionality to 3D space reveals that a monolithic NN is mostly unable to attain robust performance (S3), exhibiting complete failure in longer and more complex sequential cases (as seen in S4 and beyond). These results highlight the value of a hierarchical task decomposition as with ROMAN's hierarchical architecture. For more expansion and technical details regarding the monolithic NNs, including their architecture and hyperparameters, please consult Supplementary Tables 14 and 15.

\subsubsection*{\textbf{Evaluation Against Exteroceptive Uncertainty}} 
\label{par:Eval_Uncertainty}
All subsequent results from this section onward will present ROMAN's final stage composed of seven in total experts operating in 3D space to study the domain of robotics with complex settings. More details regarding ROMAN's hierarchical architecture and specific network settings can be seen in \autoref{table:DetailedNetworkCharacteristicsStructure}. While scaling up to 3D with 7 experts, the first objective was to evaluate the robustness of the hierarchical framework against different levels of Gaussian distributed exteroceptive noise on the position states. The rationale for introducing noise in the exteroceptive states was to thoroughly evaluate the robustness of the framework against uncertainties under realistic conditions, since such states are typically more prone to noise than proprioceptive states in robotic systems in real-life scenarios \cite{doi:10.1126/scirobotics.abb2174}.

\paragraph{Expert Networks -- Evaluation against Increasing Levels of Gaussian Noise:} First and foremost, the individual robustness of each expert was evaluated as it was deemed critical before proceeding to the evaluation of the gating network, i.e. the MN's performance during the sequential activation of those experts. This, in turn, allowed us to understand and infer whether failures were being caused by the individual expert performance or by the subsequent sequential activation of those by the MN. This further minimised the covariance between the success rates of each expert and that of the MN. From the results shown in \autoref{table:SuccessRatesNoiseLevels} were each individual expert performance is evaluated, it can be inferred that even when presented with higher levels of noise, all expert NNs are resilient against the tested levels of uncertainty. It is worth noting that all \textit{Picking} experts, were slightly more prone to errors due to their higher complexity, in line with \cite{9659344, 9811819}.

\paragraph{Manipulation Network -- Evaluation against Increasing Levels of Gaussian Noise:} For the next evaluation, the performance of the MN was tested and its ability to coordinate the different experts incorporated in the hierarchical architecture of ROMAN. From the given seven experts, we tested seven different randomised case scenarios, where each scenario requires adding an additional expert, granting the overall tasks more complex and of longer horizons. From the results shown in \autoref{table:SuccessRatesNoiseLevels} where the MN's performance is tested, it can be inferred that the MN exhibits robust performance to different noise levels. As it can be inferred, despite adding more experts naturally increasing the dimensionality of the problem, the results show that the MN is sufficiently resilient and capable of coordinating the different experts in the hierarchical architecture even in the most complex settings in scenarios 6 and 7. Nevertheless, a performance drop in scenarios 3,4 and 5, compared to 6 and 7 can be observed, which is discussed in detail further along the Results section.

\paragraph{Evaluation of Vision System:}
The next objective was to test the robustness of ROMAN against exteroceptive uncertainties from a simulated vision system in the simulation. The reasoning of extending ROMAN to a vision system was to approximate a closer to real life case and illustrate the capabilities and versatility of the framework to operate in a more realistic setting. ROMAN and its experts including the MN, were not directly trained with this vision detection module, but rather directly evaluated on it to test the versatility and robustness of the framework. More details regarding the vision system its characteristics and technical details can be found in the Methods section.

The results from the vision detection module are shown in \autoref{table:SuccessRatesNoiseLevels}. From the results, it is shown that using a pre-trained object detection module from vision attains high success rates even amongst the most complex sequential tasks with the longest horizons. While a slight decrease in success rates as more sequences are added is observed, ROMAN nonetheless exhibits robustness to the vision system. The decrease in success rates in S6 and less in S7 can be attributed to the unboxing sub-task, which is more prone to visual occlusion (see \autoref{fig:CoverImage}) and the similarity in the exteroceptive observations later analysed in a t-distributed stochastic neighbor embedding (t-SNE), (see \autoref{figure:TSNE-Depiction-ROMAN-Snapshots}).

\begin{table}[H]
\caption{\textbf{Summary of the architectural settings of ROMAN and the results of each specialised expert and the central manipulation network across different levels of uncertainty. \autoref{table:DetailedNetworkCharacteristicsStructure} summarises ROMAN's overall neural network architecture and characteristics including the state space of each neural network, and the settings of individual components in the hierarchical framework. \autoref{table:SuccessRatesNoiseLevels} summarises the results evaluated on increasing levels of Gaussian noise in the exteroceptive states and uncertainties stemming from the vision system for each expert and the main Manipulation Network (MN) in ROMAN.}}
\label{table:ROMAN_Architecture_Results_Experts_NN_Noise}
    \begin{subtable}[h]{1.0\textwidth}
        \centering 
        \begin{footnotesize}
        \setlength\tabcolsep{1pt}
        \resizebox{\textwidth}{!}{\begin{tabular}{lccccccc} \bottomrule
        \rowcolor{Gray} \multicolumn{8}{c}{{\textbf{Network Architecture, Characteristics and Demonstration Settings}}} \\ \hline
        
        \multirow{2}{*}{\textbf{Master (MN)}} & \multicolumn{7}{c}{{\textbf{Experts NNs}}} \\ \cmidrule(lr){2-8}
        &{Push-Button} & {Push} & {Pick \& Insert} & {Pick \& Place} & {Rotate Open} & {Pick \& Drop} & {Pull Open} \\ \hline
        \rowcolor{Gray} \multicolumn{8}{c}{{\textbf{State Space (Vector Size)}}} \\ \hline
        \scriptsize \textbf{{Total: 29}} & \scriptsize \textbf{{Total: 11}} & \scriptsize \textbf{{Total: 14}} & \scriptsize \textbf{{Total: 14}}  & \scriptsize \textbf{{Total: 14}}  & \scriptsize \textbf{{Total: 11}}  & \scriptsize \textbf{{Total: 14}}  & \scriptsize \textbf{{Total: 11}}  \\ \hdashline
        
        \scriptsize Agent Position (3) & \scriptsize Agent Position (3) & \scriptsize Agent Position (3) & \scriptsize Agent Position (3) & \scriptsize Agent Position (3) & \scriptsize Agent Position (3) & \scriptsize Agent Position (3) & \scriptsize Agent Position (3) \\
        \scriptsize Agent Velocity (3) & \scriptsize Agent Velocity (3) & \scriptsize Agent Velocity (3) & \scriptsize Agent Velocity (3) & \scriptsize Agent Velocity (3) & \scriptsize Agent Velocity (3) & \scriptsize Agent Velocity (3) & \scriptsize Agent Velocity (3) \\
        \scriptsize Gripper Force (2) & \scriptsize Gripper Force (2) & \scriptsize Gripper Force (2) & \scriptsize Gripper Force (2) & \scriptsize Gripper Force (2) & \scriptsize Gripper Force (2) & \scriptsize Gripper Force (2) & \scriptsize Gripper Force (2) \\
        \scriptsize Full Environment (21) & \scriptsize Button Position (3) & \scriptsize Rack Position (3) & \scriptsize Rack Position (3) & \scriptsize Rack Position (3) & \scriptsize Cabinet Position (3) & \scriptsize Box Position (3) & \scriptsize Drawer Position (3) \\
        & & \scriptsize Conveyor Position (3) & \scriptsize Vial Position (3) & \scriptsize Rack Target (3) & & \scriptsize Unbox Target (3) &\\
        
        \hline \rowcolor{Gray} \multicolumn{8}{c}{{\textbf{Action Space (Vector Size)}}} \\ \hline
        \scriptsize \textbf{{Total: 7}} & \scriptsize \textbf{{Total: 4}} & \scriptsize \textbf{{Total: 4}} & \scriptsize \textbf{{Total: 4}}  & \scriptsize \textbf{{Total: 4}}  & \scriptsize \textbf{{Total: 4}}  & \scriptsize \textbf{{Total: 4}}  & \scriptsize \textbf{{Total: 4}}  \\ \hdashline
        \scriptsize Agent Weights (7) & \scriptsize Agent Velocity (3) & \scriptsize Agent Velocity (3) & \scriptsize Agent Velocity (3) & \scriptsize Agent Velocity (3) & \scriptsize Agent Velocity (3) & \scriptsize Agent Velocity (3) & \scriptsize Agent Velocity (3) \\
        & \scriptsize Gripper State (1) & \scriptsize Gripper State (1) & \scriptsize Gripper State (1) & \scriptsize Gripper State (1) & \scriptsize Gripper State (1) & \scriptsize Gripper State (1) & \scriptsize Gripper State (1) \\
        
        \hline \rowcolor{Gray} \multicolumn{8}{c}{{\textbf{Demonstration Settings and Training Times (Number, Demo Time, Train Time)}}} \\ \hline
        \scriptsize N = 42 (N=6 per Case)& \scriptsize N = 20 & \scriptsize N = 20 & \scriptsize N = 20 & \scriptsize N = 20 & \scriptsize N = 20 & \scriptsize N = 20 & \scriptsize N = 20 \\
        \scriptsize $\scriptscriptstyle t_{demo} \approx$ 42min & \scriptsize $\scriptscriptstyle t_{demo} \approx$ 7min & \scriptsize $\scriptscriptstyle t_{demo} \approx$ 6min & \scriptsize $\scriptscriptstyle t_{demo} \approx$ 12min &\scriptsize $\scriptscriptstyle t_{demo} \approx$ 10min & \scriptsize $\scriptscriptstyle t_{demo} \approx$ 7min & \scriptsize $\scriptscriptstyle t_{demo} \approx$ 9min & \scriptsize $\scriptscriptstyle t_{demo} \approx$ 7min \\
        \scriptsize $\scriptscriptstyle t_{train}$ = 11h 22min & \scriptsize $\scriptscriptstyle t_{train}$ = 3h 1min & \scriptsize $\scriptscriptstyle t_{train}$ = 3h 59min & \scriptsize $\scriptscriptstyle t_{train}$ = 23h 30min & \scriptsize $\scriptscriptstyle t_{train}$ = 11h 46min & \scriptsize $\scriptscriptstyle t_{train}$ = 2h 39min & \scriptsize $\scriptscriptstyle t_{train}$ = 3h 43min & \scriptsize $\scriptscriptstyle t_{train}$ = 3h 18min \\
        \bottomrule
        \end{tabular}}
        \end{footnotesize}
        \caption{Overview of the state and action space, including the demonstrations provided for each NN in ROMAN. A total of $N=20$ demonstrations were provided to pre-train the expert NNs in ROMAN, and a total of $N=42$ demonstrations to the MN, corresponding to $N=6$ for each of the seven sequential cases.}
        \label{table:DetailedNetworkCharacteristicsStructure}
	\end{subtable}
	\vspace{0.5cm}
	
	\begin{subtable}[h]{1.0\textwidth}
        \begin{footnotesize}
        \setlength\tabcolsep{2.5pt}
        \resizebox{\textwidth}{!}{\begin{tabular}{l|rrrrrrr} \hline
        \rowcolor{Gray} \multicolumn{8}{c}{{\textbf{Individual Expert Success Rates [10,000 Trials per Cell]}}} \\ \hline
        \textbf{Success ($\%$)} & 
         \textbf{Push-Button} & \textbf{Push} & \textbf{Pick \& Insert} & \textbf{Pick \& Place} & \textbf{Rotate Open} & \textbf{Pick \& Drop} & \textbf{Pull Open} \\  \hline
        $\sigma = \pm 0.0 \text{ [cm]}$ & {0.996} & {0.998} & {0.919} & {0.986} &  {0.999} &  {0.997} &  {0.970}\\
        $\sigma = \pm 0.5 \text{ [cm]}$ & {0.999} & {0.999} & {0.933} & {0.989} &  {0.999} &  {0.994} &  {0.993}\\
        $\sigma = \pm 1.0 \text{ [cm]}$ & {0.999} & {0.999} & {0.939} & {0.982} & {0.999} &  {0.994} &  {0.985}\\
        $\sigma = \pm 1.5 \text{ [cm]}$ & {1.000} & {0.999} & {0.920} & {0.965} & {0.999} &  {0.988} &  {0.969}\\
        $\sigma = \pm 2.0 \text{ [cm]}$ & {0.999} & {0.998} & {0.872} & {0.941} & {0.999} &  {0.973} &  {0.962}\\
        $\sigma = \pm 2.5 \text{ [cm]}$ & {0.999} & {0.991} & {0.826} & {0.903} & {0.998} &  {0.955} &  {0.950}\\ \hline
        
        \rowcolor{Gray} \multicolumn{8}{c}{{\textbf{Manipulation Network Success Rates [1,000 Trials per Cell]}}} \\ \hline
         \multirow{3}{*}{\textbf{Success ($\%$)}} & 
         \textbf{Scenario 1} & \textbf{Scenario 2} & \textbf{Scenario 3} & \textbf{Scenario 4} & \textbf{Scenario 5} & \textbf{Scenario 6} & \textbf{Scenario 7} \\ 
          \cmidrule(lr){2-2} \cmidrule(lr){3-3} \cmidrule(lr){4-4} \cmidrule(lr){5-5} \cmidrule(lr){6-6} \cmidrule(lr){7-7} \cmidrule(lr){8-8}
         & {One Expert} & {Two Experts} & {Three Experts} & {Four Experts} & {Five Experts} & {Six Experts} & {Seven Experts} \\
         & {\footnotesize {(Push-Button)}} & {\footnotesize {(+ Push)}} & {\footnotesize {(+ Pick \& Insert)}} & {\footnotesize {(+ Pick \& Place)}} & {\footnotesize {(+ Rotate Open)}} & {\footnotesize {(+ Pick \& Drop)}} & {\footnotesize {(+ Pull-Open)}} \normalsize \\ \hline
        $\sigma = \pm 0.0 \text{ [cm]}$ & {0.976} & {0.972} & {0.847} & {0.951} &  {0.728} &  {0.954} &  {0.903}\\
        $\sigma = \pm 0.5 \text{ [cm]}$ & {0.973} & {0.975} & {0.817} & {0.959} &  {0.794} &  {0.960} &  {0.952}\\
        $\sigma = \pm 1.0 \text{ [cm]}$ & {0.977} & {0.990} & {0.798} & {0.946} & {0.776} &  {0.933} &  {0.939}\\
        $\sigma = \pm 1.5 \text{ [cm]}$ & {0.980} & {0.986} & {0.720} & {0.846} & {0.722} &  {0.836} &  {0.841}\\
        $\sigma = \pm 2.0 \text{ [cm]}$ & {0.967} & {0.986} & {0.737} & {0.837} & {0.753} &  {0.820} &  {0.815}\\
        $\sigma = \pm 2.5 \text{ [cm]}$ & {0.973} & {0.986} & {0.723} & {0.763} & {0.697} &  {0.719} &  {0.744}\\
        \hline       
        
        \rowcolor{Gray} \multicolumn{8}{c}{{\textbf{Manipulation Network Vision System Success Rates [100 Trials per Cell]}}} \\ \hline
        \textbf{Success (\%)} & \textbf{Scenario 1} & \textbf{Scenario 2} & \textbf{Scenario 3} & \textbf{Scenario 4} & \textbf{Scenario 5} & \textbf{Scenario 6} & \textbf{Scenario 7} \\ \hline
        Visual & {0.97} & {0.96} & {0.82} & {0.83} & {0.79} &  {0.51} &  {0.72}\\
        \hline
        \end{tabular}}
        \end{footnotesize}
        \caption{The success rates for all individual experts including the manipulation network in ROMAN for the 3D setting, across all scenarios, based on increased levels of Gaussian noise in the exteroceptive position observations. Moreover, the feasibility and robustness of those trained models is tested by evaluating their performance directly on a vision system that provides exteroceptive information.}
        \label{table:SuccessRatesNoiseLevels}
	\end{subtable}
\end{table}

\subsubsection*{\textbf{Ablation Study on ROMAN's Default Learning Approach}} 
\label{par:Eval_AlgorithmAblationStudy}
The next validation entails a thorough comparison with state-of-the-art learning paradigms, including HRL and HIL approaches, similar to related work \cite{9659344, 8461249}. ROMAN makes use of BC to warm-start the policy via supervised learning and thereafter uses intrinsic $r_{I}$ (IL: GAIL) and extrinsic $r_{E}$ (RL) rewards via PPO for training. In this setting, we conduct ablations to the training procedure by excluding at least one of the previous paradigms. More details on the hybrid hierarchical learning architecture of ROMAN can be found in the Methods section.

From the ablation results in \autoref{table:ROMANAlgorithmComparison} it can be inferred that the most apparent result is that the exclusive use of extrinsic rewards $r_{E}$ (RL) exhibited complete failure. This result highlights and suggests the high complexity of the tasks studied, which are effectively unattainable via random exploration of the action space. Using the intrinsic rewards $r_{I}$ provided by GAIL or coupling it with extrinsic $r_{E}$ for RL and GAIL, rendered both significantly higher success rates, however, limited to S1 to S3, with long horizon tasks such as S4 to S7 still being unattainable.

From the related work \cite{davchev2021wish, 9659344, 8461249}, we summarise that training with BC alone appears to yield rapid performance degradation as the task time horizon is increased and the overall complexity of the sequential task is rendered higher. This is in line with our results for both BC and BC with extrinsic $r_{E}$ rewards (RL, BC) at $\sigma = \pm 0.5 \text{cm}$ noise. While a significant boost in success rates is observed with both methods compared to previous, longer sequential tasks such as S4 to S7 (which exhibit larger variance in the trajectories visited due to compounding of errors throughout the trajectory) show lower performance when compared to that of ROMAN's default learning approach. In line with previous work, we find that while BC may be a simple yet effective algorithm in some settings, its performance is greatly affected when presented with out-of-distribution states, and can lead to significant distribution drifting \cite{NIPS2016_cc7e2b87, Rajeswaran-RSS-18, pmlr-v15-ross11a}. To further test this finding, we moreover evaluate both BC and RL, BC on increased levels of noise of $\sigma = \pm 1.0 \text{cm}$ and $\sigma = \pm 2.0 \text{cm}$. 

Increasing the level of noise to $\sigma = \pm 1.0 \text{cm}$, it is observed the success rates for both BC and RL, BC drop slightly. ROMAN's default settings still attain the highest success rates amongst the studied learning paradigms. Increasing the level of uncertainty to $\sigma = \pm 2.0 \text{cm}$, we observe a significant drop in success rates for both BC and RL, BC compared to lower levels of noise. It is apparent that employing BC at such levels of uncertainty further highlights its limitation. Adding a $r_{E}$ for RL, BC exhibits slightly higher success rates but not significantly higher degree. In comparison, ROMAN's success rates while dropping slightly compared to previous levels of noise, still retain significantly higher degrees of resilience, highlighting the value of avoiding ``naively" imitating demonstrations as with the BC-employed method.

Hence, we can conclude that the proposed HHL approach within ROMAN is highly advantageous in overcoming increasing exteroceptive uncertainties and the complexities associated with longer time-horizon sequential tasks. This is attributed to the combination of (i) using BC up to a given epoch for warm-starting policy optimization, (ii) thereafter using the intrinsic reward provided by GAIL to further minimize the divergence of the agent and that of the expert demonstrator, and finally (iii) the addition of an extrinsic reward from the RL paradigm to allow the agent to explore further and beyond what was demonstrated upon. Most importantly, further in the Results section, we show that the HHL architecture of ROMAN exhibits dynamic adaptation in the presence of cases not encountered in the demonstrated sequence, and extends beyond the imitated behaviour during training. This is attributed to ROMAN's balance between exploitation and exploration.

\subsubsection*{\textbf{Effects of Demonstrations}}
\label{par:Eval_DemonstrationEffect}
For the final evaluation, we compared the effect of a different number of demonstrations on the overall performance of ROMAN and more specifically on its gating network. In particular, we analysed the effects of $N=7$, $N=21$ and $N=42$ demonstrations on the success rates across all scenarios for the MN. The results are shown in \autoref{table:DemonstrationComparison}. From the results, it is observed that a relatively small number of demonstrations for the MN ($N=21$, which corresponds to only $N=3$ demonstrations for each of the seven sub-tasks), is sufficient to have a reasonable success rate. Doubling the number of demonstrations to $N=42$, yields higher success rates than $N=21$, but not to a significantly higher difference. Finally, a one-shot demonstration of each scenario (i.e. $N=7$), did not yield sufficiently acceptable success rates during more complex sequences as shown in S4 to S7. More details regarding the demonstrations and expansion on the results can be found in Supplementary Table 7. 

\begin{table}[H]
\caption{\textbf{Experimental results of ROMAN validated across different evaluation and cases. Results stem from 1,000 trials for each individual cell listed in the tables below. Uncertainty levels are in the form of Gaussian noise and indicated in the leftmost column of each table. An identical number of demonstrations, network settings and hyperparameters were used.}}
\label{table:Experimental_Resuls_ROMAN_Full}
    \vspace{-0.25cm}
    \begin{subtable}[h]{1.0\textwidth}
    		\centering
    		\begin{footnotesize}
    		\setlength\tabcolsep{12pt}
    		\begin{tabular}{l|rrrrr} \bottomrule
            \rowcolor{Gray} \multicolumn{6}{c}{{\textbf{[2-D] Preliminary Version of Single NN versus ROMAN on Case Scenarios}}} \\ \hline
             \cellcolor{gray!20} & \textbf{S1} & \textbf{S2} & \textbf{S3} & \textbf{S4} & \textbf{S5} \\
            \multirow{-2}{*}{ \cellcolor{gray!20} {$\sigma = \pm 0.5 \text{cm}$}} & {\footnotesize {Push}} & {\footnotesize {+Lift} }& {\footnotesize {+Pick \& Insert}} & {\footnotesize {+Pick \& Drop}} & {\footnotesize {+Pull}} \\ \hline
             
            Single NN & {0.997} & {0.841} & {0.699} & {0.591} &  {0.565} \\
            ROMAN &  {0.993} & {0.995} & {0.982} & {0.971} &  {0.974} \\
            \hline
            \end{tabular}
            \end{footnotesize}
            \caption{Success rates are being compared against a single NN and ROMAN's preliminary stage in 2D with five experts. A total of $N=20$ demonstrations ($t \approx 5 \text{min}$) were provided for each expert and a total of $N=35$ demonstrations ($t \approx 20 \text{min}$) for the MN. A total of $N=100$ demonstrations were provided to the single NN ($t \approx 64 \text{min}$). \textbf{Note:} $N=35$ demos for the gating network correspond to 7 demonstrations for each of the five derived case scenarios.} 
            \label{table:2DCase_SingleNN_VS_RMN}
	\end{subtable}
	\hfill
	
	\begin{subtable}[h]{1.0\textwidth}
	\vspace{0.25cm}
		\centering
		\begin{footnotesize}
        \setlength\tabcolsep{8.25pt}
        \begin{tabular}{l|rrrrrrr} \bottomrule
        \rowcolor{Gray} \multicolumn{8}{c}{{\textbf{[3-D] Single NN versus ROMAN on Case Scenarios}}} \\ \hline
        
        \cellcolor{gray!20}  $\sigma = \pm 0.5 \text{cm}$ & \textbf{S1} & \textbf{S2} & \textbf{S3} & \textbf{S4} & \textbf{S5} & \textbf{S6} & \textbf{S7} \\ \hline
        Single NN & {0.997} & {0.981} & {0.583} & {0.032} &  {0.028} &  {0.000} &  {0.000}\\
        ROMAN &  {0.973} & {0.975} & {0.817} & {0.959} &  {0.852} &  {0.960} &  {0.952}\\
        \hline
        \end{tabular}
        \end{footnotesize}
		\caption{Success rates across all seven scenarios, between a single NN and ROMAN's final stage in 3D with seven experts. A total of $N=20$ demonstrations were provided to each agent in ROMAN and a total of $N=42$ demonstrations to the MN. \textbf{Note:} A total of $N=140$ demonstrations were provided to the single NN in 3D ($t \approx 132 \text{min}$).}
        \label{table:SingleNN_VS_RMN}
	\end{subtable}
	\hfill
	
	\begin{subtable}[h]{1.0\textwidth}
	\vspace{0.25cm}
		\centering
		\begin{footnotesize}
        \setlength\tabcolsep{7.5pt}
        \begin{tabular}{l|rrrrrrr} \bottomrule
        \rowcolor{Gray} \multicolumn{8}{c}{{\textbf{Algorithm Comparison in ROMAN}}} \\ \hline
        \cellcolor{gray!20} $\sigma = \pm 0.5 \text{cm}$ & \textbf{S1} & \textbf{S2} & \textbf{S3} & \textbf{S4} & \textbf{S5} & \textbf{S6} & \textbf{S7} \\ \hline
        HRL: RL & {0.000} & {0.000} & {0.000} & {0.000} &  {0.000} &  {0.000} &  {0.000}\\        
        HIL: GAIL & {0.980} & {0.468} & {0.559} & {0.012} &  {0.003} &  {0.001} &  {0.000}\\        
        HIL: BC & {0.986} & {0.978} & {0.786} & {0.660} &  {0.525} &  {0.722} &  {0.760}\\
        HHL: RL,GAIL & {0.981} & {0.468} & {0.570} & {0.009} &  {0.005} &  {0.006} &  {0.004}\\        
        HHL: RL,BC & {0.995} & {0.897} & {0.841} & {0.683} &  {0.492} &  {0.754} &  {0.774}\\
        \textbf{ROMAN's} $\dagger$ & {0.973} & {0.975} & {0.817} & {0.959} &  {0.852} &  {0.960} &  {0.952}\\
        \hline
        \cellcolor{gray!20} $\sigma = \pm 1.0 \text{cm}$ & \textbf{S1} & \textbf{S2} & \textbf{S3} & \textbf{S4} & \textbf{S5} & \textbf{S6} & \textbf{S7} \\ \hline
        HIL: BC & {0.995} & {0.990} & {0.712} & {0.573} &  {0.474} &  {0.563} &  {0.632}\\
        HHL: RL,BC & {0.996} & {0.895} & {0.881} & {0.766} &  {0.562} &  {0.696} &  {0.729}\\ 
        \textbf{ROMAN's} $\dagger$ & {0.977} & {0.990} & {0.798} & {0.946} & {0.776} &  {0.933} &  {0.939}\\ \hline
        \cellcolor{gray!20} $\sigma = \pm 2.0 \text{cm}$ & \textbf{S1} & \textbf{S2} & \textbf{S3} & \textbf{S4} & \textbf{S5} & \textbf{S6} & \textbf{S7} \\ \hline
        HIL: BC & {0.838} & {0.678} & {0.609} & {0.205} &  {0.190} &  {0.111} &  {0.075}\\
        HHL: RL,BC & {0.947} & {0.841} & {0.725} & {0.442} &  {0.363} &  {0.246} &  {0.100}\\
        \textbf{ROMAN's} $\dagger$ & {0.967} & {0.986} & {0.737} & {0.837} & {0.753} &  {0.820} &  {0.815}\\ \hline
        \end{tabular}
        \end{footnotesize}
		\caption{Success rates across all seven scenarios, between different comparisons of HRL, HIL and their combinations. \textbf{Note BC:} Supervised learning on the demonstration dataset. \textbf{Note GAIL:} Use of IL, intrinsic rewards ($r_{I}$) provided to PPO. \textbf{Note on RL:} Use of task extrinsic rewards ($r_{E}$), provided to PPO. \textbf{ROMAN's $\dagger$:} Default HHL approach combining BC, IL (via $r_{I}$) and RL (via $r_{E}$). Tested on $\sigma = \pm 0.5 \text{cm}$ noise, increasing to $\sigma = \pm 1.0 \text{cm}$ and $\sigma = \pm 2.0 \text{cm}$ for algorithms scoring high. Where BC or GAIL is used, the same number of demonstrations ($N=42$) were employed.}
        \label{table:ROMANAlgorithmComparison}
	\end{subtable}
	\hfill
	
	\begin{subtable}[h]{1.0\textwidth}
	\vspace{0.25cm}
		\centering
        \begin{footnotesize}
        \setlength\tabcolsep{9.5pt}
        \begin{tabular}{l|rrrrrrr} \hline
        \rowcolor{Gray} \multicolumn{8}{c}{{\textbf{Demonstration Comparison N=7, 21 and 42 on Case Scenarios}}} \\ \hline
        \cellcolor{gray!20} \textbf{Total Demo No.} & \textbf{S1} & \textbf{S2} & \textbf{S3} & \textbf{S4} & \textbf{S5} & \textbf{S6} & \textbf{S7} \\ \hline
        N = 7 ($t \approx 7 \text{min}$) & {0.775} & {0.876} & {0.680} & {0.378} &  {0.360} &  {0.008} &  {0.005}\\
        N = 21 ($t \approx 25 \text{min}$) & {0.994} & {0.921} & {0.718} & {0.945} &  {0.903} &  {0.929} &  {0.958}\\
        N = 42 ($t \approx 42 \text{min}$) & {0.973} & {0.975} & {0.817} & {0.959} &  {0.852} &  {0.960} &  {0.952}\\
        \hline
        \end{tabular}
        \end{footnotesize}
\caption{Success rates based on demonstrations provided to the MN. \textbf{Note:} The total number is divided by the number of scenarios, i.e. $N=7$, $N=21$ and $N=42$ correspond to 1, 3 and 6 demonstrations per case. Evaluated on $\sigma = \pm 0.5$ level of noise.}
\label{table:DemonstrationComparison}
	\end{subtable}
	\label{tab:label_all_table}
\end{table}

\subsubsection*{\textbf{Adaptation to Recover from Local Minima}}
\begin{figure}
     \centering
     \centering
         \includegraphics[width=\textwidth]{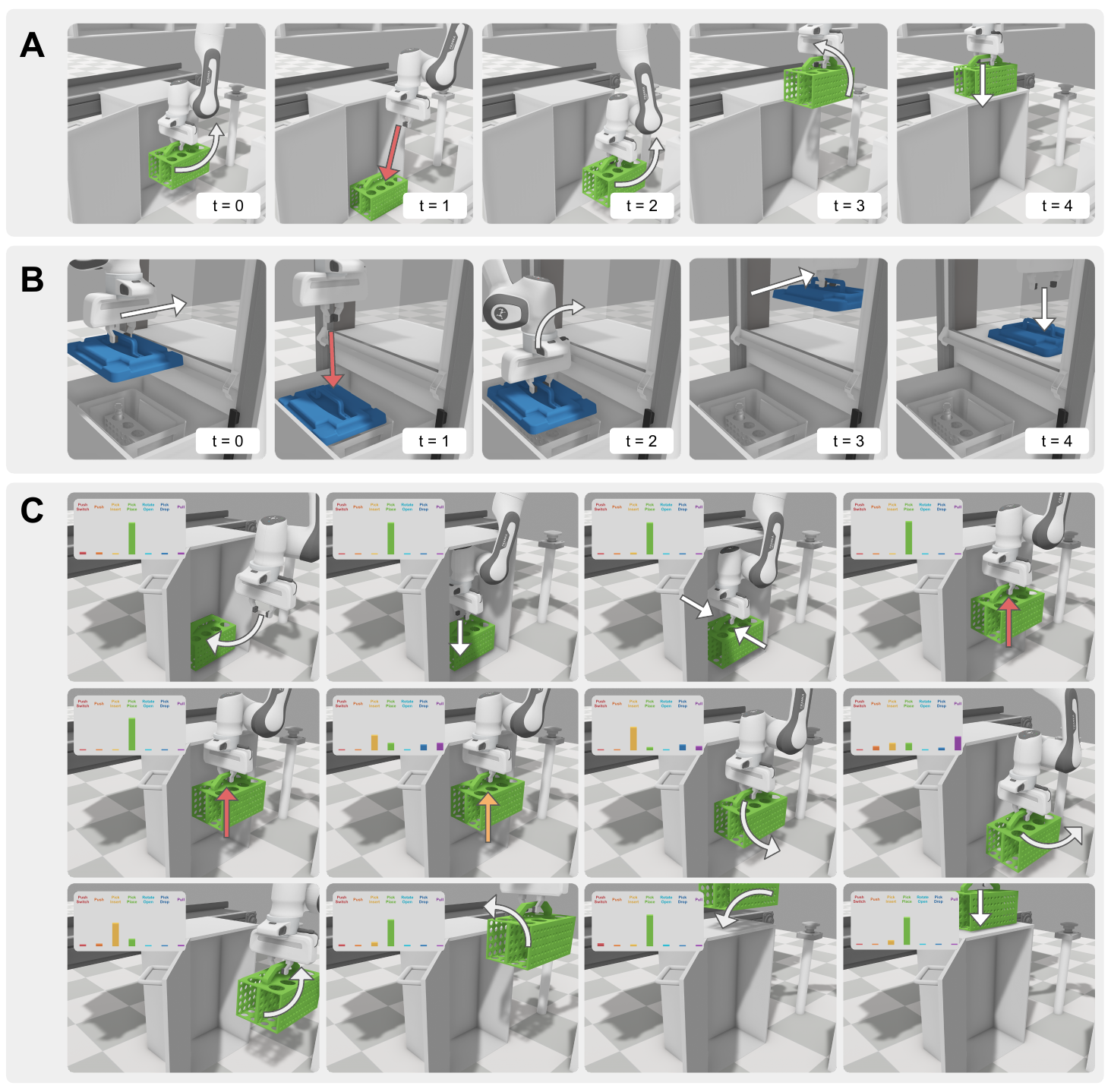}
    \caption{\textbf{ROMAN's ability to adapt to the scenarios beyond the demonstrated sequence and exhibiting behaviour beyond imitation with the most notable one being the dynamic recovery capabilities shown, by virtue of balancing exploitation and exploration via the employed HHL approach.} \textbf{Figures (A) and (B):} Policy adaptation of ROMAN during failures concerned with \textit{Picking and Placing} as well as \textit{Pick and Dropping} sub-tasks respectively. These intermediate failures are either attributed to individual expert error or a gating network error. In such seldom instances, we show the error cases ($t=1$) of these experts, which however quickly and dynamically re-adapt and re-grasp the items ($t=2$ to $t=4$) to successfully complete the sequence and more broadly the end goal. \textbf{Figure (C):} The ability of the MN of the ROMAN framework to dynamically adapt in cases that were not encountered in the demonstrated sequence, but rather visited states during the RL training as the result of balancing exploitation and exploration from the employed hybrid learning procedure. This balance ultimately resulted in new behaviours beyond imitation, leading to recovery capabilities from local minima. The figure represents 12 snapshots over time with a sequence from left to right and top to bottom, depicting and highlighting the weight assignments by the MN.}
    \label{figure:AdaptationToPickingFailures}
\end{figure}

As part of the experimental evaluation of ROMAN, it was observed that occasionally experts could fail in retaining a firm grasp when manipulating certain objects of interest in the environment, resulting, in turn, of grasped objects to drop. As shown by the success rates, this occurred in a fairly infrequent number and was primarily limited to more complex experts concerned with \textit{Picking} tasks. Further investigation found that when such rare expert-level failures occurred, the MN began to recognise the sub-task state and gradually learned a new weight assignment until the tasks were rendered successful. The use of the HHL approach which in turn balances exploitation and exploration, resulted in the hierarchical architecture of ROMAN to enable a positive adaptation of the learning agent to commence a re-grasping procedure. This adaptation of the framework is visually depicted in \autoref{figure:AdaptationToPickingFailures}.a as well as \autoref{figure:AdaptationToPickingFailures}.b.

Moreover, the MN within the hierarchical architecture of ROMAN exhibited the ability to learn to recover from local minima by rapidly switching experts during their sequential activation when it was necessary to do so. During the sequential activation of the entailed seven experts, the robotic gripper could occasionally get stuck under the cabinet while retrieving the rack, rendering it stuck under the obstacle. During such rare cases, the MN would activate other experts in order to alter the trajectory and move the gripper away from the cabinet until it was collision-free, and then recommence the task successfully, as shown in \autoref{figure:AdaptationToPickingFailures}.c. It is important to note that this was not directly demonstrated by the expert demonstrator but was rather the result of the employed hybrid learning procedure within ROMAN balancing exploration and exploitation. This result highlights the value of combining the advantages of IL and RL paradigms and leveraging intrinsic and extrinsic rewards, resulting in a robust performance in cases not encountered in the demonstrations and going beyond demonstrated behaviour. 

We can hence infer that a balance between imitating the demonstrations and the random exploration to maximize the extrinsic RL reward, is beneficial for the framework and detrimental to its success in unforeseen cases. This balance exhibited by ROMAN draws inspiration from a biological perspective as identified in the state-of-the-art \cite{Zador2019, Saxe2021}.

\subsection*{\textbf{T-SNE Analysis of the Similarity of Sequences}}
\begin{figure}
     \centering
     \centering
         \includegraphics[width=\textwidth]{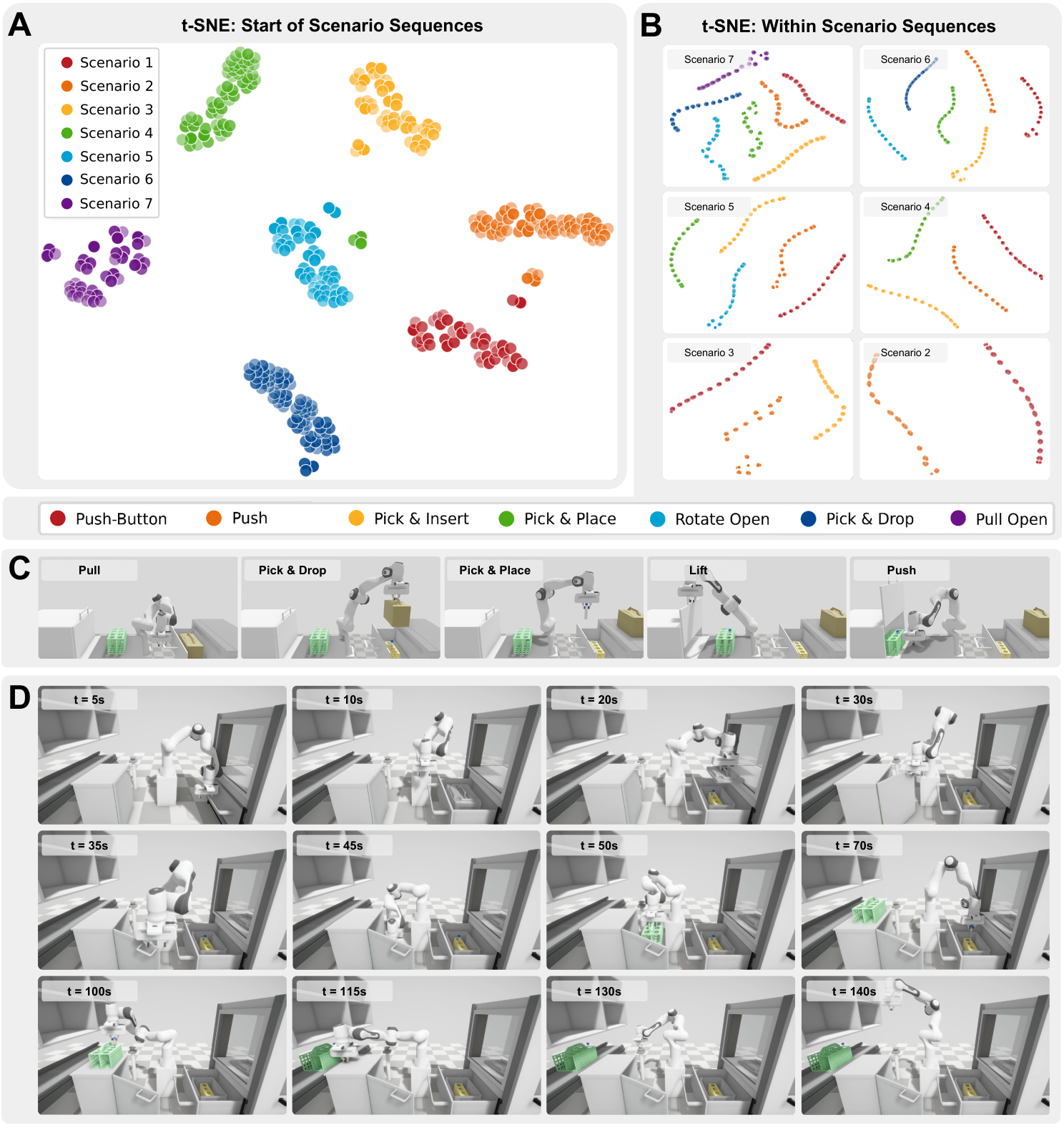}
         \caption{\textbf{The analysis of the MN observations using the t-Distributed Stochastic Neighbour Embedding (t-SNE), with visualised snapshots showing ROMAN's completion of sequential tasks in 2D as well as 3D space.} The t-SNE is projecting the 29-dimensional MN state vector into 2 dimensions. Principal Component Analysis (PCA) was used to warm-start the t-SNE projection. \textbf{Figure (A):} The depiction of the state vectors at the \textit{start} of each of the seven case scenarios, sampled at 1000Hz for 1 s. A grand total of 1000 samples were projected with a perplexity of 400. \textbf{Figure (B):} The illustration of the state vectors \textit{during} the sequence of actions contained in each case scenario, sampled for the first 1.5 s of each expert sequence. Hence, as these are sampled \textit{within} the sequence of actions, they appear ``trajectory"-like, since the robot and the objects manipulated by it are already in motion during the sampling. A total of 1500 samples were projected with a perplexity of 200. Six out of seven scenario cases are depicted as in practice the S1 case only includes a single expert activation and hence is omitted from the analysis. \textbf{Figure (C):} ROMAN in its initial 2D stage depicting the total five distinct sub-tasks managed by each expert respectively. \textbf{Figure (D):} ROMAN in its final stage in the most complex setting and longest time-horizon sequential tasks.}
    \label{figure:TSNE-Depiction-ROMAN-Snapshots}
\end{figure}

Following the performance evaluation of the MN at sequential activation of the different experts in different scenarios, we further analysed the similarities and patterns. Specifically, in most of the results from S3, S4 and S5, ROMAN exhibited lower success rates, compared to that in S6 and S7 which are more complex tasks with longer time horizons. Consequently, to qualitatively study the MN's ability to activate the necessary expert activations based on its observations, we conducted dimensionality reduction via a t-SNE. This allowed us to evaluate the similarities in the observations of the MN and its ability to distinguish between different scenarios. The t-SNE plots are shown in \autoref{figure:TSNE-Depiction-ROMAN-Snapshots}.a and  \autoref{figure:TSNE-Depiction-ROMAN-Snapshots}.b.

First and foremost, a t-SNE was conducted on the MN observations at the commencement of each scenario to analyse the similarities between the MN observations in different case scenarios. As shown in \autoref{figure:TSNE-Depiction-ROMAN-Snapshots}.a, scenarios S1 to S7 differ to a great degree, and S3, S4 and S5 present a slight overlap between each other due to the state vectors between these three being relatively similar. This in turn explains to a great degree as to why the MN may not always activate the correct sequence, particularly at the beginning of a sequence when the robotic end-effector's start position is randomised (as opposed to being the ending position of a previous sub-task), leading ultimately to slightly lower success rates. A potential mitigation strategy for this problem would be to endow ROMAN with some form of ``memory", either explicitly by expanding the observation space to include a buffer of its past actions, or implicitly by employing some recurrent NN architecture for the MN. This was left as part of our future work.

Secondly, a t-SNE was conducted on the MN observations of each separate activation for every scenario studied. \autoref{figure:TSNE-Depiction-ROMAN-Snapshots}.b reveals the similarities in the MN observations throughout different expert activations in each of the seven case scenarios. By sampling within the sequence of actions as opposed to the beginning as with the aforementioned paragraph, we obtain a low-dimensional projection of the trajectory of the MN observation vectors during the expert activations. In essence, this is due to the change in the spatial states of the objects in the scene and the end-effector being in motion during the sequence of actions. 

Overall, from \autoref{figure:TSNE-Depiction-ROMAN-Snapshots}.b, it can be inferred that no significant overlaps between the activation of the different experts within each scenario are shown. These observations suggest that the MN is capable of distinctly activating experts during the sub-task completion. Consequently, it can be concluded that the decreased performance for S3, S4, and S5 observed in \autoref{table:SuccessRatesNoiseLevels} and \autoref{tab:label_all_table} are due to the slight overlap between MN observations analysed in \autoref{figure:TSNE-Depiction-ROMAN-Snapshots}.a. In particular, we conclude that the failures that account for the slight drop in performance occurred at the \textit{beginning} of the sequences due to the randomised initialisation.

\subsection*{Discussion}
In this section, we discuss our findings as part of our evaluation and the results, as well as additional observations from the evaluation of ROMAN. More details regarding the results and further expansion can be found in the supplementary materials.

From the evaluation and the experiments, it is shown that the hierarchical task decomposition of ROMAN allows task-level experts to be trained to achieve robust performance in significantly complex sequential tasks. This higher-level task decomposition which is made possible by the hybrid learning procedure ultimately enables the MN to focus on the orchestration of the incorporated high-level experts, rather than low-level skills, thereby offloading unnecessary complexity from the MN. From the results at hand, it is shown that ROMAN can orchestrate significantly more complex sequential tasks of longer time horizons and higher dimensionalities than similar work in physics-based manipulation \cite{9659344, Rajeswaran-RSS-18, 8843293}.

Furthermore, ROMAN's HHL architecture (see \autoref{figure:NN_MN_StructureOverview}) led to the overall framework achieving successful adaptation to non-encountered scenarios and recovery from local minima that were not explicitly demonstrated. Hence, these results suggest that although IL is effective in providing a baseline, achieving a balance between imitating the demonstrations and maximising the extrinsic RL reward through random exploration as per the RL paradigm, is crucial for successful subsequent adaptation beyond the demonstrated behaviours. This balance between exploration and exploitation provided by ROMAN also shares the common ground from biological studies \cite{Zador2019, Saxe2021}. Consequently, this balance between the exploration and exploitation trade-off in the hybrid learning procedure may indicate that even non-optimal demonstrations i.e. stemming from humans, can still be employed and potentially improved upon via the random exploration of the agent.

Finally, from the results it was observed that ROMAN's central MN was able to solve and attain robustness in the most complex and longest horizon sequential manipulation tasks skillfully. Further investigation also revealed a slight performance drop in some of the tasks with lower complexity, such as S3 to S5 compared to more complex ones such as S6 and S7. The t-SNE analysis concluded that this is primarily due to the difficulties of the MN to differentiate between those states at randomised initialisation of tasks. Future work can explore more sensory feedback to differentiate ambiguous cases, or incorporate a ``memory" mechanism by expanding the observation with history states, potentially leading to better differentiation and ultimately performance.

\paragraph*{Future Work}
Part of our future work is to extend our current framework to higher-dimensionality problems, including multi-expert hierarchical learning as well as tasks requiring bi-manual operation. Leveraging immersive technologies such as VR and MR are also a promising means of providing demonstrations and rendering the execution of complex tasks for human demonstrators easier than conventional input devices \cite{9492850, 9076603}. Another improvement for future work would be the addition of some form of memory to our framework in order to further improve performance at complex sequential tasks.

Another potential future work would be the derivation of the lowest possible level of ``fundamental" manipulation skills. While this may require the use of a hierarchical framework with an increased number of levels within its hierarchy, it may be beneficial in terms of re-using demonstrations as shown in related work \cite{8843293}. The work of \cite{8843293} focused on re-using demonstrations by decomposing tasks into their fundamental primitives so as to limit the need for new demonstrations in related domestic tasks such as clearing and setting a table. While their tasks were concerned with significantly easier complexities of lower time-horizons when compared to ROMAN, their approach is potentially a promising improvement we should consider to offload the need to re-demonstrate sub-tasks. 

Lastly, to enable real-world deployment in future work, a vision system for exteroceptive information would be needed to predict object poses: for example, using AprilTags, or segmenting/detecting objects using RGB/RGB-D cameras. Additionally, a dynamic grasping controller that incorporates force control could further enhance the grasping performance. 

Overall, from the evaluation, we can support that ROMAN is able to solve complex sequential tasks with generalizing capabilities. In particular, the results validate the robustness of the ROMAN framework and its HHL approach against (i) large exteroceptive observation noise, (ii) the presence of many complex non-interrelated compositional sub-tasks, (iii) long time-horizon sequential tasks as well as (iv) cases not seen during demonstrations by dynamically coordinating experts to recover from local minima.

\section*{Methods}
In this section, the methodology and technical details of ROMAN are described. ROMAN is characterised by a hybrid hierarchical learning approach. In this architecture, multiple experts specialise in diverse and fundamental types of manipulation tasks. When these experts are subsequently activated, in a correct sequential order, by a primary gating network, known as the MN, significantly complex of long horizon robotic manipulation tasks can be solved, attaining robust performance even under increasing levels of exteroceptive uncertainty. The validation of ROMAN will by definition be among different types of manipulation tasks commonly seen in robotics and physics-based interactions.

\subsection*{System Overview}
The hierarchical architecture of ROMAN is validated in a complex medical laboratory setting, in order to highlight our approach in a setting where manipulation typically consists of (i) careful handling of small objects with high precision, (ii) the necessity to perform multiple tasks and (iii) the correct sequence of tasks to complete a long and complex end-goal. The construction of the environment was done in such a way as to derive as many sub-tasks as possible and validated our method. In regards to the robotic system, we used the 7 Degrees of Freedom (DoF) Franka Emika Robot in simulation with its default gripper configuration in 3D space, based entirely on physics-based interactions with the environment. The system overview entailing the simulation environment and the overall depiction of the ROMAN framework are visually depicted in \autoref{figure:NN_MN_StructureOverview}. Moreover, the architectural overview of the incorporated NNs in the ROMAN framework, including their individual states, actions, number of demonstrations and training times are shown in \autoref{table:DetailedNetworkCharacteristicsStructure}. For further details regarding the system and simulation overview, incorporated software tools \cite{juliani2018unity}, including the general apparatus, can be found in the Supplementary Notes and more specifically Supplementary Note S1. 

\subsection*{Vision System}
\label{visionSystem}
As part of the preliminary investigation, a vision system using an RGB camera was implemented in the simulation to predict the poses of the different Objects of Interest (OIs). The vision system implements an object detection and pose estimation module based on the VGG-16 backbone architecture \cite{8202133}. The system was initialised with pre-trained weights on the ImageNet dataset and fine-tuned using a custom dataset, which was in turn created by capturing the OIs from the simulated environment, including both the segmentation and labelling of the OIs. The output of the network predicted the poses of all OIs, specifically their 3D positions (X, Y and Z).

The rationale for the inclusion and testing with a camera setup was to validate ROMAN's robustness in a realistic setting and also underline the flexibility of the framework to operate to closer to real-life cases. In such a setting, pose prediction errors and visual occlusions naturally occur. When the target objects were occluded, the last known position was provided to the gating network, i.e. the MN. Since the pre-trained object detection module from the vision system attained variable levels of positional error \cite{8202133}, we simulated increasing levels of Gaussian distributed noise to all exteroceptive observations of all NNs. This in turn allowed for further testing ROMAN's capabilities besides its exhibited robustness to a vision system, which is in line with related work \cite{Rajeswaran-RSS-18, 8843293}. Overall, by introducing exteroceptive uncertainties, it was further made possible to assess the resilience of the ROMAN framework and highlight the importance of a hybrid learning approach within a hierarchical architecture for solving complex sequential robotic manipulation tasks.

\subsection*{Hybrid Learning Procedure and Learning Preliminaries}
\begin{figure}
     \centering
     \centering
         \includegraphics[width=\textwidth]{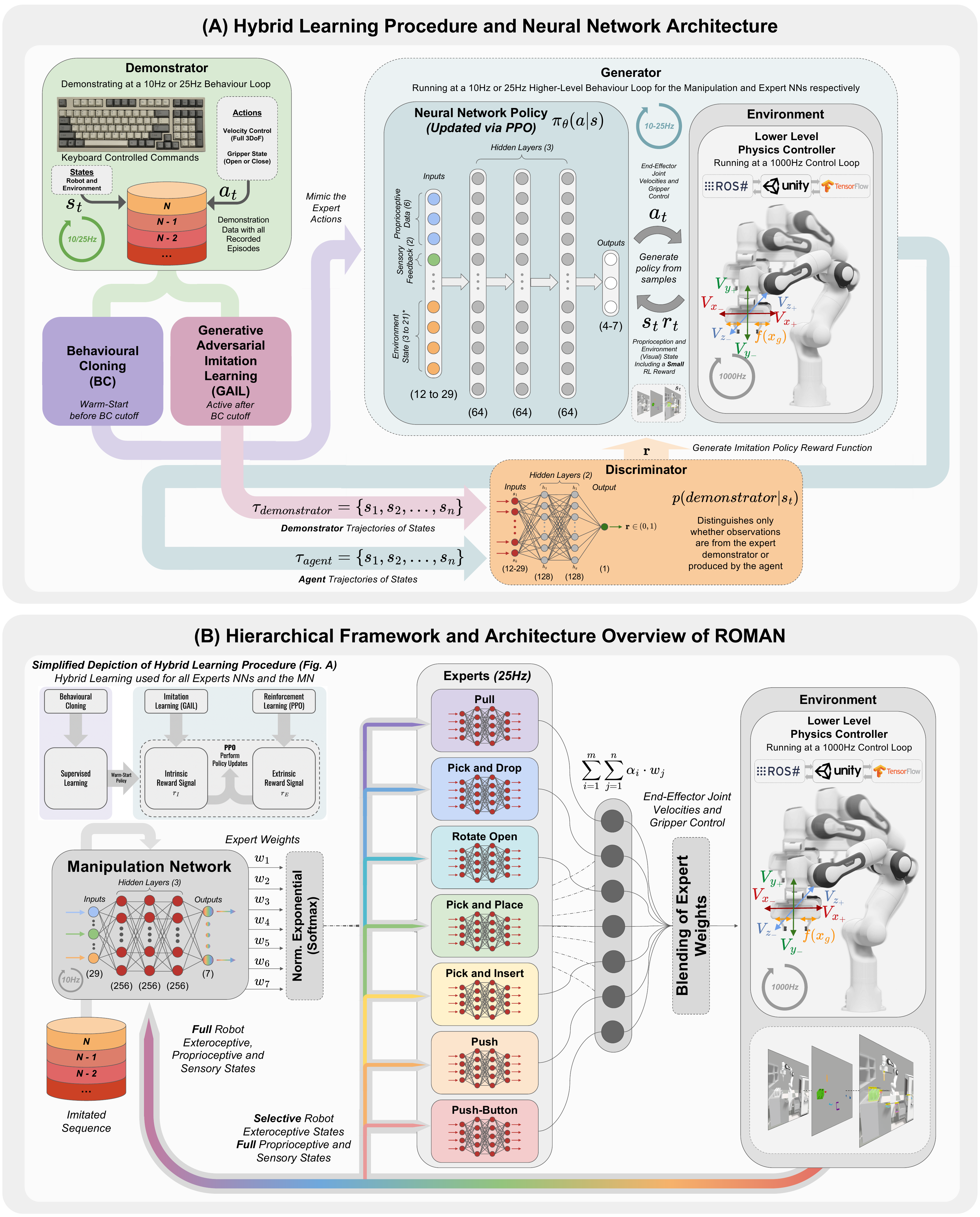}
    \caption{\textbf{The hybrid hierarchical architecture of ROMAN composed of high-level experts and the central gating network depicting the overall formation of the hierarchical framework.} \textbf{Figure (A):} The hybrid learning architecture of each high-level expert and gating network, the MN. \textbf{Figure (B):} The higher hierarchical formation of ROMAN and how the experts are orchestrated and activated by the centrally governing MN. The Multi-Layer Perceptron (MLP) is visually depicted for all NNs in both figures.}
    \label{figure:NN_MN_StructureOverview}
\end{figure}

Two imitation learning algorithms were employed, (i) Generative Adversarial Imitation Learning (GAIL) \cite{NIPS2016_cc7e2b87}, as well as (ii) Behavioural Cloning (BC) \cite{10.5555/3304652.3304697}. These two algorithms, coupled with the Reinforcement Learning (RL) algorithm Proximal Policy Optimization (PPO) \cite{schulman2017proximal}, allowed the framework and all incorporated NNs within the hierarchical formation of ROMAN to successfully and robustly imitate complex robotic tasks for the purpose of autonomous robotic operation entailing purely physics-based interactions with multiple sequential tasks. In particular, the training procedure is composed of two stages: in stage one, the policy is warm-started using BC; in stage two, the policy is updated via the PPO algorithm with rewards $r_E$ and $r_I$ stemming from the environment (RL) and from the discriminator network (GAIL) respectively. The hybrid learning procedure used for both the expert NNs and the MN in ROMAN is visually illustrated and detailed in \autoref{figure:NN_MN_StructureOverview}.a. \autoref{figure:NN_MN_StructureOverview}.b depicts the hierarchical framework formation and the architectural overview of ROMAN and how the MN is tasked with the supervision of the incorporated experts.

\paragraph{\textbf{Behavioural Cloning (Warm-Starting the Policy):}} First and foremost, to warm-start the policy, we used BC up until a given number of initial epochs. The cut-off point for BC was determined via preliminary investigations and training sessions on the performance of the policy and the complexity of the sequential tasks. Most notably, the cut-off point of BC was increased when transitioning from the 2D to 3D version of ROMAN to account for the increased complexity and increased number of experts in the hierarchical formation. The exclusive use of BC throughout the training process was avoided, so as to allow the agent to explore further samples and improve upon demonstrated behaviours, while keeping the demonstration dataset small \cite{NIPS2016_cc7e2b87, 8461249}. This decision was made and inspired by the state-of-the-art that BC is limited in its ability to generalise to out-of-distribution states, and thus is restricted to the trajectories seen in the provided demonstrations \cite{NIPS2016_cc7e2b87, reddy2019sqil}. Most notably, this can ultimately lead to drifting errors when the agent encounters new trajectories outside of those in the demonstrations \cite{NIPS2016_cc7e2b87, codevilla2019exploring}. In line with previous work concerned with robotic manipulation, the sole dependence on BC should be avoided. Instead, a viable alternative is to incorporate a reward term when computing a separate RL gradient that corresponds to the BC loss \cite{Rajeswaran-RSS-18}. In this work, using a dataset of state and action transitions ${s^{d}_t, a^{d}_t}$ provided by the demonstrator, we implement BC by training a NN policy $\pi(s_t) = a_t$ using supervised learning to minimise the mean-squared error (MSE) loss between $a^{d}_t$ and $a_t$ for the demonstration dataset.

\paragraph{\textbf{GAIL (Commenced after BC and Active Throughout Training): }} To effectively match the provided human demonstration dataset over a period, also known as a horizon, we made use of inverse RL and in this case, GAIL \cite{NIPS2016_cc7e2b87}. Contrary to BC, GAIL was used after BC's cutoff point, at which point GAIL commenced and was active throughout training to attempt in minimising the divergence between the agent's policy and that of the expert demonstrator. However, it is important to point out that GAIL was not directly used to update policy parameters, instead a proxy imitation reward signal obtained by GAIL was used, described further in this section. 

This is achieved by sampling a set of expert ($\tau_E$) and agent ($\tau_A$) trajectories of states and actions $(s_t, a_t)$. The expert trajectories are sampled from a provided demonstration dataset while the agent trajectories are sampled from a generative model also known as Generator (G). The generator, however, instead of being rewarded solely by the environment, is instead being rewarded by a scalar score provided by the Discriminator (D), implemented as a separate NN in this process. In this procedure, the discriminator attempts to differentiate between the expert and agent trajectories, rewarding the generator if the divergence between these trajectories decreases. The discriminator is also trained to become ``stricter" over time, resulting in the Generator, e.g. agent, improving its performance at imitating and converging towards the behaviour that was demonstrated by the human expert. This process can be formulated as follows:
\begin{equation}
\label{equation:GAIL}
\begin{gathered}
     E_{\tau_E}[\nabla log(D(s_t,a_t))]+E_{\tau_A}[\nabla log(1-D(s_{t},a_{t}))]
\end{gathered}
\end{equation}
where $E_{\tau_E}$ and $E_{\tau_A}$ represent the expert and agent trajectories from the training, which are represented as inputs to the discriminator network ($D$). The discriminator outputs a continuous value ranging between 0 and 1, with a value closer to 1 indicating that the agent or generator, is resembling a trajectory closer to that of the expert's. This process, in turn, essentially minimises the divergence between the two sets of trajectories and maximises imitation over time. Consequently, D can be used as a reward signal to train G to mimic the expert's demonstrated data. Furthermore, to allow the agent to further explore additional actions that can potentially lead to improved performance on what was demonstrated upon, we modify the above formulation for the discriminator to only use the states ($s_{t}$) but not the actions ($a_{t}$) of the demonstrated trajectories. This ultimately leads to increased exploration of the agent which should encourage behaviours beyond those encountered in a demonstrated sequence when coupled with RL. More details are described in the Results and Discussion sections. 

Consequently, we reformulate as with \cite{NEURIPS2018_943aa0fc}, \autoref{equation:GAIL} as:
\begin{equation}
\label{equation:GAILModified}
\begin{gathered}
     E_{\tau_E}[\nabla log(D(s_t))]+E_{\tau_A}[\nabla log(1-D(s_{t}))].
\end{gathered}
\end{equation}
Sampling only the states for GAIL allowed the policy to be less restrictive in terms of imitation. Discriminating against both states and actions between the demonstrator and the expert as with the original formulation of GAIL \cite{NIPS2016_cc7e2b87}, would have potentially led to disallowing the agent to further explore other actions. These actions could in actuality lead to better adaptation based on the state space and avoid a ``naive" copying of identical actions during imitation learning.

The result of using the above two IL algorithms with slight modifications to GAIL, translated into a significantly reduced necessary dataset, compared to related work to train the agents successfully in complex long-horizon sequential tasks \cite{8461249, 9659344}.

\paragraph{\textbf{Reinforcement Learning (Exploration Beyond Imitation):}} In addition to the IL approaches outlined above, a small task-related extrinsic reward signal was further used. We use extrinsic rewards to provide a small contribution towards the final policy to avoid exclusive dependence on pure imitation and encourage exploration and ultimately balancing exploration and exploitation. As described below, we use intrinsic rewards stemming from IL (GAIL) as well as extrinsic task-related rewards as per the RL paradigm, to update the policy. It is noteworthy to point out that before updating the policy, the intrinsic and extrinsic rewards are being scaled, with the weight of the intrinsic reward (i.e. IL) being the main learning signal provider. Most notably, the resulting HHL architecture exhibited the ability to adapt to new cases that were not encountered during the demonstrated sequence by the human expert and further attain resilience in the presence of sensor uncertainty. More specifically, this allowed the ROMAN framework to recover from local minima during the most complex sequence activation of experts, even when the sequence is not activated precisely or seldom errors occur during the individual expert performance. We chose PPO as our RL algorithm because it is robust and flexible across various hyperparameter settings, as supported by the related work \cite{pmlr-v80-haarnoja18b, schulman2017proximal, 10.5555/3294996.3295141}.

Denoting our policy $\pi_{\theta}$ as a NN parameterised by weights $\theta$, the PPO update at step $k$ is given by:
\begin{equation}
\label{equation:PPO}
    \begin{gathered}
    \theta_{k+1} = \arg \max_{\theta} \mathbb{E}_{s,a \sim \pi_{\theta_k}} \left[
    L(s,a,\theta_k, \theta)\right]
\end{gathered}
\end{equation}
with a clipped loss function $L(s,a,\theta_k, \theta)$ that has a surrogate term, a value term and an entropy term \cite{schulman2017proximal}.

\paragraph{\textbf{Integration of BC, GAIL and RL: }} 
\label{par:FusingBC_GAIL_RL}
In order to learn to solve long and complex sequential tasks using a low number of demonstration data, we integrate a set of algorithms for an effective balance between exploitation and exploration. While using BC, we perform supervised learning on the policy using the demonstrations as a dataset, i.e. policy updates are driven by the MSE loss on the demonstration dataset. While using GAIL and or RL, we use the PPO algorithm in this process as the general-purpose algorithm to perform policy updates. We thereafter combine these methods by using, as aforementioned, different reward terms for intrinsic rewards $r_{I}$ and extrinsic rewards $r_{E}$, where intrinsic rewards are provided by the discriminator score from GAIL, and extrinsic rewards are provided by the environment as per the RL formalism. 

In regards to GAIL, as mentioned above, we modify the original framework to only use states in the discriminator, instead of states and actions, hence making use of \autoref{equation:GAILModified}. We define the intrinsic reward term as $r_{I} = -log(1-D(s_t))$, whereby $D(s_t) \in (0,1)$ is the score provided by the discriminator, which acts as a proxy reward term that can be used by PPO to ultimately maximise the GAIL objective. When training with GAIL and RL, we use a linear combination of reward terms such that $r = r_{I} \cdot w_{I} + r_{E} \cdot w_{E}$, with $w_{I}$ and $w_{E}$ as fixed scaling parameters for intrinsic and extrinsic rewards respectively. Our HHL control policy focuses more on imitation, i.e. on the intrinsic provided reward compared to the extrinsic reward. More specifically, the rewards for $r_{I}$ are several magnitudes larger than that for $r_{E}$ ($w_{I} > w_{E}$). Using the latter reward combination, the returns are computed as the discounted sum of rewards which are used for the PPO update on the policy as in \autoref{equation:PPO}. 

ROMAN's robustness is attributed to the above employed hybrid learning architecture. In particular, this process consists of the combination of (i) using BC up to a given epoch for warm-starting policy optimisation, (ii) thereafter using the intrinsic reward provided by GAIL to further minimise the divergence of the agent and that of the expert demonstrator, and finally (iii) the addition of an extrinsic reward term from the RL paradigm to allow the agent to explore further and beyond what was demonstrated upon. 

The individual NN architecture of each expert and the manipulation network incorporated in ROMAN's hierarchical architecture, are depicted in \autoref{figure:NN_MN_StructureOverview}.a and \autoref{figure:NN_MN_StructureOverview}.b respectively. \autoref{figure:NN_MN_StructureOverview}.b illustrates the hierarchical formation of ROMAN and more specifically, that the exteroceptive information provided to each NN from the environment is determined by the individual objective of each expert and the relevance of that information for the successful completion of the given sub-task goal. In contrast, the MN observes the entirety of the environment, overseeing all experts and sub-task related goals as to infer their necessary sequential activation of achieving the primarily end goal of the sequential task.

\subsection*{Demonstration Acquisition and Settings}
The demonstrations provided to the NNs were achieved via keybindings stemming from a generic keyboard. This was coupled with a commercially available 2D display monitor to allow the human expert to imitate the tasks and visually observe the behaviour. Two cameras in an orthographic projection were rendered onto the monitor, visually depicting the environment from an upper and side-view perspective. This design decision, regarding the visual overseeing of the simulation by the human, was made to allow the determination of depth-associated distances in the simulation significantly easier than without, in line with previous work \cite{9076603, 10.1145/3290605.3300437, 10.1145/3411763.3443442}. An alternative to this visual design would have been projecting the environment visually with either 3D displays or Mixed Reality (MR) technologies similar to related work on tele-manipulation \cite{9076603, 8461249}. The use of more immersive technologies with better depth estimations was left as part of our future work.

\subsection*{The Sequential Task}
The incorporated physics engine, NVIDIA PhysX, in the simulation, allowed the derivation of numerous tasks all containing physical properties with advanced physical characteristics such as hinges, linearly moving objects as well as spring joints. However, more complex mechanical tasks such as unlocking a lock were disregarded due to their significantly more difficult mechanical design in the simulation and their lack of relevance for the overall studied system validation. The full task as seen in the simulation is visually shown in \autoref{fig:CoverImage}. Moreover, the full sequence decomposed into its relevant sub-tasks can be visually observed in \autoref{figure:TSNE-Depiction-ROMAN-Snapshots}.d.

The task we conceived was based on a medical laboratory setting which allowed the derivation of specialised manipulation experts of varying nature and objectives. The main goal of the sequential task was to retrieve a small vial, insert it into a rack and push it all together onto a conveyor belt. Within this three-action sequence, we derived additional sub-tasks while also ensuring their interdependence and unique specialising manipulation type of task. All derived tasks in this work are commonly seen in robotic manipulation and physics-based interactions \cite{Billardeaat8414}. With a total of seven experts in the hierarchical architecture of ROMAN as seen in \autoref{fig:CoverImage}, a corresponding total of seven sequential activation cases were derived, which are hereinafter referred to as \textit{\textbf{scenarios}}. Thereafter, the numbering of scenarios also indicates how many experts are involved in itself, as each sequence builds upon the previous by adding a new task to the sequence and granting the task all the more complex of longer horizons. Lastly, the episode in the simulation would terminate either (i) once the button next to the conveyor belt would be pushed, (ii) the maximum step count for the episode would be reached or (iii) the robot end-effector would spatially deviate too far from the centre of the scene.

\subsection*{Expert Network Characteristics and Architecture}
During the derivation of the full sequential task, we attempted to derive as many fundamental manipulation primitives in 3D space as possible while also being inspired by the most widely used daily manipulation tasks, henceforth referred to as \textit{\textbf{experts}}. This, in turn, allowed for the validation of the robustness of the architecture against increased complexity, uncertainty and dimensionality. The manipulation experts are derived with diverse and distinct specialised skills to cover a broad range of common tasks in real-life cases and robotic manipulation \cite{9492850, Ortenzi2019}. During the derivation of these experts, it was also deemed important to not derive experts that are too closely interrelated to one another, as to cover the diverse types of tasks seen in daily life and thereby offering greater versatility and flexibility while used in combination. The total number of expert NNs that were trained is seven. These are visually depicted in \autoref{fig:CoverImage} and are listed in detail below:
\begin{itemize}
  \item \textbf{Pull-Opening (Opens Drawer) [$\pi_{Pull}$]:} An expert responsible for pulling a linearly moving object, such as a sliding drawer across an axis.
  \item \textbf{Picking and Dropping (Unboxes) [$\pi_{PickDrop}$]:} An expert responsible for picking and dropping an object without regard to a height offset when placing, hence dropping an item. This is commonly seen when removing the lid or the cover of a disposable box to retrieve an object of interest.
  \item \textbf{Rotating Opening (Opens Cabinet) [$\pi_{RotateOpen}$]:} An expert responsible for rotating a door handle configured around a single axis, a very common scenario seen when opening a cabinet, door or rotating drawer.
  \item \textbf{Picking and Placing (Places Rack) [$\pi_{PickPlace}$]:} An expert responsible for picking and placing an object carefully on a target surface, with zero or close to minimal height drop.
  \item \textbf{Picking and Inserting (Inserts Vial) [$\pi_{PickInsert}$]:} An expert responsible for picking and inserting an object with significantly high levels of precision to a respective docking target location.
  \item \textbf{Pushing (Pushes Rack and Vial) [$\pi_{Push}$]:} An expert responsible for pushing an object across a surface.
  \item \textbf{Pushing-Button (Pushes Button) [$\pi_{Button}$]:} An expert responsible for pushing and activating a human-made switch or button.
\end{itemize}

\paragraph*{Action Space --} All aforementioned experts are listed in the form of high-level types of abstract manipulation tasks (specific task on validated environment). All experts shared identical actions. These actions including full end-effector velocities in three dimensions ($\alpha_{1}:\pm v_{x}$, $\alpha_{2}:\pm v_{y}$, $\alpha_{3}:\pm v_{z}$), as well as controlling the gripper state in a binary approach ($\alpha_{4}:f({\pm x_{g}}$)). Sharing an identical action space across the incorporated experts in the hierarchical architecture of ROMAN is relevant to highlight the value of the proposed hierarchical framework, as expert specialisation is not aided by having constrained the actions available to each expert to those that are only relevant for their respective specialisation.

\paragraph*{State Space --} Foremost, the state space of each expert was identical for the proprioceptive and sensory states. However, the exteroceptive states differed from expert to expert, dependent on each expert's individual specialised type of manipulation skill. Consequently, only the relevant information from the environment for the successful completion of each individual task was made available in the exteroceptive state for each expert. Hence, the exteroceptive states were decided based on the nature of each expert's specialised skill and end goal. This subsequently allowed each expert NN to only focus on its own core exteroceptive information relevant to its sub-task, and omit non-relevant ones. This decision was inspired and is seen from a neuro-scientific perspective whereby during the human decision-making process, the relevance of information during a motor task is determined and specified \cite{Wolpert2011}. The full state and action space, including the demonstration settings, training times and specific details of these for each expert and the MN are detailed in \autoref{table:DetailedNetworkCharacteristicsStructure}. 

\paragraph*{Focus on High-Level Task Decomposition --} The derived experts were composed in such a way as to allow the decomposition of high-level tasks, thereby off-loading the central MN of ROMAN from combining a large number of low-level action-based experts that can otherwise be solved by a single sub-task-based expert. This task decomposition was made possible by virtue of the employed hybrid-learning procedure in the hierarchical architecture of ROMAN, which incorporates and orchestrates multiple NNs specialising in sub-tasks to efficiently and effectively solve complex long horizon sequences. 

In contrast to the high-level task decomposition employed in this work, most related work decomposed manipulation experts into rather basic action-based primitives or action-level skills \cite{8461249, 9659344}. While the employed low-level task decomposition in the related work allows for the derivation of more abstract cases, it does limit the potential of a hierarchical model as with ROMAN's. In particular, a decomposition of low-level action-based skills limits, to a great extent, the gating network potential from learning high-level scene understandings or solving complex sequences as it focuses more on composing skills such as \textit{Picking and Placing} which can be instead solved by one single expert. For instance, in the work of \cite{9659344}, the skill of \textit{Picking and Placing} was learned using a three-expert hierarchical architecture composed of (i) \textit{Approaching}, (ii) \textit{Manipulating} and (iii) \textit{Retracting}. ROMAN's framework shows that by virtue of the employed hybrid learning procedure which balances exploration and exploitation, the derivation of \textit{Picking and Placing} as a single high-level expert is made possible. This is how the employed HHL architecture of ROMAN overcomes such limitations by deriving experts specialised in high-level sub-task-based manipulation skills, offloading the MN in turn from lower-level skill supervision.

Furthermore, each derived single task-level expert was trained via the same hybrid learning procedure. This translates, as per the evaluation and the exhibited recovery capabilities in the Results section, an inherent individual expert robustness in facing new states during the exploration of the RL process. This allowed the MN to be trained more effectively in solving highly complex sequences over long horizons, without the need of learning how to re-combine primitive action-based experts to achieve a sub-task.

From preliminary investigations and results, it was observed that when the MN was switching between the different experts, there was the possibility of dropping the object when suddenly switching from any expert involved with \textit{Picking} to another. This was due to the limited control interface of the incorporated gripper in the simulation, providing only binary commands for opening and closing. To compensate for this, a \textit{Dead Zone (DZ)} implementation was introduced to account for the expert switching process. This relationship is shown below as:
\begin{equation}
\text{DZ}(x_{g})=\begin{cases}
         \text{close} \quad &\text{if } \, x_{g} \in {[-1.0, -0.9]}, \\
         \text{remain the same} \quad &\text{if } \, x_{g} \in {(-0.9 , 0.9)}, \\
         \text{open} \quad &\text{if } \, x_{g} \in {[0.9, 1.0]}. \\
     \end{cases}
\end{equation}
Hence, the DZ ($\in (0,1)$) implementation improved the overall stability of grasping, by only switching the gripper action to open or close when \(x_{g}\) goes beyond the zone of $(-0.9, 0.9)$, effectively ignoring the remainder. As a possible future work, the DZ implementation could potentially be substituted by incorporating a dynamic controller with force control or tactile sensing to render grasping more reliable.

\subsection*{Manipulation Network Characteristics}
The MN acts as a master control policy, overseeing and supervising the incorporated expert NNs and assigning weights ($\in (0,1)$) to them. The final output is defined as the sum of those weighted actions:
\begin{equation}
\label{equation:blendingActionsWeights}
\begin{gathered}
     \sum_{i=1}^{m}\sum_{j=1}^{n} \alpha_{i} \cdot w_{j}
\end{gathered}
\end{equation}
whereby the number ($m=4$) of all actions ($\alpha_{i}$) of each expert is controlled by a set of weights ($w_{j}$) corresponding to the total number ($n=7$) of experts in the hierarchy. One of the main limitations and issues arising in assigning the weights is the potential of the sum of all weights exceeding one, which can lead to unwanted behaviour, e.g. most notably torques and forces going beyond the robot's capabilities. Consequently, it was deemed crucial to ensure that the sum of weights does not exceed one.

To account for the above, we normalise the sum of weights assigned by the MN to activate experts, using a normalised exponential function, i.e., softmax. The softmax in turn provides us with a probability distribution to better isolate the expert activations during long sequences. This is represented as:
\begin{equation}
\label{equation:softmax}
\begin{gathered}
     \sigma (\mathbf {z} )_{i}={\frac {e^{z_{i}}}{\sum _{j=1}^{K}e^{z_{j}}}}\ \ \ \ {\text{ for }}i=1,\dotsc ,K{\text{ and }}\mathbf {z} =(z_{1},\dotsc ,z_{K})\in \mathbb {R} ^{K}.
\end{gathered}
\end{equation}
where $\sigma$ is the softmax function, $z$ is the input vector and represented as a function of $e^{z_{i}}$ denoting the standard exponential for each input, divided by the sum of all inputs $K$. In our case, the input vector is represented as a weight vector, with each element representing the weight of every expert in the hierarchical architecture, with a sum equal to $K=7$.

Contrary to the state space of the expert NNs being dependent on their individual studied nature of tasks and sub-task goals, the observation space of the MN contains the union of all of the observation spaces of the incorporated experts. Consequently, the MN, in essence, observes the entirety of the relevant sub-tasks allowing for a better distinction as to which expert should be activated and at which time step. \autoref{figure:NN_MN_StructureOverview} depicts the overall ROMAN framework and in particular highlighting the MN as a gating mechanism which centrally governs the control policy in the HHL control framework.

\section*{Data availability}
The evaluation data stemming from the experiments are made publicly available and can be accessed at \url{https://github.com/etriantafyllidis/ROMAN_Data}. The data can be downloaded as a compressed file (.zip) and consist of Comma-Separated Values (CSV) formatted files for each evaluated scenario, including success rates for sub-tasks and end-goal sequences. A ReadMe file is included for context and for further details on the data format.

\section*{Code availability}
ROMAN's code is made available at \url{https://github.com/etriantafyllidis/ROMAN} \cite{eleftherios_triantafyllidis_2023_8059565}.

\section*{Acknowledgements}
We would like to thank Prof. Taku Komura, Prof. Robert Fisher, Dr. Quentin Rouxel, and Dr. Filippos Christianos for their valuable feedback. We thank Dr. Ruoshi Wen for providing certain 3D models used in this work. We would also like to thank Dr. Valentina Andries for proofreading contents of this work. In regards to funding, E.T. is supported by the EPSRC CDT in Robotics and Autonomous Systems (EP/L016834/1), and F.A. is supported by the UKRI CDT in Foundational Artificial Intelligence (EP/S021566/1).

\section*{Author Contributions Statement}
E.T. contributed to the conceptualisation and implementation of the ROMAN architecture, the simulation setup, data acquisition and analysis, experimentation, designing the visuals and figures, and authored the manuscript. F.A. contributed to the derivation of the ROMAN architecture, data analysis, t-SNE analysis, types of experiments conducted, visuals and figures, and writing of the manuscript. Z.Liu contributed to the ROMAN architecture, data analysis, types of experiments conducted, visuals and figures, and writing of the manuscript. Z.Li directed the research, provided management across all aspects of this research, supported solving research and technical problems, edited figures, and wrote the manuscript. All authors contributed to the manuscript.

\section*{Competing Interests Statement}
The authors declare no competing interests.

\bibliography{bibliography}
\bibliographystyle{naturemag}

\newcommand{\beginsupplement}{%
	\setcounter{table}{0}
	\renewcommand{\tablename}{Supplementary Table}
	\setcounter{figure}{0}
	\renewcommand{\figurename}{Supplementary Figure}
}

\newpage
\vspace*{\fill}
\begin{center}
{\Large Supplementary Materials for \\ \vspace{-0.2cm} \vspace{0.25cm} \textbf{RObotic MAnipulation Network (ROMAN)} -- \\ \vspace{0.25cm}  Hybrid Hierarchical Learning for Solving Complex Sequential Tasks}
\end{center}

\captionsetup[figure]{name=Supplementary Figure}
\captionsetup[table]{name=Supplementary Table}
\renewcommand{\figureautorefname}{Supplementary Figure}
\renewcommand{\tableautorefname}{Supplementary Table}
\setcounter{figure}{0}
\setcounter{table}{0}

\vspace{1cm} 

\textbf{\Large Supplementary Table of Contents:}
\begin{itemize}
 \setlength{\itemsep}{0.0pt}%
    \item \hyperref[SN1]{Supplementary Note 1}: System Overview and Apparatus
    \item \hyperref[SN2]{Supplementary Note 2}: Contributions and Key Summary Points of ROMAN 
    \item \hyperref[SN3]{Supplementary Note 3}: Figures and Tables - Detailed Expansion
    \item \hyperref[SN4]{Supplementary Note 4}: Expanding Upon the Results In Detail
    \item \hyperref[SN5]{Supplementary Note 5}: Simulation Environment and Control Framework
    \item \hyperref[SN6]{Supplementary Note 6}: ROMAN's Hybrid Learning and Architecture
    \item \hyperref[SN7]{Supplementary Note 7}: The Architectures and Training of Monolithic Networks
    \item 
    \hyperref[alg:training_procedure]{Supplementary Algorithm 1}: Algorithmic Depiction of the Hybrid-Learning Procedure within ROMAN
    \item \hyperref[sup_fig:trainingFlowDiagram]{Supplementary Figure 1}: Flow chart of the hybrid training procedure.
    \item \hyperref[sup_fig:rewardPlotIndividualExperts]{Supplementary Figure 2}: Training plot of each individual expert that depicts the normalised reward over the environment steps in millions.
    \item \hyperref[sup_fig:rewardPlotDemos]{Supplementary Figure 3}: Training plot of a different number of demonstrations on the ROMAN framework, with the normalised reward over the environment steps in millions.
    \item \hyperref[sup_fig:rewardPlotSingleNNvsROMAN]{Supplementary Figure 4}: Training plot of the hierarchical framework: ROMAN versus a single NN in full 3D space, with the overall normalised reward over the environment steps in tens of millions.
    \item \hyperref[sup_fig:demonstrationActionVariance]{Supplementary Figure 5}: Variation of actions in the demonstration dataset of N=21 and N=42 for scenario case S5.
    \item \hyperref[table_sup::TasksWithGoalsDependency]{Supplementary Table 1}: Derived and investigated manipulation tasks and their goals and interdependencies.
    \item \hyperref[table_sup:DetailROMANDifferentLevelsNoise]{Supplementary Table 2}: Detailed results of ROMAN expanding upon the Main Manuscript Table 1.b, where the overall success of each sequential case scenario is depicted over different levels of uncertainty, with the individual expert success highlighted.
    \item \hyperref[table_sup:2DDetailedSingleNNvsROMAN]{Supplementary Table 3}: Detailed results of a single NN vs the preliminary stage of the hierarchical framework ROMAN in 2D, expanding upon the Main Manuscript Table 2.a.
    \item \hyperref[table_sup:3DDetailedSingleNNvsROMAN]{Supplementary Table 4}: Detailed results of a single NN vs the hierarchical framework ROMAN, expanding upon the Main Manuscript Table 2.b.
    \item \hyperref[table_sup:DetailedAlgorithmROMAN_0_5cm]{Supplementary Table 5}: Detailed results of ROMAN expanding upon the Main Manuscript Table 2.c, where the overall success of each sequential case scenario is depicted with the use of different learning algorithm paradigms at a $\sigma = \pm 0.5 \text{ [cm]}$ level of Gaussian noise.
    \item \hyperref[table_sup:DetailedAlgorithmROMAN_1_and_2cm]{Supplementary Table 6}: Detailed results of ROMAN expanding upon the Main Manuscript Table 2.c, where the overall success of each sequential case scenario is depicted with the use of different learning algorithm paradigms at $\sigma = \pm 1.0 \text{ [cm]}$ and $\sigma = \pm 2.0 \text{ [cm]}$ levels of Gaussian noise.
    \item \hyperref[table_sup:DetailedDemonstrationsROMAN]{Supplementary Table 7}: Detailed results of ROMAN expanding upon Manuscript.Table.2d, where the overall success of each sequential case scenario is depicted over the different number of demonstrations, with the individual expert success also highlighted.
    \item \hyperref[table_sup:Duration_Experts_of_ROMAN_on_Noise_Levels]{Supplementary Table 8}: Duration of each expert within ROMAN completing their individual sub-task.
    \item \hyperref[table_sup:Duration_Sequences_of_ROMAN_on_Noise_Levels]{Supplementary Table 9}: Duration of the full sequential tasks by ROMAN and all included experts within those.
    \item \hyperref[table_sup:StandardDevRewardDemos]{Supplementary Table 10}: Differences in the standard deviation of the rewards between N=21 and N=42.
    \item \hyperref[table_sup:ROMAN_2D_DetailedNetworkCharacteristicsStructure]{Supplementary Table 11}: Architecture details of ROMAN in its preliminary 2D setting composed of five experts.
    \item \hyperref[table_sup:roman_nn_architectures]{Supplementary Table 12}: Neural network architectures of all incorporated NNs in the hierarchical formation of ROMAN.
    \item \hyperref[table_sup:Hyperparameters]{Supplementary Table 13}: Hyperparameters for all experts in the ROMAN framework, including the gating network.
    \item \hyperref[table_sup:SingleNN_2D_3D]{Supplementary Table 14}: Single NN architecture settings for both the 2D and 3D cases.
    \item \hyperref[table_sup:SingleNN_2D_3D_Hyperparameters]{Supplementary Table 15}: Hyperparameters for the single NNs for both the 2D and 3D cases.
\end{itemize}

\newpage
\section*{Supplementary Notes}
In the supplementary materials, additional technical details of the ROMAN framework are provided. These include the different hyperparameter settings used for training, details of the raw data collected during experiments, and supplementary analyses and results that complement the main manuscript. Furthermore, a detailed description of the individual neural network architectures, state and action representations, dimension and hyperparameters used for the single NNs in both 2D and 3D manipulation tasks is listed. Data acquired in this work did not involve human participants and consequently, the submitted data does not contain sensitive information subject to data anonymisation.

\subsection*{Supplementary Note S1: System Overview and Apparatus}
\label{SN1}
The derivation of ROMAN and its subsequent evaluation was made possible via the use of the simulation engine Unity3D (2019.3.0f6), containing the built-in NVIDIA physics engine (PhysX 4.1). Moreover, to allow for training, the simulation was coupled with the PyTorch-based ML-Agents toolkit \footnote{\urlstyle{same}\url{https://github.com/Unity-Technologies/ml-agents}} \cite{juliani2018unity}. With the aforementioned simulation and tools, a realistic simulation environment was achieved for the purpose of training and validating robotic manipulation tasks for autonomous operation. To furthermore ensure reliable physical modelling of the robotic system, we incorporated the Robotics Operating System (ROS) ROS\# plugin to import physics models of robots and objects (Unified Robot Description Format) into the Unity3D engine. The physics simulation frequency was set at $1000\,$Hz. This allowed us to ensure robust and stable physics performance entailing realistic physical properties and frictions in the simulation. 

The robotic system used for the experiments in the simulation environment was the Franka Panda Emika robot, with its default gripper configuration. All simulated tasks were within the robot's operational workspace. All control policies of the expert NNs, from this point onward, operated in a direct velocity control of the gripper end-effector gripper as well as the state of activating i.e. opening and closing the gripper. An Inverse Kinematics (IK) solver was used, for visualisation purposes, to solve the kinematic chain to the end-effector position. The IK solver and the control of the robot arm operated at $1000\,$Hz and a low-pass filter was introduced for filtering and de-noising the exteroceptive states (i.e. the observation signals) to the NNs. The full training processes were conducted on a desktop computer Central Processing Unit (CPU), incorporating an 18-core Intel i9-9980XE. The parallelism in the training procedure of the robot allowed the use of concurrent training instances, significantly speeding up wall-clock training time.

\subsection*{Supplementary Note S2: Contributions and Key Summary Points of ROMAN}
\label{SN2}
\begin{itemize}
    \item The proposal of ROMAN, an event-based Hybrid Hierarchical Learning (HHL) framework for hierarchical task learning, leveraging behavioural cloning, imitation learning (GAIL) and reinforcement learning for solving multiple sequences of robotic manipulation tasks;
    \item The hybrid learning approach and the higher-level decomposition of tasks enable ROMAN's incorporated experts to match and surpass the performance of equivalent hierarchies in the related work;
    \item Comprehensive quantitative evaluation of ROMAN's framework against a multitude of different settings, in particular: (i) multiple levels of increasing exteroceptive state noise, (ii) increasing levels of long-time horizon sequential tasks, (iii) different number of demonstrations, (iv) comparison against monolithic neural networks, and (v) a thorough ablation study investigating the effects of different learning paradigms on the hierarchy and its incorporated experts.
    \item A novel use of a Mixture of Experts (MoE) architecture for robotic manipulation, whereby the gating network, referred to as the Manipulation Network (MN), is able to solve complex, long-horizon sequential end goals by (i) exhibiting robustness to different expert activation sequences and their orchestration, (ii) higher-level scene understandings even under increasingly exteroceptive uncertainties as well as (iii) exhibiting adaptation and recovery capabilities to cases not encountered in the demonstrated sequence as the result of the employed hybrid learning procedure which in turn balances exploitation and exploration beyond imitation. 
\end{itemize}

\subsection*{Supplementary Note S3: Figures and Tables - Detailed Expansion}
\label{SN3}
In this section, additional details and analyses of the results are presented. In particular, contrary to the main manuscript, in these supplementary materials all sub-task sequences and individual expert success as well as completion times are included during the full sequence of actions.

First and foremost, the sequential manipulation tasks and their interdependence that were derived in the studied experiments to evaluate ROMAN against are detailed in \autoref{table_sup::TasksWithGoalsDependency}. The overall success rates of ROMAN based on different Gaussian levels of exteroceptive noise are detailed in \autoref{table_sup:DetailROMANDifferentLevelsNoise}, expanding upon the Main Manuscript Table 1.b. The comparison of the hierarchical formation of ROMAN compared against monolithic neural network equivalents for 2D and 3D sharing an identical hybrid learning procedure as to retain consistency and conduct a fair comparison, are detailed in \autoref{table_sup:2DDetailedSingleNNvsROMAN} and \autoref{table_sup:3DDetailedSingleNNvsROMAN}. These two tables expand upon Main Manuscript Table 2.a and Main Manuscript Table 2.b respectively. 

The comparison of different learning paradigms during the ablation study detailed in the main manuscript (RL, BC and GAIL), including their different combinations within ROMAN against noise levels equivalent to $\pm 0.5$ cm are detailed in \autoref{table_sup:DetailedAlgorithmROMAN_0_5cm}. For increasing thereafter noise levels of $\pm 1.0$ cm and $\pm 2.0$ cm, details can be found in \autoref{table_sup:DetailedAlgorithmROMAN_1_and_2cm}. \autoref{table_sup:DetailedAlgorithmROMAN_0_5cm} and \autoref{table_sup:DetailedAlgorithmROMAN_1_and_2cm} expand upon the Main Manuscript Table 2.c. Lastly, the number of different demonstrations provided to ROMAN's NNs and by extent to its learning paradigms where relevant, i.e. for BC and GAIL, are outlined \autoref{table_sup:DetailedDemonstrationsROMAN}, expanding upon Main Manuscript Table 2.d. The duration in seconds to complete all case scenarios by ROMAN against different levels of noise ranging from no noise $\pm 0.0$ cm to $\pm 2.5$ cm, in $0.5$ cm increments, are all outlined in \autoref{table_sup:Duration_Experts_of_ROMAN_on_Noise_Levels} for each individual expert and in \autoref{table_sup:Duration_Sequences_of_ROMAN_on_Noise_Levels} for the full sequential scenario tasks respectively.

The analyses of these supplementary data provided us with a thorough and comprehensive understanding of not only the overall sequential success rates but also the entailed individual sub-tasks as part of the full sequential task decomposition. This allowed, in turn, for a thorough investigation as to whether sequential failures were attributed to either specific expert NNs within the sequence of actions, or higher-level errors stemming from the orchestration of weight assignments by the MN. 

In \autoref{sup_fig:rewardPlotIndividualExperts} as well as \autoref{sup_fig:rewardPlotDemos}, the maximum attainable (normalised) reward plots during training for each individual expert within ROMAN are shown and in particular how the different number of demonstrations  ($N=7$, $N=21$ and $N=42$) affected the training procedure respectively. Finally, in \autoref{sup_fig:rewardPlotSingleNNvsROMAN}, the normalised reward plot of a single NN compared against ROMAN in full 3D space is depicted. 

From the supplementary tables as well as figures, and in line with the inferred result discussions from the main manuscript, it was observed that \textit{Picking and Dropping, Placing and Inserting}, were arguably the most complex experts/sub-tasks due to their higher complexity compared to the other experts in the ROMAN framework. In particular, this was observed in three main aspects, more specifically: (i) requiring overall longer training duration, (ii) requiring higher completion times, and (iii) exhibiting overall lower success rates than other expert NNs within the ROMAN hierarchical architecture.

\subsection*{Supplementary Note S4: Expanding Upon the Results in Detail}
\label{SN4}
In this section, additional details on select results from ROMAN are provided, that were not deemed critical for the primary understanding and implications of the hierarchical framework and were therefore omitted from the main manuscript text.

\begin{figure}[H]
	\begin{center}
		\includegraphics[width=0.67\textwidth]{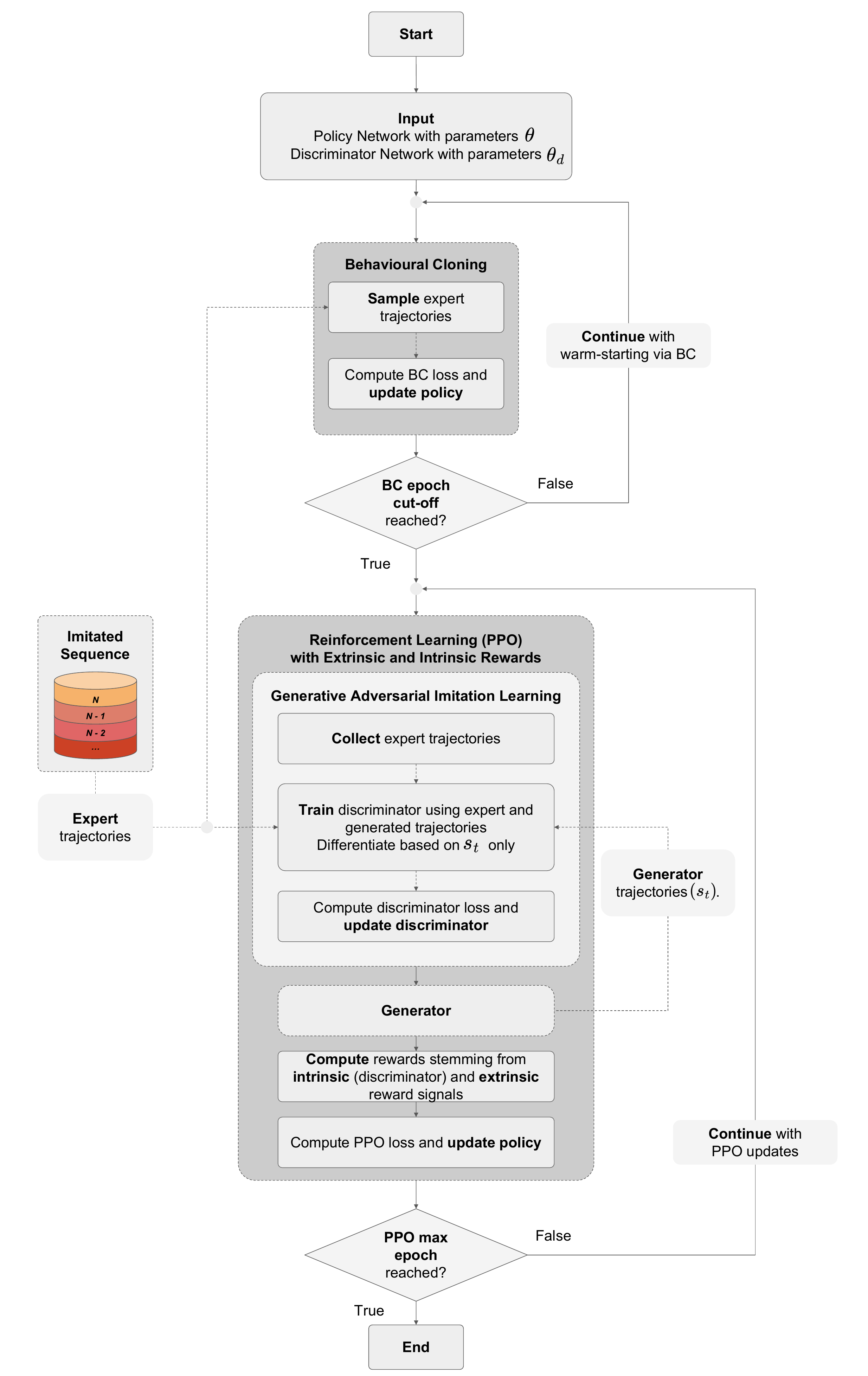}
	\end{center}
	\caption{\textbf{The flow chart of the hybrid training procedure.} The depiction of the main stages of the training procedure, including the use of demonstrations to warm-start the policy via behavioural cloning (BC). Thereafter the policy is updated following the use of PPO, primarily acting as the general purpose update rule, with extrinsic ($r_E$) and intrinsic ($r_I$) rewards provided by the environment and GAIL's discriminator respectively. This training procedure is employed for all expert NNs incorporated in ROMAN's hierarchical framework. Given the pre-trained expert NNs, the MN is subsequently trained with the same hybrid learning procedure.}
	\label{sup_fig:trainingFlowDiagram}
\end{figure}

\begin{figure}[H]
	\begin{center}
		\includegraphics[width=0.8\textwidth]{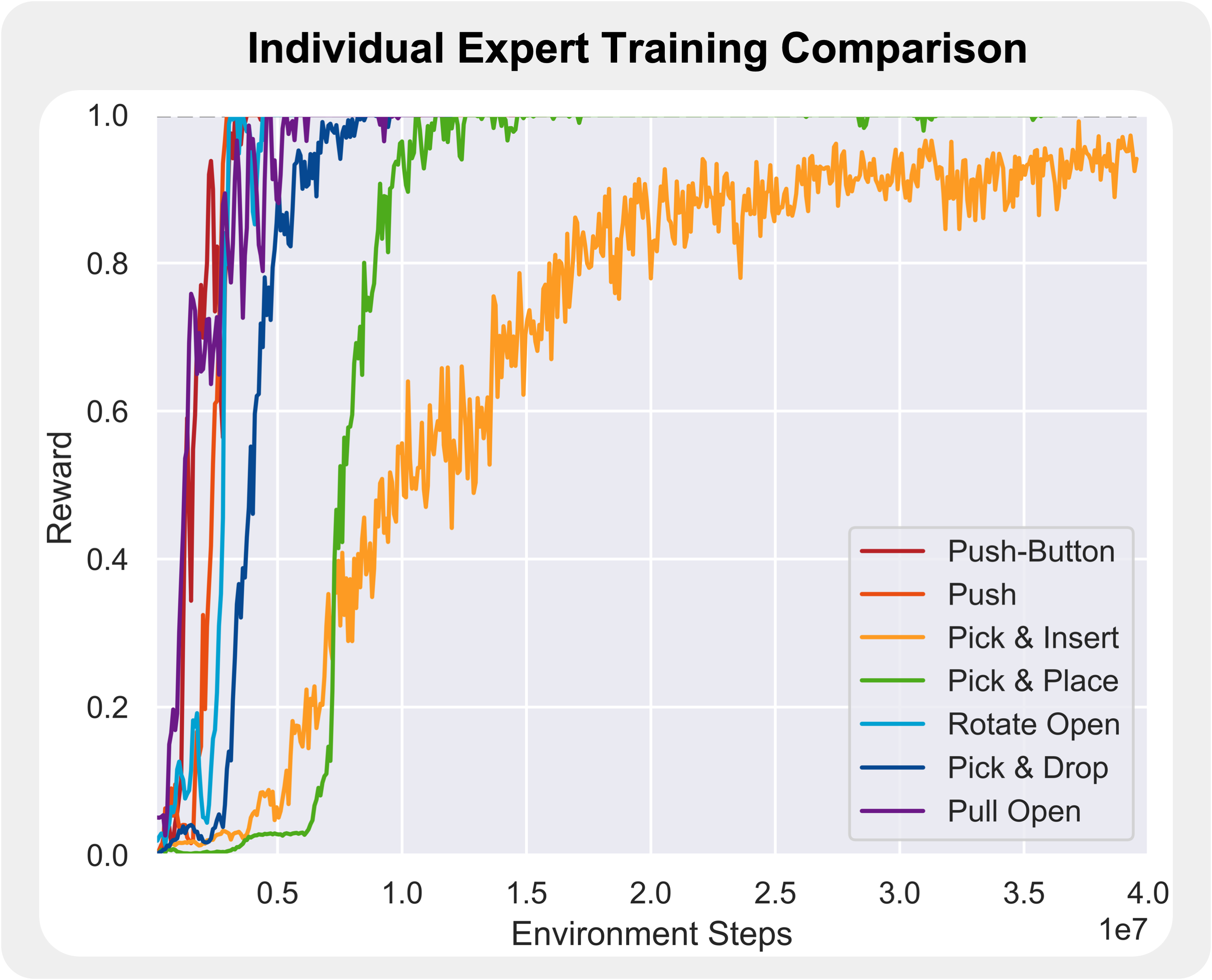}
	\end{center}
	\caption{\textbf{Training plot of each individual expert depicting the normalised reward over the environment steps in millions.} The figure shows the different training steps of each expert in the ROMAN framework with the returns over the duration of the training steps. Notice that the training requirements in environment steps depend on the nature and complexity of each specialising expert. The most apparent observation is that every expert concerned with a higher-level complexity goal, such as those concerned with \textit{Picking \& Dropping, Placing or Inserting}, were admittedly the most complex and longest in time horizons compared to other experts. As discussed in detail the main manuscript, developing task-specific experts allowed for reducing the subsequent burden on the primarily gating network. This is because rather than learning to schedule low-level sub-tasks, the gating network can focus entirely on orchestrating the higher-level tasks using specialised experts. This approach minimises the amount of unnecessary information that the gating network needs to process during the sequential supervision and orchestration of the included experts, ultimately resulting in a more efficient and effective task execution. As observed in the reward plot, the highest complexity was undoubtedly presented with the \textit{Picking and Inserting} expert, requiring the most training in environment steps compared to other experts. This is furthermore evidenced by the qualitative difficulty of obtaining the demonstration data from a human expert which was also the most demanding in regards to effort in this specific sub-task. All experts depicted and used in ROMAN were pre-trained with $N=20$ demonstrations.}
	\label{sup_fig:rewardPlotIndividualExperts}
\end{figure}

\begin{figure}[H]
	\begin{center}
		\includegraphics[width=0.8\textwidth]{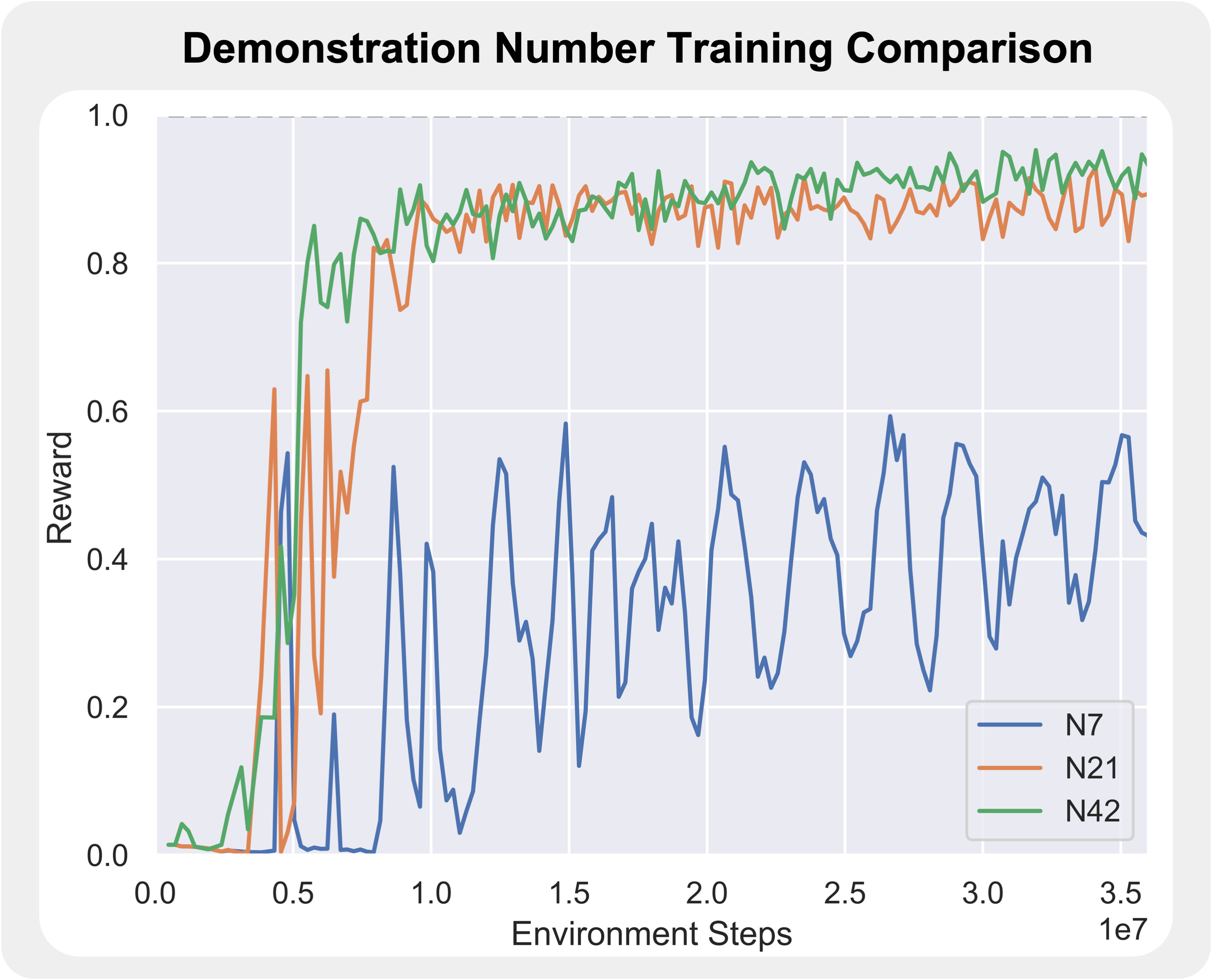}
	\end{center}
	\caption{\textbf{The training plot of the different number of demonstrations employed in the ROMAN framework, with the normalised reward over the environment steps in millions.} The figure depicts the different number of demonstrations acquired $N=7$, $N=21$ and $N=42$, thereafter used for the gating network of ROMAN and how it affected training. As described in the manuscript, a total of $N=7$ demonstrations, corresponded essentially to one demonstration for each of the 7 different sequential tasks. Consequently, $N=21$ and $N=42$ correspond to 3 and 6 demos per sequential case scenario respectively. After further analysing the above reward plot and testing ROMAN on these different numbers of demonstrations, we concluded that $N=7$ was insufficient for even a minimal level of acceptable performance of the gating network; rather, a minimum of $N=21$ demonstrations were necessary to attain high success rates, which in turn did not differ by much when doubling the demonstrations to $N=42$. This suggests that a comparable number of demonstrations provided to the MN as with ROMAN's experts, can render high success rates, primarily attributed to the employed hybrid learning procedure. In the above cases, a maximum environment step of 3.5 million was set after which we terminated the training, as depicted in the figure.}
	\label{sup_fig:rewardPlotDemos}
\end{figure}

\begin{figure}[H]
	\begin{center}
		\includegraphics[width=0.8\textwidth]{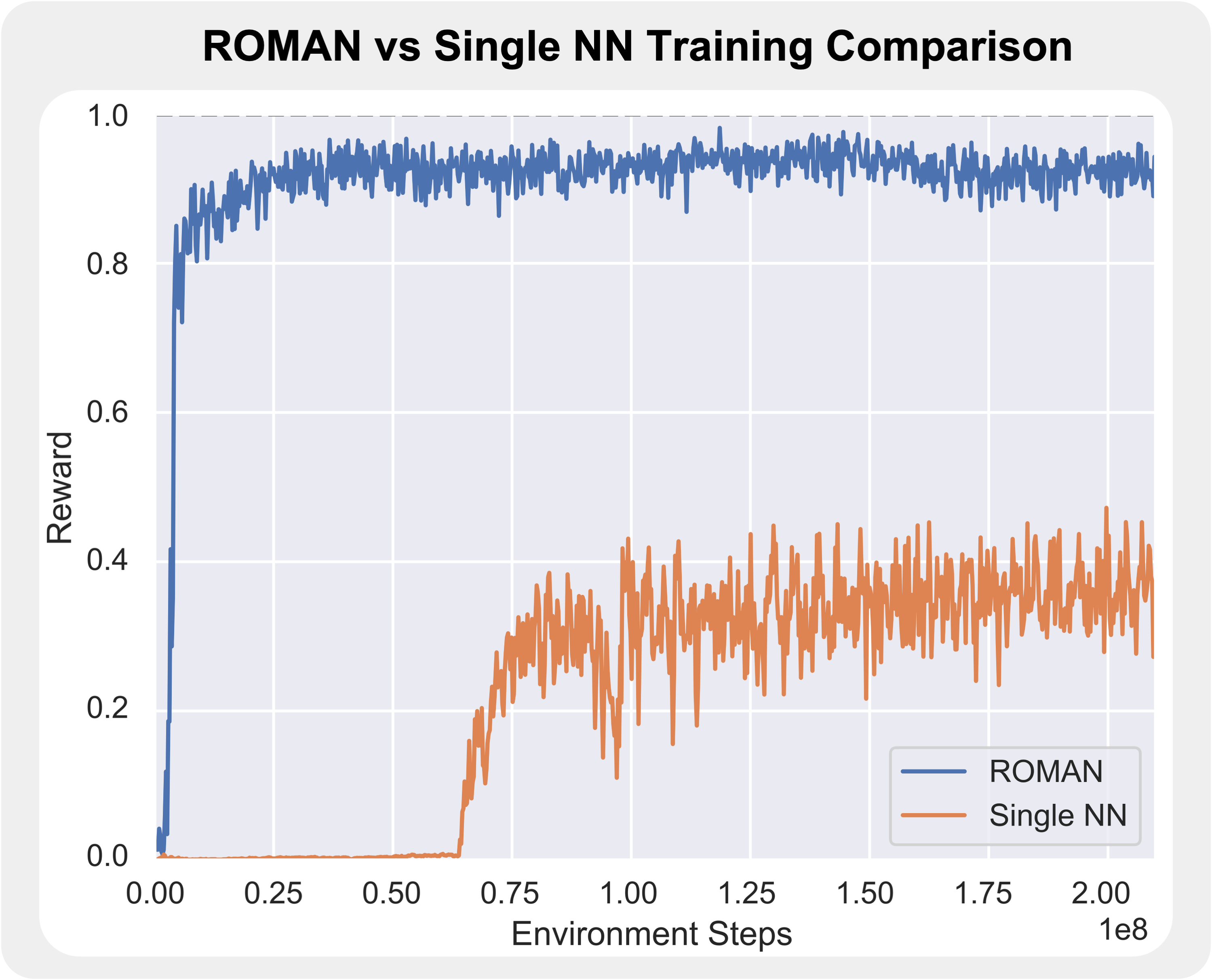}
	\end{center}
	\caption{\textbf{The training plot of the hierarchical framework of ROMAN compared against a monolithic NN in full 3D space, with the overall normalised reward over the environment steps in tens of millions.} The reward plot depicts the training of ROMAN vs a single NN, both being trained in full 3D space. As it can be observed only ROMAN's hierarchical architecture is able to attain, to the closest possible extent, the maximum attainable reward. ROMAN's MN was given a total of $N=42$ demonstrations, corresponding to $N=6$ demonstrations for each of the seven sequential case scenarios. The single NN in 3D was given $N=140$ demonstrations. This decision was made to conduct the fairest possible comparison by accounting for ROMAN's 7 pre-trained experts with $N=20$ demonstrations each, thus giving the monolithic NN an equivalent number of demonstrations to the relative hierarchical architecture compared against. It is noteworthy to point out that both ROMAN's NNs in its hierarchical framework as well as the single NN were trained with an identical hybrid learning procedure to retain consistency. The reward plot suggests that a hierarchical formation is of necessity for solving and attaining robust performance in complex sequential and long-time horizon tasks as studied in this work. From the above plot and the details laid out in the main manuscript, it is inferred that a monolithic NN, even though trained with the same hyperparameters and overall hybrid learning procedure, is unable to attain robust performance, especially amongst the most difficult sequential tasks with increasing time horizons. In conclusion, this highlights the value of using a hierarchical task decomposition, in order to solve complex long-horizon sequential tasks as commonly encountered in robotic manipulation.}
	\label{sup_fig:rewardPlotSingleNNvsROMAN}
\end{figure}

\paragraph*{\textbf{Result Implications on a Monolithic NN vs ROMAN's Hierarchical Architecture in 2D: }} Prior to scaling ROMAN to the more complex 3D Euclidean space, a preliminary benchmark was conducted as to evaluate the performance of a monolithic approach versus ROMAN's initial version based on 2D settings entailing a total of five experts. By sharing an identical hybrid learning procedure, this evaluation allowed a direct comparison of a monolithic versus a hierarchical approach. This in turn allowed us to evaluate and demonstrate the advantages of a hierarchical task decomposition. The hybrid learning procedure similarly to ROMAN's framework, shown in \autoref{sup_fig:trainingFlowDiagram}, was used to train the single NN. Furthermore, identical states and hyperparameters were used for both ROMAN's MN and the single NN. In particular, the action space of the single NN was identical to ROMAN's experts for controlling the end-effector and the gripper. In regards to the demonstrations, a total number of $N=100$ demonstrations were provided to the single NN to match ROMAN's 2D setting composed of 5 experts pre-trained with $N=20$ (5 experts $\times$ 20 demos each). 

From \autoref{table_sup:2DDetailedSingleNNvsROMAN} it is inferred that while a single NN can accommodate the least complex and with shorter time-horizon sequential scenario cases 1 and 2 well, cases 3, 4 and 5 exhibit significantly lower success rates (less than $\approx $70\%), dropping to a low of $\approx $56\%, even in a fairly simplistic 2D case setting. More specifically, S3 introduces the \textbf{Pick and Insert} expert, followed by \textbf{Pick and Drop} in S4 which are arguably the most demanding ones, as they in essence combine three low-level sub-tasks composed of (i) \textit{reaching}, (ii) \textit{grasping}, (iii) \textit{inserting}. Overall, it is observed that from the results a single NN has significantly lower success rates due to the complexities associated with longer horizons. This, in turn, underlines the value of ROMAN's hierarchical framework for complex and long in-sequence tasks and the necessity of a hierarchical architecture.

\paragraph*{\textbf{Result Implications on a Monolithic NN vs ROMAN in 3D: }} Similarly to the 2D case of ROMAN outlined above, a further investigation was conducted as to whether a single NN can solve sequences requiring up to seven experts in 3D space. To retain consistency as before and similarly to the baseline evaluation of the monolithic NN versus ROMAN in the 2D case setting, an identical training procedure via the hybrid learning approach (see \autoref{sup_fig:trainingFlowDiagram}) was used for the 3D single NN. Identical states (to ROMAN's MN), actions (to ROMAN's experts) and hyperparameters were used for the monolithic NN. For the 3D case, a total of $N=140$ demonstrations were employed and provided to the single NN to match ROMAN's seven experts, each pre-trained with a total of $N=20$ demonstrations (7  experts $\times$ 20 demos).

In line with \autoref{table_sup:2DDetailedSingleNNvsROMAN}, results in \autoref{table_sup:3DDetailedSingleNNvsROMAN} suggest that a single NN is unable to solve the complex nature and long sequential task of the validated manipulation scenario given an identical training procedure. Although a monolithic NN approach exhibits some robustness against cases S1 to S3 in 2D space, it cannot robustly attain stable performance for scenarios S4 to S7 in 3D space due to their inherently more complex and longer horizon sequences. Despite S1 and S2 initially showing promising success rates, the monolithic NN is unable to converge to a stable performance with longer and more complex sequences, achieving only 58.3\% success in S3. Subsequent evaluation found that the monolithic NN exhibits less than 3\% success rates for S4 and S5, and completely failing for S6 and S7. Extending the noise level comparison beyond $\sigma = \pm 0.5$ cm was disregarded due to the already significantly lower success rates of the monolithic NN at such noise levels. 

It is hence inferred that in 2D and especially in 3D space, a monolithic NN, even though trained with an identical hybrid learning procedure as with ROMAN's incorporated NNs, is unable to solve the complex, long-time-horizon sequential tasks studied. This observation underlines and highlights the value of the proposed hierarchical architecture of ROMAN. 

\paragraph*{\textbf{Result Implications on Increasing Levels of Uncertainty and the Effects of the Employed Learning Algorithms: }} In this work, ROMAN's experts were provided with exteroceptive information relevant to their specific task goals and specialised skills, while the MN used the combined state space of the experts to oversee and supervise the context of the environment and the long-term sequential task. ROMAN was subsequently evaluated on a vision-based detection system. However, before proceeding to the vision system, it was deemed important to initially evaluate the framework beyond the exteroceptive uncertainties the vision system alone would yield. In regards to increasing levels of Gaussian uncertainty, it was observed that even a five-fold increase in Gaussian noise for exteroceptive states on ROMAN ($\sigma = \pm 2.5\text{cm}$) rendered high success rates, as supported by \autoref{table_sup:DetailROMANDifferentLevelsNoise}. The lowest success rate at that particular noise level was $76.2\%$ for the \textbf{Pick and Insert} expert which was arguably the most complex compared to the rest of the NNs in ROMAN's framework. It is also worth noting that the entailed experts operate at a higher level in terms of manipulation skills when compared to related work, for more details please consult the main manuscript.

In regards to the increasing levels of noise for the gating network (the MN), it was observed that by virtue of training ROMAN's MN with the employed hybrid learning approach of ``BC + GAIL + RL'' as detailed in the main manuscript. The employed hybrid learning procedure resulted in an overall higher and more robust performance than when comparing their different combinations. This suggests that an HHL approach exhibits significantly higher resilience against a multitude of different settings and in particular: (i) increased exteroceptive uncertainties as frequently encountered in real-life robotic cases, (ii) the presence of more complex non-interrelated sub-tasks, (iii) longer time-horizon sequential tasks, as well as (iv) the adaptation to cases beyond those encountered in the demonstration sequence with the ability to dynamically recover from local minima. 

\paragraph*{\textbf{Result Implications on Demonstrations: }} From the results based on the different number of demonstrations, it was observed that even a relatively small amount of demonstrations for the MN allowed the overall framework to retain ``acceptable" success rates even when presented with the most complex sequences of S6 and S7. Upon detailed evaluation of the results in \autoref{table_sup:DetailedDemonstrationsROMAN}, it was furthermore observed that the proposed HHL approach (BC + GAIL ($r_{I}$) + RL  ($r_{E}$)) allowed ROMAN to benefit from the initial warm-starting of the policy via the employed BC approach and to train a complex gating network with an overall small number of demonstrations due to the intrinsic reward provided by GAIL, while also retaining robust performance due to the added extrinsic RL reward encouraging exploration. 

\begin{figure}[H]
	\begin{center}
		\includegraphics[width=1.0\textwidth]{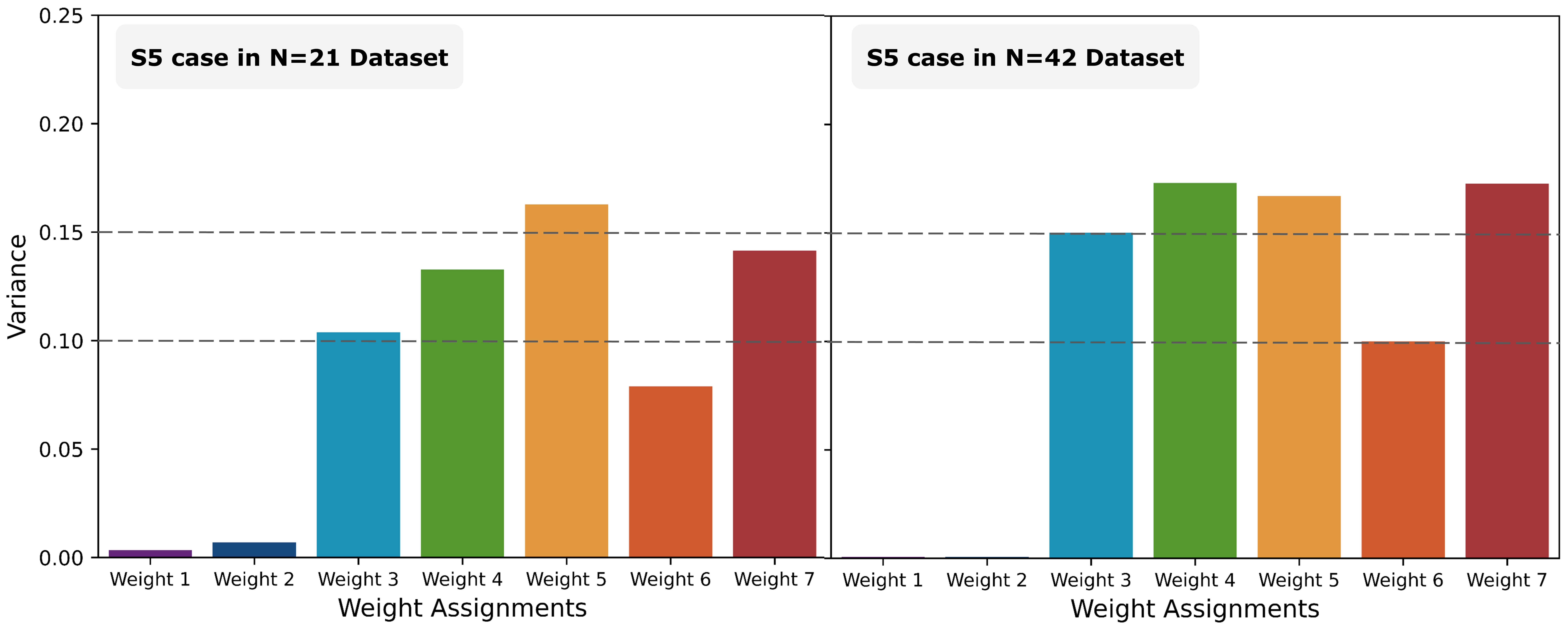}
	\end{center}
	\caption{\textbf{Depiction of the variation of actions between the demonstration datasets of N=21 and N=42 for scenario case S5.} The figure illustrates the variance of actions of the MN, which correspond to the weight orchestration and subsequent weight assignments to the incorporated experts. From the figure it is inferred that the variance of actions for scenario case S5 is overall higher in the N=42 compared to the N=21 dataset.}
	\label{sup_fig:demonstrationActionVariance}
\end{figure}

It is noteworthy to point out that there was an observed discrepancy regarding scenario case S5 in the results and in particular for $N=21$ and $N=42$ demonstrations. In particular, a 5\% performance difference for Scenario 5 (S5) between the N=21 and N=42 demonstration cases was observed. Upon further investigation, it was observed that the weight assignments of the learned MN policy corresponding to the demonstration datasets for the N=42 dataset had greater weight variation compared to the N=21 dataset. Due to the introduction of a human-generated dataset, these slight deviations around the ideally-optimal trajectories are naturally occurring over time and are not detrimental. Non-optimal demonstrations can still affect the learning process to some extent, however by virtue of the employed hybrid learning procedure which balances exploration and exploitation, even non-optimal provided demonstrations, to a certain degree, can still result in a robust policy. The variation of these weight assignments in the dataset of N=21 and N=42 specific to case S5 are visually depicted in \autoref{sup_fig:demonstrationActionVariance}.

To shed additional light into this discrepancy a reward function was applied to quantify task performance, and the standard deviation of the rewards was quantified and expressed as a percentage difference between the two datasets. It was found that with the exception of S5, most cases had roughly similar reward scores between the N=21 and N=42 datasets. Notably, S5 had 53.23\% more standard deviation in the case of N=42 compared to that of N=21. The standard deviations of these rewards are listed in \autoref{table_sup:StandardDevRewardDemos}.

It is worth noting that these results are based purely on imitation learning, which exhibits larger variations in the demonstration dataset which can, in turn, ultimately lead to larger variations in task performance. Although imperfect demonstrations can still affect the learned policy (i.e., 53.23\% more standard deviation for the imitation part), the hybrid learning procedure can balance exploration and exploitation via extrinsic task-related rewards, resulting in comparable task performance for both N=21 and N=42 cases (i.e., 90.3\% and 85.2\% success rate).

From the results, it can be initially inferred that a one-shot demonstration of each sequential case scenario ($N=7$) was not sufficient to solve scenario levels S4 to S7 but still retained relatively good success levels of more than 68\% for S1 to S3. We found that at $N=21$ demonstrations, which correspond to three demonstrations for every scenario, stable and significantly higher success rates were achieved compared to $N=7$, and were almost to the same level as that with $N=42$ dataset. Even though ROMAN was evaluated on $N=42$ demonstrations for most of the comparisons in the main manuscript, providing half of those demonstrations ($N=21$) to the MN is still sufficient for the framework to attain high success, even amongst the most complex and long-time horizon sequential tasks as evidenced by \autoref{table_sup:DetailedDemonstrationsROMAN}.

\subsection*{Supplementary Note S5: Simulation Environment and Control Framework}
\label{SN5}
In regards to the control framework used in the simulation environment, realistic physical properties were preserved to minimise future Sim2Real gaps and to approximate real-life physics as much as possible. The robotic system Franka Emika was controlled via a Proportional–Integral–Derivative (PID) controller. A Jacobian pseudoinverse method was used to compute the inverse kinematics for the end-effector/gripper. The inverse kinematics control was independently running, whenever the end-effector position is commanded by a human demonstration or by the neural networks that are controlling the end-effector during training and subsequent inference. Two levels of demonstrations were provided, in particular: (i) direct control of the end-effector gripper via a velocity command and the state of the gripper (open and close), which are used for training the expert NNs; and (ii) demonstration of sequences of activating expert skills for training the weight orchestration for training the MN. The binary control of opening and closing of the gripper was achieved via a binary signal with a Dead Zone (DZ) implementation as detailed in the manuscript. The environment was observed visually by the human expert via a generic monocular display monitor, illustrating the simulation in an orthographic view.

\subsection*{Supplementary Note S6: ROMAN's Hybrid Learning and Architecture}
\label{SN6}
In addition to the training details outlined in the main manuscript, in this section, the procedure used to train all incorporated NNs within the hierarchical formation of ROMAN's hybrid learning architecture is elaborated upon. \autoref{sup_fig:trainingFlowDiagram} visually depicts the employed hybrid training procedure within ROMAN's hierarchical architecture in the form of a flow diagram. \autoref{alg:training_procedure} details the training procedure in the form of a pseudo-code. The symbol $r_{E}$ represents the extrinsic task-related reward collected from the environment as per the RL paradigm, while $r_{I}$ the intrinsic reward provided by GAIL's discriminator to the RL policy (PPO). Finally, $w_{E}$ and $w_{I}$ represent the extrinsic and intrinsic weights to their respective reward terms. 

\autoref{table_sup:ROMAN_2D_DetailedNetworkCharacteristicsStructure} illustrates ROMAN's preliminary stage composed of five in total experts in 2D space, detailing the states and action space of all the incorporated NNs in the hierarchical formation. The equivalent table for ROMAN's hierarchical architecture at its final stage composed of seven experts, operating in 3D space can be found in the Main Manuscript Table 1.a. \autoref{table_sup:roman_nn_architectures} depicts the architectural overview of ROMAN's final stage in 3D space composed of seven experts, including the details of the states, actions, demonstrations and dimensions overview of all incorporated NNs. Furthermore, the hyperparameters used for the HHL procedure of ROMAN's NNs are listed in \autoref{table_sup:Hyperparameters}.

In regards to the reward design of the expert NNs, their rewards were specific to their sub-tasks and the nature of their objectives. These were through a combination of sparse rewards with a terminal reward at the end of a successful episode. The rewards for the Manipulation Network (MN) were sparse. More specifically, a reward signal was provided to the MN after successfully completing each sub-task and a terminal reward for successfully completing the entire sequence of tasks successfully. 

\subsection*{Supplementary Note S7: The Architectures and Training of Monolithic Networks}
\label{SN7}
Aside from the hierarchical composition of ROMAN, as aforementioned, two monolithic neural networks were trained for 2D and 3D manipulation tasks respectively as baseline evaluations. The performance of these monolithic networks was subsequently compared to that of ROMAN's hierarchical architecture in both its preliminary 2D stage composed of five experts and its final 3D stage composed of seven experts. \autoref{table_sup:SingleNN_2D_3D} depicts the architectures of these monolithic NNs in 2D and 3D spaces, detailing their characteristics and more specifically their states, actions, demonstrations and neural network architectures. As aforementioned, an identical hybrid learning procedure was used to train the monolithic NNs and ROMAN to adhere to consistency and to conduct fair comparisons. Moreover, the same hyperparameter values were used for the monolithic NNs as for ROMAN's MN, in 2D and 3D respectively. The specific hyperparameters used for the monolithic NNs in 2D and 3D are listed in \autoref{table_sup:SingleNN_2D_3D_Hyperparameters}. Ultimately, this allowed for the direct comparison of monolithic approaches and underlining their limitations and eventually the necessity and usefulness of a hierarchical task decomposition. 

\newpage
\section*{Algorithm Pseudo-Code}

\begin{algorithm}
\caption{Algorithmic of the Hybrid-Learning Procedure employed in ROMAN}\label{alg:training_procedure}
\begin{algorithmic}[1]
\State \textbf{Input}: A policy network with parameters $\theta$, a discriminator network with parameters $\theta_d$ 
\State
\State \textbf{Behavioural Cloning to Warm-Start the Policy}
\For{$epoch$ in $[1, num\_bc\_epochs]$}
    \State Collect expert trajectories
    \State $\text{trajectories} \gets \text{collect\_expert\_trajectories}(\text{env}, \text{expert\_policy})$
    
    \State Compute BC loss and update policy network
    \State $\text{loss\_bc} \gets \text{compute\_bc\_loss}(\text{policy}, \text{trajectories})$
    \State $\nabla_\theta \gets \text{compute\_gradient}(\text{loss\_bc}, \theta)$
    \State $\theta \gets \text{update\_weights}(\theta, \nabla_\theta)$
\EndFor
\State
\State \textbf{Reinforcement Learning (Proximal Policy Optimization) with Extrinsic and Intrinsic (GAIL) Rewards:}
\For{$epoch$ in $[1, num\_ppo\_epochs]$}
    \State Collect trajectories using PPO with both $r_{E}$ and $r_{I}$
    \State $\text{trajectories} \gets \text{collect\_ppo\_trajectories}(\text{env}, \text{policy}, \text{$w_{E}$}, \text{$w_{I}$})$
    
    \State Train discriminator with both expert and generated trajectories, using only ($s_{t}$)
    \State $\text{loss\_discriminator} \gets \text{compute\_gail\_loss}(\text{discriminator}, \text{expert\_traj}, \text{generated\_traj})$
    \State $\nabla_{\theta_d} \gets \text{compute\_gradient}(\text{loss\_discriminator}, \theta_d)$
    \State $\theta_d \gets \text{update\_weights}(\theta_d, \nabla_{\theta_d})$
    
    \State Update policy network with PPO using extrinsic and intrinsic (GAIL discriminator) reward
    \State $\text{loss\_policy} \gets \text{compute\_ppo\_loss}(\text{policy}, \text{trajectories}, \text{discriminator}, \text{$w_{E}$}, \text{$w_{I}$})$
    \State $\nabla_\theta \gets \text{compute\_gradient}(\text{loss\_policy}, \theta)$
    \State $\theta \gets \text{update\_weights}(\theta, \nabla_\theta)$
\EndFor
\end{algorithmic}
\end{algorithm}

\newpage
\section*{Supplementary Figures}

\clearpage

\clearpage

\clearpage

\clearpage

\newpage
\section*{Supplementary Tables}

\begin{table}[H]
\centering
\caption{\textbf{The derived and investigated manipulation tasks and their goals and interdependencies with respect to the evaluated simulation task.} A successful task is attained when the successful completion of all seven of the aforementioned task goals is met. Consequently, a single failure amongst the sequences of the below-mentioned tasks is treated as an overall error in our subsequent results analyses. The derived tasks represent an example of a long-time horizon sequential task that is decomposed into smaller sub-tasks, commonly encountered in sequential robotic tasks.}
\begin{small}
\begin{tabular}{llcc} \bottomrule
\rowcolor{DarkerGray} \multicolumn{4}{c}{{\textbf{Derived Sequence of Tasks}}} \\ 
\rowcolor{DarkGray} \multicolumn{2}{c}{\textbf{Manipulation Task and Goal}} & \textbf{Dependency} & \textbf{Scenario Case} \\ \hline
(i) & Open drawer to expose container encasing vial & - & Scenario 7 \\ \hline
\rowcolor{Gray} (ii) & Unbox container to expose vial & (i) & Scenario 6 \\ \hline
(iii) & Open cabinet to expose rack & - & Scenario 5 \\ \hline
\rowcolor{Gray} (iv) & Place rack on top of cabinet & (iii) & Scenario 4 \\ \hline
(v) & Retrieve vial and place it onto rack  & All of the above & Scenario 3 \\ \hline
\rowcolor{Gray} (vi) & Push rack containing vial onto conveyor & All of the above & Scenario 2 \\ \hline
(vii) & Activate conveyor by pressing button & All of the above & Scenario 1 \\
\bottomrule
\end{tabular}
\end{small}
\label{table_sup::TasksWithGoalsDependency}
\end{table}

\clearpage
\begin{table}[H]
\centering
\caption{\textbf{Detailed results of ROMAN expanding upon the Main Manuscript Table 1.b, where the overall success of each sequential case scenario is depicted over the different levels of Gaussian uncertainty, with the individual expert success highlighted.} This table summarises the overall success rates of the individual expert sequences within the action sequence over the given scenario task and the studied Gaussian level of noise. The grey cells are denoted as n/a and hence omitted from the analysis as these are not part of the sequential actions needed to satisfy the end-goal task. In other words, only the cells that are relevant to the scenario are studied. For an overall scenario to be deemed successful, all relevant sub-tasks needed to be satisfied. A total of $N=42$ demonstrations were provided to the MN, which corresponds to $N=6$ per each of the seven case scenarios.}
\begin{footnotesize}
\setlength\tabcolsep{5pt}
        \resizebox{\textwidth}{!}{
\begin{tabular}{ll|rrrrrrr|r} \bottomrule
\rowcolor{DarkerGray} \multicolumn{10}{c}{{\textbf{Detailed Success Rates of ROMAN Against Different Levels of Uncertainty}}} \\ \hline
\rowcolor{DarkGray} &  \textbf{Scenario Cases} & \textbf{Push-Button} & \textbf{Push} & \textbf{Pick \& Insert} & \textbf{Pick \& Place} & \textbf{Rotate Open} & \textbf{Pick \& Drop} & \textbf{Pull-Open} & \textbf{Overall} \\ \hline

{\multirow{7}{*}{\rotatebox[origin=c]{90}{$\underline{\sigma = \pm 0.0 \text{ [cm]}}$}}} & \textbf{S1:} \textit{Push-Button}  & {0.976} & \cellcolor{Gray} n/a & \cellcolor{Gray} n/a & \cellcolor{Gray} n/a & \cellcolor{Gray} n/a & \cellcolor{Gray} n/a & \cellcolor{Gray} n/a & {0.976}\\

& \textbf{S2:} \textit{+Push} & {0.972} & {0.991} & \cellcolor{Gray} n/a & \cellcolor{Gray} n/a & \cellcolor{Gray} n/a &  \cellcolor{Gray} n/a & \cellcolor{Gray} n/a & {0.972}\\

& \textbf{S3:} \textit{+Pick \& Insert} & {0.869} & {0.876} & {0.881} & \cellcolor{Gray} n/a & \cellcolor{Gray} n/a &  \cellcolor{Gray} n/a & \cellcolor{Gray} n/a & {0.847}\\

& \textbf{S4:} \textit{+Pick \& Place} & {0.955} & {0.975} & {0.971} & {0.982} & \cellcolor{Gray} n/a &  \cellcolor{Gray} n/a & \cellcolor{Gray} n/a & {0.951}\\

& \textbf{S5:} \textit{+Rotate Open} & {0.743} & {0.750} & {0.753} & {0.768} & {0.771} &  \cellcolor{Gray} n/a & \cellcolor{Gray} n/a & {0.728}\\

& \textbf{S6:} \textit{+Pick \& Drop} & {0.961} & {0.968} & {0.965} & {0.976} & {0.999} &  {0.993} & \cellcolor{Gray} n/a & {0.954}\\

& \textbf{S7:} \textit{+Pull-Open} & {0.912} & {0.924} & {0.917} & {0.942} & {0.972} &  {0.973} &  {0.972} & {0.903} \\ 

\hline

{\multirow{7}{*}{\rotatebox[origin=c]{90}{$\underline{\sigma = \pm 0.5 \text{ [cm]}}$}}} & \textbf{S1:} \textit{Push-Button}  & {0.973} & \cellcolor{Gray} n/a & \cellcolor{Gray} n/a & \cellcolor{Gray} n/a & \cellcolor{Gray} n/a & \cellcolor{Gray} n/a & \cellcolor{Gray} n/a & {0.973}\\

& \textbf{S2:} \textit{+Push} & {0.975} & {0.989} & \cellcolor{Gray} n/a & \cellcolor{Gray} n/a & \cellcolor{Gray} n/a &  \cellcolor{Gray} n/a & \cellcolor{Gray} n/a & {0.975}\\

& \textbf{S3:} \textit{+Pick \& Insert} & {0.841} & {0.854} & {0.869} & \cellcolor{Gray} n/a & \cellcolor{Gray} n/a &  \cellcolor{Gray} n/a & \cellcolor{Gray} n/a & {0.817}\\

& \textbf{S4:} \textit{+Pick \& Place} & {0.965} & {0.966} & {0.966} & {0.985} & \cellcolor{Gray} n/a &  \cellcolor{Gray} n/a & \cellcolor{Gray} n/a & {0.959}\\

& \textbf{S5:} \textit{+Rotate Open} & {0.811} & {0.822} & {0.820} & {0.835} & {0.833} &  \cellcolor{Gray} n/a & \cellcolor{Gray} n/a & {0.794}\\

& \textbf{S6:} \textit{+Pick \& Drop} & {0.967} & {0.973} & {0.971} & {0.983} & {0.987} &  {0.992} & \cellcolor{Gray} n/a & {0.960}\\

& \textbf{S7:} \textit{+Pull-Open} & {0.958} & {0.964} & {0.962} & {0.983} & {0.991} &  {0.993} &  {0.994} & {0.952} \\ 

\hline

{\multirow{7}{*}{\rotatebox[origin=c]{90}{$\underline{\sigma = \pm 1.0 \text{ [cm]}}$}}} & \textbf{S1:} \textit{Push-Button}  & {0.977} & \cellcolor{Gray} n/a & \cellcolor{Gray} n/a & \cellcolor{Gray} n/a & \cellcolor{Gray} n/a & \cellcolor{Gray} n/a & \cellcolor{Gray} n/a & {0.977}\\

& \textbf{S2:} \textit{+Push} & {0.999} & {0.997} & \cellcolor{Gray} n/a & \cellcolor{Gray} n/a & \cellcolor{Gray} n/a &  \cellcolor{Gray} n/a & \cellcolor{Gray} n/a & {0.990}\\

& \textbf{S3:} \textit{+Pick \& Insert} & {0.831} & {0.841} & {0.843} & \cellcolor{Gray} n/a & \cellcolor{Gray} n/a &  \cellcolor{Gray} n/a & \cellcolor{Gray} n/a & {0.798}\\

& \textbf{S4:} \textit{+Pick \& Place} & {0.955} & {0.960} & {0.955} & {0.980} & \cellcolor{Gray} n/a &  \cellcolor{Gray} n/a & \cellcolor{Gray} n/a & {0.946}\\

& \textbf{S5:} \textit{+Rotate Open} & {0.796} & {0.802} & {0.799} & {0.813} & {0.816} &  \cellcolor{Gray} n/a & \cellcolor{Gray} n/a & {0.776}\\

& \textbf{S6:} \textit{+Pick \& Drop} & {0.952} & {0.955} & {0.946} & {0.956} & {0.974} &  {0.984} & \cellcolor{Gray} n/a & {0.933}\\

& \textbf{S7:} \textit{+Pull-Open} & {0.954} & {0.955} & {0.944} & {0.970} & {0.987} &  {0.999} &  {0.989} & {0.939} \\ 

\hline

{\multirow{7}{*}{\rotatebox[origin=c]{90}{$\underline{\sigma = \pm 1.5 \text{ [cm]}}$}}} & \textbf{S1:} \textit{Push-Button}  & {0.980} & \cellcolor{Gray} n/a & \cellcolor{Gray} n/a & \cellcolor{Gray} n/a & \cellcolor{Gray} n/a & \cellcolor{Gray} n/a & \cellcolor{Gray} n/a & {0.980}\\

& \textbf{S2:} \textit{+Push} & {0.986} & {0.996} & \cellcolor{Gray} n/a & \cellcolor{Gray} n/a & \cellcolor{Gray} n/a &  \cellcolor{Gray} n/a & \cellcolor{Gray} n/a & {0.986}\\

& \textbf{S3:} \textit{+Pick \& Insert} & {0.756} & {0.767} & {0.794} & \cellcolor{Gray} n/a & \cellcolor{Gray} n/a &  \cellcolor{Gray} n/a & \cellcolor{Gray} n/a & {0.720}\\

& \textbf{S4:} \textit{+Pick \& Place} & {0.881} & {0.882} & {0.874} & {0.930} & \cellcolor{Gray} n/a &  \cellcolor{Gray} n/a & \cellcolor{Gray} n/a & {0.846}\\

& \textbf{S5:} \textit{+Rotate Open} & {0.767} & {0.766} & {0.758} & {0.808} & {0.853} &  \cellcolor{Gray} n/a & \cellcolor{Gray} n/a & {0.722}\\

& \textbf{S6:} \textit{+Pick \& Drop} & {0.875} & {0.870} & {0.860} & {0.912} & {0.973} &  {0.978} & \cellcolor{Gray} n/a & {0.836}\\

& \textbf{S7:} \textit{+Pull-Open} & {0.875} & {0.871} & {0.862} & {0.933} & {0.979} &  {0.983} &  {0.986} & {0.841} \\ 

\hline

{\multirow{7}{*}{\rotatebox[origin=c]{90}{$\underline{\sigma = \pm 2.0 \text{ [cm]}}$}}} & \textbf{S1:} \textit{Push-Button}  & {0.967} & \cellcolor{Gray} n/a & \cellcolor{Gray} n/a & \cellcolor{Gray} n/a & \cellcolor{Gray} n/a & \cellcolor{Gray} n/a & \cellcolor{Gray} n/a & {0.967}\\

& \textbf{S2:} \textit{+Push} & {0.986} & {0.994} & \cellcolor{Gray} n/a & \cellcolor{Gray} n/a & \cellcolor{Gray} n/a &  \cellcolor{Gray} n/a & \cellcolor{Gray} n/a & {0.986}\\

& \textbf{S3:} \textit{+Pick \& Insert} & {0.768} & {0.783} & {0.805} & \cellcolor{Gray} n/a & \cellcolor{Gray} n/a &  \cellcolor{Gray} n/a & \cellcolor{Gray} n/a & {0.737}\\

& \textbf{S4:} \textit{+Pick \& Place} & {0.862} & {0.865} & {0.859} & {0.901} & \cellcolor{Gray} n/a &  \cellcolor{Gray} n/a & \cellcolor{Gray} n/a & {0.837}\\

& \textbf{S5:} \textit{+Rotate Open} & {0.785} & {0.790} & {0.790} & {0.829} & {0.892} &  \cellcolor{Gray} n/a & \cellcolor{Gray} n/a & {0.753}\\

& \textbf{S6:} \textit{+Pick \& Drop} & {0.842} & {0.843} & {0.850} & {0.882} & {0.943} &  {0.954} & \cellcolor{Gray} n/a & {0.820}\\

& \textbf{S7:} \textit{+Pull-Open} & {0.840} & {0.839} & {0.840} & {0.886} & {0.973} &  {0.977} &  {0.983} & {0.815} \\ 

\hline

{\multirow{7}{*}{\rotatebox[origin=c]{90}{$\underline{\sigma = \pm 2.5 \text{ [cm]}}$}}} & \textbf{S1:} \textit{Push-Button}  & {0.973} & \cellcolor{Gray} n/a & \cellcolor{Gray} n/a & \cellcolor{Gray} n/a & \cellcolor{Gray} n/a & \cellcolor{Gray} n/a & \cellcolor{Gray} n/a & {0.973}\\

& \textbf{S2:} \textit{+Push} & {0.986} & {0.995} & \cellcolor{Gray} n/a & \cellcolor{Gray} n/a & \cellcolor{Gray} n/a &  \cellcolor{Gray} n/a & \cellcolor{Gray} n/a & {0.986}\\

& \textbf{S3:} \textit{+Pick \& Insert} & {0.767} & {0.774} & {0.778} & \cellcolor{Gray} n/a & \cellcolor{Gray} n/a &  \cellcolor{Gray} n/a & \cellcolor{Gray} n/a & {0.723}\\

& \textbf{S4:} \textit{+Pick \& Place} & {0.789} & {0.790} & {0.800} & {0.856} & \cellcolor{Gray} n/a &  \cellcolor{Gray} n/a & \cellcolor{Gray} n/a & {0.763}\\

& \textbf{S5:} \textit{+Rotate Open} & {0.740} & {0.748} & {0.739} & {0.784} & {0.898} &  \cellcolor{Gray} n/a & \cellcolor{Gray} n/a & {0.697}\\

& \textbf{S6:} \textit{+Pick \& Drop} & {0.751} & {0.749} & {0.769} & {0.807} & {0.931} &  {0.935} & \cellcolor{Gray} n/a & {0.719}\\

& \textbf{S7:} \textit{+Pull-Open} & {0.788} & {0.785} & {0.762} & {0.830} & {0.962} &  {0.964} &  {0.980} & {0.744} \\ 

\hline
\end{tabular}}
\end{footnotesize}
\label{table_sup:DetailROMANDifferentLevelsNoise}
\end{table}

\clearpage
\begin{table}[H]
\centering
\caption{\textbf{The detailed results of a single NN vs the preliminary stage of the hierarchical framework ROMAN in 2D space consisting of five experts, expanding upon Main Manuscript Table 2.a.} The table summarises the percentage of successful sequential scenarios over the five in total decomposed sub-tasks. Consequently, the sub-tasks listed under the five different columns are perceived here as different sub-tasks within the time horizon of the full sequence of the episode, rather than individual experts. This allowed, in turn, for a more in-depth analysis as to which sub-tasks in the sequence of tasks failed when comparing the monolithic NN vs the hierarchical formation of ROMAN. A total of $N=35$ demonstrations were provided to the MN of ROMAN, while for the single NN, a total of $N=100$ demonstrations were given. This decision was made as to conduct a fair comparison and to account for ROMAN's five experts in total that were pre-trained with $N=20$ demonstrations ($5 \text{\tiny experts} \times 20 \text{\tiny demos}$). Identical hyperparameters were used for both the monolithic NN and ROMAN's framework, the states of the single NN corresponded to that of the MN's, while the actions of the single NN were in direct control of the robotic-end effector and the gripper state, similar to the expert NNs of ROMAN's. Evaluated on $\sigma = \pm 0.5 \text{ [cm]}$ level of noise.}
\begin{footnotesize}
\setlength\tabcolsep{10pt}
        \resizebox{\textwidth}{!}{
\begin{tabular}{ll|rrrrr|r} \bottomrule
\rowcolor{DarkerGray} \multicolumn{8}{c}{{\textbf{[2D] Monolithic Single NN vs Preliminary Stage of the Hierarchical Framework ROMAN}}} \\ \hline
\rowcolor{DarkGray} &  \textbf{Scenario Cases} & \textbf{Push} & \textbf{Lift} & \textbf{Pick \& Insert}& \textbf{Pick \& Drop} & \textbf{Pull} & \textbf{Overall} \\ \hline

{\multirow{5}{*}{\rotatebox[origin=c]{90}{\underline{\footnotesize Single NN}}}} & \textbf{S1:} \textit{Push}  & {0.997} & \cellcolor{Gray} n/a & \cellcolor{Gray} n/a & \cellcolor{Gray} n/a & \cellcolor{Gray} n/a & {0.997}\\

& \textbf{S2:} \textit{+Lift} & {0.851} & {0.987} & \cellcolor{Gray} n/a & \cellcolor{Gray} n/a & \cellcolor{Gray} n/a & {0.841}\\

& \textbf{S3:} \textit{+Pick \& Insert} & {0.843} & {0.945} & {0.823} & \cellcolor{Gray} n/a & \cellcolor{Gray} n/a & {0.699}\\

& \textbf{S4:} \textit{+Pick \& Drop} & {0.702} & {0.782} & {0.679} & {0.869} & \cellcolor{Gray} n/a & {0.591}\\

& \textbf{S5:} \textit{+Pull} & {0.718} & {0.832} & {0.700} & {0.919} & {0.987} & {0.565}\\

\hline
 
{\multirow{5}{*}{\rotatebox[origin=c]{90}{\underline{\footnotesize Single NN}}}} & \textbf{S1:} \textit{Push}  & {0.993} & \cellcolor{Gray} n/a & \cellcolor{Gray} n/a & \cellcolor{Gray} n/a & \cellcolor{Gray} n/a & {0.993}\\

& \textbf{S2:} \textit{+Lift} & {0.996} & {0.997} & \cellcolor{Gray} n/a & \cellcolor{Gray} n/a & \cellcolor{Gray} n/a & {0.995}\\

& \textbf{S3:} \textit{+Pick \& Insert} & {0.996} & {0.997} & {0.987} & \cellcolor{Gray} n/a & \cellcolor{Gray} n/a & {0.982}\\

& \textbf{S4:} \textit{+Pick \& Drop} & {0.977} & {0.981} & {0.978} & {0.991} & \cellcolor{Gray} n/a & {0.971}\\

& \textbf{S5:} \textit{+Pull} & {0.985} & {0.986} & {0.986} & {0.993} & {0.998} & {0.974}\\

\hline
\end{tabular}}
\end{footnotesize}
\label{table_sup:2DDetailedSingleNNvsROMAN}
\end{table}

\begin{table}[H]
\centering
\caption{\textbf{Detailed results of a monolithic NN vs the hierarchical architecture of ROMAN, expanding upon Main Manuscript Table 2.b.} The table presents the percentage of successful sequential scenarios for each decomposed sub-task, each listed under the seven different columns and considered as part of the full sequence of the episode. This approach allowed for an in-depth analysis at to which sub-tasks in the sequence of tasks failed when comparing the monolithic NN with the hierarchical architecture of ROMAN. ROMAN's MN was provided $N=42$, while the single NN in 3D was given a total of $N=140$ demonstrations to account for ROMAN's 7 in total experts pre-trained with $N=20$ demonstrations ($7 \text{\tiny experts} \times 20 \text{\tiny demos}$). Moreover, to retain consistency and conduct a fair comparison, an identical hybrid learning procedure as outlined in the main manuscript was employed between the two, with identical states (to ROMAN's MN), actions (to ROMAN's expert NNs) and hyperparameters. Evaluated on $\sigma = \pm 0.5 \text{ [cm]}$ level of noise.}
\begin{footnotesize}
\setlength\tabcolsep{5pt}
        \resizebox{\textwidth}{!}{
\begin{tabular}{ll|rrrrrrr|r} \bottomrule
\rowcolor{DarkerGray} \multicolumn{10}{c}{{\textbf{[3D] Monolithic Single NN vs Hierarchical Framework ROMAN}}} \\ \hline
\rowcolor{DarkGray} &  \textbf{Scenario Cases} & \textbf{Push-Button} & \textbf{Push} & \textbf{Pick \& Insert} & \textbf{Pick \& Place} & \textbf{Rotate Open} & \textbf{Pick \& Drop} & \textbf{Pull-Open} & \textbf{Overall} \\ \hline

{\multirow{7}{*}{\rotatebox[origin=c]{90}{\underline{Single NN ($\pm 0.5 \text{ cm}$)}}}} & \textbf{S1:} \textit{Push-Button}  & {0.997} & \cellcolor{Gray} n/a & \cellcolor{Gray} n/a & \cellcolor{Gray} n/a & \cellcolor{Gray} n/a & \cellcolor{Gray} n/a & \cellcolor{Gray} n/a & {0.997}\\

& \textbf{S2:} \textit{+Push} & {0.982} & {0.987} & \cellcolor{Gray} n/a & \cellcolor{Gray} n/a & \cellcolor{Gray} n/a &  \cellcolor{Gray} n/a & \cellcolor{Gray} n/a & {0.981}\\

& \textbf{S3:} \textit{+Pick \& Insert} & {0.671} & {0.678} & {0.587} & \cellcolor{Gray} n/a & \cellcolor{Gray} n/a &  \cellcolor{Gray} n/a & \cellcolor{Gray} n/a & {0.583}\\

& \textbf{S4:} \textit{+Pick \& Place} & {0.041} & {0.046} & {0.036} & {0.103} & \cellcolor{Gray} n/a &  \cellcolor{Gray} n/a & \cellcolor{Gray} n/a & {0.032}\\

& \textbf{S5:} \textit{+Rotate Open} & {0.041} & {0.044} & {0.035} & {0.076} & {0.934} &  \cellcolor{Gray} n/a & \cellcolor{Gray} n/a & {0.028}\\

& \textbf{S6:} \textit{+Pick \& Drop} & {0.000} & {0.000} & {0.000} & {0.006} & {0.006} &  {0.024} & \cellcolor{Gray} n/a & {0.000}\\

& \textbf{S7:} \textit{+Pull-Open} & {0.000} & {0.000} & {0.000} & {0.000} & {0.010} &  {0.028} &  {0.877} & {0.000} \\ 
\hline

{\multirow{7}{*}{\rotatebox[origin=c]{90}{\underline{ROMAN ($\pm 0.5 \text{ cm}$)}}}} & \textbf{S1:} \textit{Push-Button}  & {0.973} & \cellcolor{Gray} n/a & \cellcolor{Gray} n/a & \cellcolor{Gray} n/a & \cellcolor{Gray} n/a & \cellcolor{Gray} n/a & \cellcolor{Gray} n/a & {0.973}\\

& \textbf{S2:} \textit{+Push} & {0.975} & {0.989} & \cellcolor{Gray} n/a & \cellcolor{Gray} n/a & \cellcolor{Gray} n/a &  \cellcolor{Gray} n/a & \cellcolor{Gray} n/a & {0.975}\\

& \textbf{S3:} \textit{+Pick \& Insert} & {0.841} & {0.854} & {0.869} & \cellcolor{Gray} n/a & \cellcolor{Gray} n/a &  \cellcolor{Gray} n/a & \cellcolor{Gray} n/a & {0.817}\\

& \textbf{S4:} \textit{+Pick \& Place} & {0.965} & {0.966} & {0.966} & {0.985} & \cellcolor{Gray} n/a &  \cellcolor{Gray} n/a & \cellcolor{Gray} n/a & {0.959}\\

& \textbf{S5:} \textit{+Rotate Open} & {0.882} & {0.892} & {0.892} & {0.938} & {0.949} &  \cellcolor{Gray} n/a & \cellcolor{Gray} n/a & {0.852}\\

& \textbf{S6:} \textit{+Pick \& Drop} & {0.967} & {0.973} & {0.971} & {0.983} & {0.987} &  {0.992} & \cellcolor{Gray} n/a & {0.960}\\

& \textbf{S7:} \textit{+Pull-Open} & {0.958} & {0.964} & {0.962} & {0.983} & {0.991} &  {0.993} &  {0.994} & {0.952} \\ 
\hline
\end{tabular}}
\end{footnotesize}
\label{table_sup:3DDetailedSingleNNvsROMAN}
\end{table}

\clearpage
\begin{table}[H]
\centering
\caption{\textbf{Detailed results of ROMAN expanding upon Main Manuscript Table 2.c, where the overall success of each sequential case scenario is depicted with the use of different learning algorithms at a $\sigma = \pm 0.5 \text{ [cm]}$ level of Gaussian noise.} The table summarises the overall success rates of each sequential scenario. Moreover, the individual expert success rate is detailed, based on different combinations of BC, RL and GAIL. \textbf{Note on BC:} Supervised learning on the demonstration dataset. \textbf{Note on GAIL:} Use of IL intrinsic rewards ($r_{I}$) provided to PPO. \textbf{Note on RL:} Use of task extrinsic rewards ($r_{E}$) provided to PPO. \textbf{ROMAN's $\dagger$:} Default HHL approach combining BC, IL (via $r_{I}$) and RL (via $r_{E}$).}
\begin{footnotesize}
\setlength\tabcolsep{5pt}
        \resizebox{\textwidth}{!}{
\begin{tabular}{ll|rrrrrrr|r} \bottomrule
\rowcolor{DarkerGray} \multicolumn{10}{c}{{\textbf{Detailed Success Rates of Different Algorithms within ROMAN at Noise Level: $\pm 0.5$ cm}}} \\ \hline
\rowcolor{DarkGray} &  \textbf{Scenario Cases} & \textbf{Push-Button} & \textbf{Push} & \textbf{Pick \& Insert} & \textbf{Pick \& Place} & \textbf{Rotate Open} & \textbf{Pick \& Drop} & \textbf{Pull-Open} & \textbf{Overall} \\ \hline

{\multirow{7}{*}{\rotatebox[origin=c]{90}{\underline{RL}}}} & \textbf{S1:} \textit{Push-Button}  & {0.000} & \cellcolor{Gray} n/a & \cellcolor{Gray} n/a & \cellcolor{Gray} n/a & \cellcolor{Gray} n/a & \cellcolor{Gray} n/a & \cellcolor{Gray} n/a & {0.000}\\

& \textbf{S2:} \textit{+Push} & {0.000} & {0.035} & \cellcolor{Gray} n/a & \cellcolor{Gray} n/a & \cellcolor{Gray} n/a &  \cellcolor{Gray} n/a & \cellcolor{Gray} n/a & {0.000}\\

& \textbf{S3:} \textit{+Pick \& Insert} & {0.000} & {0.000} & {0.000} & \cellcolor{Gray} n/a & \cellcolor{Gray} n/a &  \cellcolor{Gray} n/a & \cellcolor{Gray} n/a & {0.000}\\

& \textbf{S4:} \textit{+Pick \& Place} & {0.000} & {0.000} & {0.000} & {0.000} & \cellcolor{Gray} n/a &  \cellcolor{Gray} n/a & \cellcolor{Gray} n/a & {0.000}\\

& \textbf{S5:} \textit{+Rotate Open} & {0.000} & {0.000} & {0.000} & {0.000} & {0.000} &  \cellcolor{Gray} n/a & \cellcolor{Gray} n/a & {0.000}\\

& \textbf{S6:} \textit{+Pick \& Drop} & {0.000} & {0.000} & {0.000} & {0.000} & {0.000} &  {0.000} & \cellcolor{Gray} n/a & {0.000}\\

& \textbf{S7:} \textit{+Pull-Open} & {0.000} & {0.000} & {0.000} & {0.000} & {0.000} &  {0.000} &  {0.000} & {0.000} \\ 
\hline

{\multirow{7}{*}{\rotatebox[origin=c]{90}{\underline{GAIL}}}} & \textbf{S1:} \textit{Push-Button}  & {0.980} & \cellcolor{Gray} n/a & \cellcolor{Gray} n/a & \cellcolor{Gray} n/a & \cellcolor{Gray} n/a & \cellcolor{Gray} n/a & \cellcolor{Gray} n/a & {0.980}\\

& \textbf{S2:} \textit{+Push} & {0.946} & {0.493} & \cellcolor{Gray} n/a & \cellcolor{Gray} n/a & \cellcolor{Gray} n/a &  \cellcolor{Gray} n/a & \cellcolor{Gray} n/a & {0.468}\\

& \textbf{S3:} \textit{+Pick \& Insert} & {0.610} & {0.649} & {0.708} & \cellcolor{Gray} n/a & \cellcolor{Gray} n/a &  \cellcolor{Gray} n/a & \cellcolor{Gray} n/a & {0.559}\\

& \textbf{S4:} \textit{+Pick \& Place} & {0.014} & {0.013} & {0.015} & {0.028} & \cellcolor{Gray} n/a &  \cellcolor{Gray} n/a & \cellcolor{Gray} n/a & {0.012}\\

& \textbf{S5:} \textit{+Rotate Open} & {0.005} & {0.006} & {0.008} & {0.034} & {0.757} &  \cellcolor{Gray} n/a & \cellcolor{Gray} n/a & {0.003}\\

& \textbf{S6:} \textit{+Pick \& Drop} & {0.004} & {0.005} & {0.006} & {0.025} & {0.755} &  {0.980} & \cellcolor{Gray} n/a & {0.001}\\

& \textbf{S7:} \textit{+Pull-Open} & {0.003} & {0.004} & {0.006} & {0.020} & {0.798} &  {0.972} &  {0.847} & {0.000} \\
\hline

{\multirow{7}{*}{\rotatebox[origin=c]{90}{\underline{BC}}}} & \textbf{S1:} \textit{Push-Button}  & {0.986} & \cellcolor{Gray} n/a & \cellcolor{Gray} n/a & \cellcolor{Gray} n/a & \cellcolor{Gray} n/a & \cellcolor{Gray} n/a & \cellcolor{Gray} n/a & {0.986}\\

& \textbf{S2:} \textit{+Push} & {0.979} & {0.994} & \cellcolor{Gray} n/a & \cellcolor{Gray} n/a & \cellcolor{Gray} n/a &  \cellcolor{Gray} n/a & \cellcolor{Gray} n/a & {0.978}\\

& \textbf{S3:} \textit{+Pick \& Insert} & {0.828} & {0.844} & {0.826} & \cellcolor{Gray} n/a & \cellcolor{Gray} n/a &  \cellcolor{Gray} n/a & \cellcolor{Gray} n/a & {0.786}\\

& \textbf{S4:} \textit{+Pick \& Place} & {0.694} & {0.689} & {0.677} & {0.830} & \cellcolor{Gray} n/a &  \cellcolor{Gray} n/a & \cellcolor{Gray} n/a & {0.660}\\

& \textbf{S5:} \textit{+Rotate Open} & {0.565} & {0.557} & {0.546} & {0.650} & {0.698} &  \cellcolor{Gray} n/a & \cellcolor{Gray} n/a & {0.525}\\

& \textbf{S6:} \textit{+Pick \& Drop} & {0.768} & {0.764} & {0.745} & {0.887} & {0.937} &  {0.983} & \cellcolor{Gray} n/a & {0.722}\\

& \textbf{S7:} \textit{+Pull-Open} & {0.813} & {0.798} & {0.776} & {0.889} & {0.934} &  {0.958} &  {0.964} & {0.760} \\ 
\hline

{\multirow{7}{*}{\rotatebox[origin=c]{90}{\underline{RL, GAIL}}}} & \textbf{S1:} \textit{Push-Button}  & {0.981} & \cellcolor{Gray} n/a & \cellcolor{Gray} n/a & \cellcolor{Gray} n/a & \cellcolor{Gray} n/a & \cellcolor{Gray} n/a & \cellcolor{Gray} n/a & {0.981}\\

& \textbf{S2:} \textit{+Push} & {0.942} & {0.492} & \cellcolor{Gray} n/a & \cellcolor{Gray} n/a & \cellcolor{Gray} n/a &  \cellcolor{Gray} n/a & \cellcolor{Gray} n/a & {0.468}\\

& \textbf{S3:} \textit{+Pick \& Insert} & {0.639} & {0.663} & {0.717} & \cellcolor{Gray} n/a & \cellcolor{Gray} n/a &  \cellcolor{Gray} n/a & \cellcolor{Gray} n/a & {0.570}\\

& \textbf{S4:} \textit{+Pick \& Place} & {0.010} & {0.010} & {0.012} & {0.025} & \cellcolor{Gray} n/a &  \cellcolor{Gray} n/a & \cellcolor{Gray} n/a & {0.009}\\

& \textbf{S5:} \textit{+Rotate Open} & {0.007} & {0.009} & {0.009} & {0.022} & {0.757} &  \cellcolor{Gray} n/a & \cellcolor{Gray} n/a & {0.005}\\

& \textbf{S6:} \textit{+Pick \& Drop} & {0.015} & {0.016} & {0.013} & {0.024} & {0.737} &  {0.982} & \cellcolor{Gray} n/a & {0.006}\\

& \textbf{S7:} \textit{+Pull-Open} & {0.004} & {0.006} & {0.005} & {0.021} & {0.786} &  {0.968} &  {0.864} & {0.004} \\ 
\hline

{\multirow{7}{*}{\rotatebox[origin=c]{90}{\underline{RL, BC}}}} & \textbf{S1:} \textit{Push-Button}   & {0.995} & \cellcolor{Gray} n/a & \cellcolor{Gray} n/a & \cellcolor{Gray} n/a & \cellcolor{Gray} n/a & \cellcolor{Gray} n/a & \cellcolor{Gray} n/a & {0.995}\\

& \textbf{S2:} \textit{+Push} & {0.965} & {0.912} & \cellcolor{Gray} n/a & \cellcolor{Gray} n/a & \cellcolor{Gray} n/a &  \cellcolor{Gray} n/a & \cellcolor{Gray} n/a & {0.897}\\

& \textbf{S3:} \textit{+Pick \& Insert} & {0.907} & {0.902} & {0.875} & \cellcolor{Gray} n/a & \cellcolor{Gray} n/a &  \cellcolor{Gray} n/a & \cellcolor{Gray} n/a & {0.841}\\

& \textbf{S4:} \textit{+Pick \& Place} & {0.894} & {0.703} & {0.689} & {0.772} & \cellcolor{Gray} n/a &  \cellcolor{Gray} n/a & \cellcolor{Gray} n/a & {0.683}\\

& \textbf{S5:} \textit{+Rotate Open} & {0.532} & {0.509} & {0.500} & {0.549} & {0.572} &  \cellcolor{Gray} n/a & \cellcolor{Gray} n/a & {0.492}\\

& \textbf{S6:} \textit{+Pick \& Drop} & {0.866} & {0.846} & {0.831} & {0.893} & {0.869} &  {0.956} & \cellcolor{Gray} n/a & {0.754}\\

& \textbf{S7:} \textit{+Pull-Open} & {0.911} & {0.866} & {0.848} & {0.923} & {0.892} &  {0.967} &  {0.967} & {0.774} \\ 
\hline

{\multirow{7}{*}{\rotatebox[origin=c]{90}{\underline{ROMAN's $\dagger$}}}} & \textbf{S1:} \textit{Push-Button}  & {0.973} & \cellcolor{Gray} n/a & \cellcolor{Gray} n/a & \cellcolor{Gray} n/a & \cellcolor{Gray} n/a & \cellcolor{Gray} n/a & \cellcolor{Gray} n/a & {0.973}\\

& \textbf{S2:} \textit{+Push} & {0.975} & {0.989} & \cellcolor{Gray} n/a & \cellcolor{Gray} n/a & \cellcolor{Gray} n/a &  \cellcolor{Gray} n/a & \cellcolor{Gray} n/a & {0.975}\\

& \textbf{S3:} \textit{+Pick \& Insert} & {0.841} & {0.854} & {0.869} & \cellcolor{Gray} n/a & \cellcolor{Gray} n/a &  \cellcolor{Gray} n/a & \cellcolor{Gray} n/a & {0.817}\\

& \textbf{S4:} \textit{+Pick \& Place} & {0.965} & {0.966} & {0.966} & {0.985} & \cellcolor{Gray} n/a &  \cellcolor{Gray} n/a & \cellcolor{Gray} n/a & {0.959}\\

& \textbf{S5:} \textit{+Rotate Open} & {0.882} & {0.892} & {0.892} & {0.938} & {0.949} &  \cellcolor{Gray} n/a & \cellcolor{Gray} n/a & {0.852}\\

& \textbf{S6:} \textit{+Pick \& Drop} & {0.967} & {0.973} & {0.971} & {0.983} & {0.987} &  {0.992} & \cellcolor{Gray} n/a & {0.960}\\

& \textbf{S7:} \textit{+Pull-Open} & {0.958} & {0.964} & {0.962} & {0.983} & {0.991} &  {0.993} &  {0.994} & {0.952} \\
\hline
\end{tabular}}
\end{footnotesize}
\label{table_sup:DetailedAlgorithmROMAN_0_5cm}
\end{table}

\begin{table}[H]
\caption{\textbf{Detailed results of ROMAN expanding upon Main Manuscript Table 2.c, where the overall success of each sequential case scenario is depicted with the use of different learning algorithms at $\sigma = \pm 1.0 \text{ [cm]}$ and $\sigma = \pm 2.0 \text{ [cm]}$ levels of Gaussian noise.} The table summarises the overall success rates for each sequential case scenario, in addition to the individual expert success rates. The evaluation is based on different combinations of BC, RL and GAIL. \textbf{Note on BC:} Supervised learning on the demonstration dataset. \textbf{Note on GAIL:} Use of IL intrinsic rewards ($r_{I}$) provided to PPO. \textbf{Note on RL:} Use of task extrinsic rewards ($r_{E}$) provided to PPO. \textbf{ROMAN's $\dagger$:} Default HHL approach combining BC, IL (via $r_{I}$) and RL (via $r_{E}$).}
	\label{table_sup:DetailedAlgorithmROMAN_1_and_2cm}
        \begin{subtable}[h]{1.0\textwidth}
    	\centering
    	\begin{footnotesize}
    	\setlength\tabcolsep{5pt}
            \resizebox{\textwidth}{!}{
            \begin{tabular}{ll|rrrrrrr|r} \bottomrule
            \rowcolor{DarkerGray} \multicolumn{10}{c}{{\textbf{Detailed Success Rates of Different Algorithms within ROMAN at Noise Level: $\pm 1.0$ cm}}} \\ \hline
            \rowcolor{DarkGray} &  \textbf{Scenario Cases} & \textbf{Push-Button} & \textbf{Push} & \textbf{Pick \& Insert} & \textbf{Pick \& Place} & \textbf{Rotate Open} & \textbf{Pick \& Drop} & \textbf{Pull-Open} & \textbf{Overall} \\ \hline
            
            {\multirow{7}{*}{\rotatebox[origin=c]{90}{\underline{BC}}}} & \textbf{S1:} \textit{Push-Button}  & {0.995} & \cellcolor{Gray} n/a & \cellcolor{Gray} n/a & \cellcolor{Gray} n/a & \cellcolor{Gray} n/a & \cellcolor{Gray} n/a & \cellcolor{Gray} n/a & {0.995}\\
            
            & \textbf{S2:} \textit{+Push} & {0.991} & {0.996} & \cellcolor{Gray} n/a & \cellcolor{Gray} n/a & \cellcolor{Gray} n/a &  \cellcolor{Gray} n/a & \cellcolor{Gray} n/a & {0.990}\\
            
            & \textbf{S3:} \textit{+Pick \& Insert} & {0.798} & {0.786} & {0.740} & \cellcolor{Gray} n/a & \cellcolor{Gray} n/a &  \cellcolor{Gray} n/a & \cellcolor{Gray} n/a & {0.712}\\
            
            & \textbf{S4:} \textit{+Pick \& Place} & {0.650} & {0.629} & {0.578} & {0.810} & \cellcolor{Gray} n/a &  \cellcolor{Gray} n/a & \cellcolor{Gray} n/a & {0.573}\\
            
            & \textbf{S5:} \textit{+Rotate Open} & {0.543} & {0.522} & {0.483} & {0.655} & {0.750} &  \cellcolor{Gray} n/a & \cellcolor{Gray} n/a & {0.474}\\
            
            & \textbf{S6:} \textit{+Pick \& Drop} & {0.657} & {0.606} & {0.571} & {0.723} & {0.804} &  {0.973} & \cellcolor{Gray} n/a & {0.563}\\
            
            & \textbf{S7:} \textit{+Pull-Open} & {0.748} & {0.684} & {0.637} & {0.823} & {0.921} &  {0.978} &  {0.992} & {0.632} \\ 
            \hline

            {\multirow{7}{*}{\rotatebox[origin=c]{90}{\underline{RL, BC}}}} & \textbf{S1:} \textit{Push-Button}  & {0.996} & \cellcolor{Gray} n/a & \cellcolor{Gray} n/a & \cellcolor{Gray} n/a & \cellcolor{Gray} n/a & \cellcolor{Gray} n/a & \cellcolor{Gray} n/a & {0.996}\\
            
            & \textbf{S2:} \textit{+Push} & {0.990} & {0.897} & \cellcolor{Gray} n/a & \cellcolor{Gray} n/a & \cellcolor{Gray} n/a &  \cellcolor{Gray} n/a & \cellcolor{Gray} n/a & {0.895}\\
            
            & \textbf{S3:} \textit{+Pick \& Insert} & {0.929} & {0.923} & {0.916} & \cellcolor{Gray} n/a & \cellcolor{Gray} n/a &  \cellcolor{Gray} n/a & \cellcolor{Gray} n/a & {0.881}\\
            
            & \textbf{S4:} \textit{+Pick \& Place} & {0.941} & {0.780} & {0.768} & {0.811} & \cellcolor{Gray} n/a &  \cellcolor{Gray} n/a & \cellcolor{Gray} n/a & {0.766}\\
            
            & \textbf{S5:} \textit{+Rotate Open} & {0.605} & {0.595} & {0.586} & {0.608} & {0.613} &  \cellcolor{Gray} n/a & \cellcolor{Gray} n/a & {0.562}\\
            
            & \textbf{S6:} \textit{+Pick \& Drop} & {0.839} & {0.830} & {0.816} & {0.849} & {0.774} &  {0.944} & \cellcolor{Gray} n/a & {0.696}\\
            
            & \textbf{S7:} \textit{+Pull-Open} & {0.890} & {0.868} & {0.863} & {0.895} & {0.785} &  {0.976} &  {0.980} & {0.729} \\ 
            \hline

            {\multirow{7}{*}{\rotatebox[origin=c]{90}{\underline{ROMAN's}}}} & \textbf{S1:} \textit{Push-Button}  & {0.977} & \cellcolor{Gray} n/a & \cellcolor{Gray} n/a & \cellcolor{Gray} n/a & \cellcolor{Gray} n/a & \cellcolor{Gray} n/a & \cellcolor{Gray} n/a & {0.977}\\

            & \textbf{S2:} \textit{+Push} & {0.999} & {0.997} & \cellcolor{Gray} n/a & \cellcolor{Gray} n/a & \cellcolor{Gray} n/a &  \cellcolor{Gray} n/a & \cellcolor{Gray} n/a & {0.990}\\
            
            & \textbf{S3:} \textit{+Pick \& Insert} & {0.831} & {0.841} & {0.843} & \cellcolor{Gray} n/a & \cellcolor{Gray} n/a &  \cellcolor{Gray} n/a & \cellcolor{Gray} n/a & {0.798}\\
            
            & \textbf{S4:} \textit{+Pick \& Place} & {0.955} & {0.960} & {0.955} & {0.980} & \cellcolor{Gray} n/a &  \cellcolor{Gray} n/a & \cellcolor{Gray} n/a & {0.946}\\
            
            & \textbf{S5:} \textit{+Rotate Open} & {0.796} & {0.802} & {0.799} & {0.813} & {0.816} &  \cellcolor{Gray} n/a & \cellcolor{Gray} n/a & {0.776}\\
            
            & \textbf{S6:} \textit{+Pick \& Drop} & {0.952} & {0.955} & {0.946} & {0.956} & {0.974} &  {0.984} & \cellcolor{Gray} n/a & {0.933}\\
            
            & \textbf{S7:} \textit{+Pull-Open} & {0.954} & {0.955} & {0.944} & {0.970} & {0.987} &  {0.999} &  {0.989} & {0.939} \\ 
            \hline
            \end{tabular}}
            \end{footnotesize}
            \caption{Algorithm comparison at $\pm 1.0$ cm of Gaussian noise.} 
            \label{table_sup:DetailedAlgorithmROMAN_1cm}
	\end{subtable}

	\hfill
        \vspace{0.5cm}

        \begin{subtable}[h]{1.0\textwidth}
    	\centering
    	\begin{footnotesize}
    	\setlength\tabcolsep{5pt}
            \resizebox{\textwidth}{!}{
            \begin{tabular}{ll|rrrrrrr|r} \bottomrule
            \rowcolor{DarkerGray} \multicolumn{10}{c}{{\textbf{Detailed Success Rates of Different Algorithms within ROMAN at Noise Level: $\pm 2.0$ cm}}} \\ \hline
            \rowcolor{DarkGray} &  \textbf{Scenario Cases} & \textbf{Push-Button} & \textbf{Push} & \textbf{Pick \& Insert} & \textbf{Pick \& Place} & \textbf{Rotate Open} & \textbf{Pick \& Drop} & \textbf{Pull-Open} & \textbf{Overall} \\ \hline
            
            {\multirow{7}{*}{\rotatebox[origin=c]{90}{\underline{BC}}}} & \textbf{S1:} \textit{Push-Button}  & {0.838} & \cellcolor{Gray} n/a & \cellcolor{Gray} n/a & \cellcolor{Gray} n/a & \cellcolor{Gray} n/a & \cellcolor{Gray} n/a & \cellcolor{Gray} n/a & {0.838}\\
            
            & \textbf{S2:} \textit{+Push} & {0.678} & {0.902} & \cellcolor{Gray} n/a & \cellcolor{Gray} n/a & \cellcolor{Gray} n/a &  \cellcolor{Gray} n/a & \cellcolor{Gray} n/a & {0.678}\\
            
            & \textbf{S3:} \textit{+Pick \& Insert} & {0.657} & {0.732} & {0.733} & \cellcolor{Gray} n/a & \cellcolor{Gray} n/a &  \cellcolor{Gray} n/a & \cellcolor{Gray} n/a & {0.609}\\
            
            & \textbf{S4:} \textit{+Pick \& Place} & {0.222} & {0.235} & {0.227} & {0.338} & \cellcolor{Gray} n/a &  \cellcolor{Gray} n/a & \cellcolor{Gray} n/a & {0.205}\\
            
            & \textbf{S5:} \textit{+Rotate Open} & {0.210} & {0.225} & {0.216} & {0.290} & {0.881} &  \cellcolor{Gray} n/a & \cellcolor{Gray} n/a & {0.190}\\
            
            & \textbf{S6:} \textit{+Pick \& Drop} & {0.124} & {0.128} & {0.121} & {0.163} & {0.417} &  {0.454} & \cellcolor{Gray} n/a & {0.111}\\
            
            & \textbf{S7:} \textit{+Pull-Open} & {0.085} & {0.085} & {0.081} & {0.096} & {0.215} &  {0.229} &  {0.949} & {0.075} \\ 
            \hline

            {\multirow{7}{*}{\rotatebox[origin=c]{90}{\underline{RL, BC}}}} & \textbf{S1:} \textit{Push-Button}  & {0.947} & \cellcolor{Gray} n/a & \cellcolor{Gray} n/a & \cellcolor{Gray} n/a & \cellcolor{Gray} n/a & \cellcolor{Gray} n/a & \cellcolor{Gray} n/a & {0.947}\\
            
            & \textbf{S2:} \textit{+Push} & {0.886} & {0.899} & \cellcolor{Gray} n/a & \cellcolor{Gray} n/a & \cellcolor{Gray} n/a &  \cellcolor{Gray} n/a & \cellcolor{Gray} n/a & {0.841}\\
            
            & \textbf{S3:} \textit{+Pick \& Insert} & {0.817} & {0.815} & {0.799} & \cellcolor{Gray} n/a & \cellcolor{Gray} n/a &  \cellcolor{Gray} n/a & \cellcolor{Gray} n/a & {0.725}\\
            
            & \textbf{S4:} \textit{+Pick \& Place} & {0.599} & {0.478} & {0.456} & {0.525} & \cellcolor{Gray} n/a &  \cellcolor{Gray} n/a & \cellcolor{Gray} n/a & {0.442}\\
            
            & \textbf{S5:} \textit{+Rotate Open} & {0.420} & {0.393} & {0.385} & {0.453} & {0.718} &  \cellcolor{Gray} n/a & \cellcolor{Gray} n/a & {0.363}\\
            
            & \textbf{S6:} \textit{+Pick \& Drop} & {0.298} & {0.285} & {0.282} & {0.332} & {0.472} &  {0.518} & \cellcolor{Gray} n/a & {0.246}\\
            
            & \textbf{S7:} \textit{+Pull-Open} & {0.131} & {0.121} & {0.119} & {0.147} & {0.214} &  {0.236} &  {0.943} & {0.100} \\ 
            \hline

            {\multirow{7}{*}{\rotatebox[origin=c]{90}{\underline{ROMAN's}}}} & \textbf{S1:} \textit{Push-Button}  & {0.967} & \cellcolor{Gray} n/a & \cellcolor{Gray} n/a & \cellcolor{Gray} n/a & \cellcolor{Gray} n/a & \cellcolor{Gray} n/a & \cellcolor{Gray} n/a & {0.967}\\

            & \textbf{S2:} \textit{+Push} & {0.986} & {0.994} & \cellcolor{Gray} n/a & \cellcolor{Gray} n/a & \cellcolor{Gray} n/a &  \cellcolor{Gray} n/a & \cellcolor{Gray} n/a & {0.986}\\
            
            & \textbf{S3:} \textit{+Pick \& Insert} & {0.768} & {0.783} & {0.805} & \cellcolor{Gray} n/a & \cellcolor{Gray} n/a &  \cellcolor{Gray} n/a & \cellcolor{Gray} n/a & {0.737}\\
            
            & \textbf{S4:} \textit{+Pick \& Place} & {0.862} & {0.865} & {0.859} & {0.901} & \cellcolor{Gray} n/a &  \cellcolor{Gray} n/a & \cellcolor{Gray} n/a & {0.837}\\
            
            & \textbf{S5:} \textit{+Rotate Open} & {0.785} & {0.790} & {0.790} & {0.829} & {0.892} &  \cellcolor{Gray} n/a & \cellcolor{Gray} n/a & {0.753}\\
            
            & \textbf{S6:} \textit{+Pick \& Drop} & {0.842} & {0.843} & {0.850} & {0.882} & {0.943} &  {0.954} & \cellcolor{Gray} n/a & {0.820}\\
            
            & \textbf{S7:} \textit{+Pull-Open} & {0.840} & {0.839} & {0.840} & {0.886} & {0.973} &  {0.977} &  {0.983} & {0.815} \\  
            \hline
            \end{tabular}}
            \end{footnotesize}
            \caption{Algorithm comparison at $\pm 2.0$ cm of Gaussian noise.} 
            \label{table_sup:DetailedAlgorithmROMAN_2cm}
	\end{subtable}
\end{table}

\clearpage
\begin{table}[H]
\centering
\caption{\textbf{Detailed results of ROMAN expanding upon Manuscript.Table.2d, where the overall success of each sequential case scenario is depicted over a different number of demonstrations provided to the MN, with additional details regarding the individual expert success also highlighted.} The table summarises the overall success rates of each sequential case scenario, in addition to the individual expert success rate, based on the different number of demonstrations employed to the primary gating network (the MN) for a total of $N=7$, $N=21$ and $N=42$. It is worthwhile to point out that $N=7$ correspond to one demonstration for each of the seven case scenarios. Essentially this constitutes $N=7$, $N=21$ and $N=42$ to a total of $N=1$, $N=3$ and $N=6$ demonstrations for each sequential case respectively. Trained and tested on $\sigma = \pm 0.5 \text{ [cm]}$ noise.}
\begin{footnotesize}
\setlength\tabcolsep{5pt}
        \resizebox{\textwidth}{!}{
\begin{tabular}{ll|rrrrrrr|r} \bottomrule
\rowcolor{DarkerGray} \multicolumn{10}{c}{{\textbf{Detailed Success Rates of ROMAN with Different Number of Demonstrations}}} \\ \hline
\rowcolor{DarkGray} &  \textbf{Scenario Cases} & \textbf{Push-Button} & \textbf{Push} & \textbf{Pick \& Insert} & \textbf{Pick \& Place} & \textbf{Rotate Open} & \textbf{Pick \& Drop} & \textbf{Pull-Open} & \textbf{Overall} \\ \hline

{\multirow{7}{*}{\rotatebox[origin=c]{90}{\underline{N = 7}}}} & \textbf{S1:} \textit{Push-Button}  & {0.775} & \cellcolor{Gray} n/a & \cellcolor{Gray} n/a & \cellcolor{Gray} n/a & \cellcolor{Gray} n/a & \cellcolor{Gray} n/a & \cellcolor{Gray} n/a & {0.775}\\

& \textbf{S2:} \textit{+Push} & {0.888} & {0.994} & \cellcolor{Gray} n/a & \cellcolor{Gray} n/a & \cellcolor{Gray} n/a &  \cellcolor{Gray} n/a & \cellcolor{Gray} n/a & {0.876}\\

& \textbf{S3:} \textit{+Pick \& Insert} & {0.732} & {0.775} & {0.818} & \cellcolor{Gray} n/a & \cellcolor{Gray} n/a &  \cellcolor{Gray} n/a & \cellcolor{Gray} n/a & {0.680}\\

& \textbf{S4:} \textit{+Pick \& Place} & {0.391} & {0.394} & {0.409} & {0.498} & \cellcolor{Gray} n/a &  \cellcolor{Gray} n/a & \cellcolor{Gray} n/a & {0.378}\\

& \textbf{S5:} \textit{+Rotate Open} & {0.377} & {0.386} & {0.414} & {0.515} & {0.970} &  \cellcolor{Gray} n/a & \cellcolor{Gray} n/a & {0.360}\\

& \textbf{S6:} \textit{+Pick \& Drop} & {0.008} & {0.008} & {0.008} & {0.013} & {0.031} &  {0.033} & \cellcolor{Gray} n/a & {0.008}\\

& \textbf{S7:} \textit{+Pull-Open} & {0.005} & {0.009} & {0.010} & {0.013} & {0.033} &  {0.035} &  {0.978} & {0.005} \\ 
\hline

{\multirow{7}{*}{\rotatebox[origin=c]{90}{\underline{N = 21}}}} & \textbf{S1:} \textit{Push-Button}  & {0.994} & \cellcolor{Gray} n/a & \cellcolor{Gray} n/a & \cellcolor{Gray} n/a & \cellcolor{Gray} n/a & \cellcolor{Gray} n/a & \cellcolor{Gray} n/a & {0.994}\\

& \textbf{S2:} \textit{+Push} & {0.923} & {0.934} & \cellcolor{Gray} n/a & \cellcolor{Gray} n/a & \cellcolor{Gray} n/a &  \cellcolor{Gray} n/a & \cellcolor{Gray} n/a & {0.921}\\

& \textbf{S3:} \textit{+Pick \& Insert} & {0.731} & {0.745} & {0.739} & \cellcolor{Gray} n/a & \cellcolor{Gray} n/a &  \cellcolor{Gray} n/a & \cellcolor{Gray} n/a & {0.718}\\

& \textbf{S4:} \textit{+Pick \& Place} & {0.948} & {0.958} & {0.957} & {0.986} & \cellcolor{Gray} n/a &  \cellcolor{Gray} n/a & \cellcolor{Gray} n/a & {0.945}\\

& \textbf{S5:} \textit{+Rotate Open} & {0.944} & {0.951} & {0.945} & {0.969} & {0.947} &  \cellcolor{Gray} n/a & \cellcolor{Gray} n/a & {0.902}\\

& \textbf{S6:} \textit{+Pick \& Drop} & {0.935} & {0.940} & {0.938} & {0.957} & {0.969} &  {0.974} & \cellcolor{Gray} n/a & {0.929}\\

& \textbf{S7:} \textit{+Pull-Open} & {0.962} & {0.969} & {0.969} & {0.988} & {0.992} &  {0.995} &  {0.996} & {0.958} \\ 
\hline

{\multirow{7}{*}{\rotatebox[origin=c]{90}{\underline{N = 42}}}} & \textbf{S1:} \textit{Push-Button}  & {0.973} & \cellcolor{Gray} n/a & \cellcolor{Gray} n/a & \cellcolor{Gray} n/a & \cellcolor{Gray} n/a & \cellcolor{Gray} n/a & \cellcolor{Gray} n/a & {0.973}\\

& \textbf{S2:} \textit{+Push} & {0.975} & {0.989} & \cellcolor{Gray} n/a & \cellcolor{Gray} n/a & \cellcolor{Gray} n/a &  \cellcolor{Gray} n/a & \cellcolor{Gray} n/a & {0.975}\\

& \textbf{S3:} \textit{+Pick \& Insert} & {0.841} & {0.854} & {0.869} & \cellcolor{Gray} n/a & \cellcolor{Gray} n/a &  \cellcolor{Gray} n/a & \cellcolor{Gray} n/a & {0.817}\\

& \textbf{S4:} \textit{+Pick \& Place} & {0.965} & {0.966} & {0.966} & {0.985} & \cellcolor{Gray} n/a &  \cellcolor{Gray} n/a & \cellcolor{Gray} n/a & {0.959}\\

& \textbf{S5:} \textit{+Rotate Open} & {0.882} & {0.892} & {0.892} & {0.938} & {0.949} &  \cellcolor{Gray} n/a & \cellcolor{Gray} n/a & {0.852}\\

& \textbf{S6:} \textit{+Pick \& Drop} & {0.967} & {0.973} & {0.971} & {0.983} & {0.987} &  {0.992} & \cellcolor{Gray} n/a & {0.960}\\

& \textbf{S7:} \textit{+Pull-Open} & {0.958} & {0.964} & {0.962} & {0.983} & {0.991} &  {0.993} &  {0.994} & {0.952} \\ 
\hline
\end{tabular}}
\end{footnotesize}
\label{table_sup:DetailedDemonstrationsROMAN}
\end{table}

\clearpage
\begin{table}[H]
\centering
\caption{\textbf{The duration of each expert within ROMAN's hierarchical architecture completing their individual sub-task.} This table details the average duration, in seconds, as well as the standard deviation of each expert completing their specialised sub-task goal. Results were obtained amongst 10,000 trials per cell at noise levels ranging from  $\sigma = \pm 0.0 \text{ [cm]}$ to $\sigma = \pm 2.5 \text{ [cm]}$ with $0.5 \text{ [cm]}$ increments. Symbols of $\downarrow$ and $\uparrow$ denote a decrease and increase in time differences compared to the previous noise level respectively.}
\begin{footnotesize}
\setlength\tabcolsep{5pt}
        \resizebox{\textwidth}{!}{
\begin{tabular}{l|rrrrrrr} \bottomrule
\rowcolor{DarkerGray} \multicolumn{8}{c}{{\textbf{The Duration of Individual Expert Sub-Task of ROMAN}}} \\ \hline
\rowcolor{DarkGray} \textbf{Noise Levels} & \textit{Push-Button} & \textit{Push}  & \textit{Pick \& Insert} & \textit{Pick \& Place} & \textit{Rotate Open} & \textit{Pick \& Drop} & \textit{Pull-Open} \\ \hline

$\mathbf{\sigma = \pm 0.0 \text{ [cm]}}$ & {$24.32 \pm \phantom{0}7.55s$} & {$16.25 \pm \phantom{0}9.03s$} & {$32.87 \pm 13.89s$} & {$28.53 \pm \phantom{0}6.77s$} & {$21.83 \pm \phantom{0}5.80s$} & {$22.61 \pm \phantom{0}5.05s$} & {$27.66 \pm 14.34s$} \\

$\mathbf{\sigma = \pm 0.5 \text{ [cm]}}$ & {$^{\downarrow} 23.69 \pm \phantom{0}6.89s$} & {$^{\downarrow} 15.72 \pm \phantom{0}8.38s$} & {$^{\downarrow} 32.15 \pm 12.93s$} & {$^{\uparrow} 28.61 \pm \phantom{0}6.32s$} & {$^{\downarrow} 21.48 \pm \phantom{0}4.75s$} & {$^{\uparrow} 22.67 \pm \phantom{0}5.85s$} & {$^{\downarrow} 21.18 \pm \phantom{0}5.90s$} \\

$\mathbf{\sigma = \pm 1.0 \text{ [cm]}}$ & {$^{\downarrow} 22.62 \pm \phantom{0}4.21s$} & {$^{\downarrow} 15.64 \pm \phantom{0}7.66s$} & {$^{\uparrow} 32.48 \pm 12.27s$} & {$^{\uparrow} 29.90 \pm \phantom{0}7.76s$} & {$^{\uparrow} 21.60 \pm \phantom{0}5.29s$} & {$^{\uparrow} 23.60 \pm \phantom{0}6.91s$} & {$^{\uparrow} 21.19 \pm \phantom{0}6.37s$} \\ 

$\mathbf{\sigma = \pm 1.5 \text{ [cm]}}$ & {$^{\downarrow} 22.36 \pm \phantom{0}4.05s$} & {$^{\uparrow} 15.90 \pm \phantom{0}7.41s$} & {$^{\uparrow} 34.42 \pm 14.15s$} & {$^{\uparrow} 32.42 \pm 10.29s$} & {$^{\uparrow} 21.77 \pm \phantom{0}5.35s$} & {$^{\uparrow} 23.99 \pm \phantom{0}7.51s$}  & {$^{\uparrow} 21.41 \pm \phantom{0}6.70s$} \\ 

$\mathbf{\sigma = \pm 2.0 \text{ [cm]}}$ & {$^{\uparrow} 22.49 \pm \phantom{0}4.72s$} & {$^{\uparrow} 16.85 \pm \phantom{0}8.89s$} & {$^{\uparrow} 37.16 \pm 16.10s$} & {$^{\uparrow} 36.29 \pm 13.20s$} & {$^{\uparrow} 21.79 \pm \phantom{0}5.45s$} & {$^{\uparrow} 25.70 \pm 10.08s$} & {$^{\uparrow} 22.21 \pm \phantom{0}7.41s$} \\ 

$\mathbf{\sigma = \pm 2.5 \text{ [cm]}}$ & {$^{\uparrow} 22.86 \pm \phantom{0}5.42s$} & {$^{\uparrow} 18.22 \pm 10.25s$} & {$^{\uparrow} 40.41 \pm 17.95s$} & {$^{\uparrow} 40.98 \pm 15.56s$} & {$^{\uparrow} 22.32 \pm \phantom{0}6.05s$}  & {$^{\uparrow} 27.19 \pm 11.92s$} & {$^{\uparrow} 23.47 \pm \phantom{0}9.22s$} \\

\hline
\end{tabular}}
\end{footnotesize}
\label{table_sup:Duration_Experts_of_ROMAN_on_Noise_Levels}
\end{table}

\begin{table}[H]
\centering
\caption{\textbf{The duration of the full sequential tasks by ROMAN's hierarchical architecture and all included experts within the framework.} The table details the average duration, in seconds, as well as the standard deviation of the full sequential scenario case tasks studied. The results were obtained amongst 1,000 trials per cell at noise levels ranging from  $\sigma = \pm 0.0 \text{ [cm]}$ to $\sigma = \pm 2.5 \text{ [cm]}$ at  $0.5 \text{ [cm]}$ increments. Symbols of $\downarrow$ and $\uparrow$ denote a decrease and increase in the time differences compared to the previous noise level respectively.}
\begin{footnotesize}
\setlength\tabcolsep{5pt}
        \resizebox{\textwidth}{!}{
\begin{tabular}{l|rrrrrrr} \bottomrule
\rowcolor{DarkerGray} \multicolumn{8}{c}{{\textbf{The Duration of the Full Sequential Tasks of ROMAN}}} \\ \hline

\rowcolor{DarkGray} & \textbf{Scenario 1} & \textbf{Scenario 2}  & \textbf{Scenario 3} & \textbf{Scenario 4} & \textbf{Scenario 5} & \textbf{Scenario 6} & \textbf{Scenario 7} \\ \hline

\rowcolor{DarkGray} & \textit{One Expert} & \textit{Two Expert} & \textit{Three Expert} & \textit{Four Expert} & \textit{Five Expert} & \textit{Six Expert} & \textit{Seven Expert} \\ 

\rowcolor{DarkGray} \textbf{Noise Levels} & \textit{Push-Button} & \textit{+Push}  & \textit{+Pick \& Insert} & \textit{+Pick \& Place} & \textit{+Rotate Open} & \textit{+Pick \& Drop} & \textit{+Pull-Open} \\ \hline

$\mathbf{\sigma = \pm 0.0 \text{ [cm]}}$ & {$23.44 \pm 8.78s$} & {$47.63 \pm 16.34s$} & {$93.97 \pm 36.55s$} & {$113.74 \pm 29.94s$} & {$178.18 \pm 72.36s$} & {$165.00 \pm 46.29s$} & {$210.29 \pm 64.49s$} \\

$\mathbf{\sigma = \pm 0.5 \text{ [cm]}}$ & {$^{\downarrow} 22.80 \pm 8.37s$} & {$^{\downarrow} 45.34 \pm 15.12s$} & {$^{\uparrow} 94.07 \pm 38.41s$} & {$^{\downarrow} 110.94 \pm 27.45s$} & {$^{\downarrow} 168.76 \pm 66.81s$} & {$^{\downarrow} 163.41 \pm 42.59s$} & {$^{\downarrow} 194.88 \pm 50.44s$} \\

$\mathbf{\sigma = \pm 1.0 \text{ [cm]}}$ & {$^{\downarrow} 21.76 \pm 7.09s$} & {$^{\downarrow} 43.36 \pm 12.23s$} & {$^{\uparrow} 95.89 \pm 40.36s$} & {$^{\uparrow} 111.55 \pm 31.06s$} & {$^{\downarrow} 168.20 \pm 69.28s$} & {$^{\uparrow} 168.79 \pm 50.85s$} & {$^{\downarrow} 192.29 \pm 50.12s$} \\

$\mathbf{\sigma = \pm 1.5 \text{ [cm]}}$ & {$^{\downarrow} 21.46 \pm 6.89s$} & {$^{\downarrow} 42.56 \pm 12.39s$} & {$^{\uparrow} 104.76 \pm 44.33s$} & {$^{\uparrow} 123.30 \pm 44.86s$} & {$^{\uparrow} 176.17 \pm 71.14s$} & {$^{\uparrow} 183.79 \pm 69.00s$} & {$^{\uparrow} 211.75 \pm 79.52s$} \\

$\mathbf{\sigma = \pm 2.0 \text{ [cm]}}$ & {$^{\uparrow} 21.59 \pm 7.37s$} & {$^{\downarrow} 41.75 \pm 12.64s$} & {$^{\downarrow} 103.96 \pm 44.37s$} & {$^{\uparrow} 125.12 \pm 47.27s$} & {$^{\downarrow} 171.14 \pm 69.50s$} & {$^{\uparrow} 194.38 \pm 74.57s$} & {$^{\uparrow} 220.25 \pm 86.50s$} \\

$\mathbf{\sigma = \pm 2.5 \text{ [cm]}}$ & {$^{\downarrow} 21.36 \pm 6.73s$} & {$^{\downarrow} 41.40 \pm 13.65s$} & {$^{\uparrow} 104.22 \pm 44.55s$} & {$^{\uparrow} 136.62 \pm 54.38s$} & {$^{\uparrow} 181.10 \pm 72.83s$} & {$^{\uparrow} 213.84 \pm 85.52s$} & {$^{\uparrow} 236.69 \pm 97.32s$} \\

\hline
\end{tabular}}
\end{footnotesize}
\label{table_sup:Duration_Sequences_of_ROMAN_on_Noise_Levels}
\end{table}

\clearpage
\begin{table}[H]
\centering
\caption{\textbf{The differences in the standard deviation of the rewards observed between the $N=21$ and $N=42$ demonstration datasets provided to the MN.} The table details the differences in the standard deviation expressed as a \% value for $N=21$ and $N=42$ across all case scenarios. Most notably, a significant difference in the standard deviation between $N=21$ and $N=42$ is observed with S5, relative to the other scenario cases. This indicates the higher variance in the demonstration dataset of $N=42$ compared to $N=21$ dataset and the subsequent discrepancy observed in the S5 successes between the two.}
\begin{footnotesize}
\setlength\tabcolsep{5pt}
        \resizebox{\textwidth}{!}{
\begin{tabular}{rrrrrrr} \bottomrule
\rowcolor{DarkerGray} \multicolumn{7}{c}{{\textbf{Standard Deviation Differences of Rewards for Demonstration Dataset N=21 and N=42}}} \\ \hline

\rowcolor{DarkGray} \textbf{Scenario 1} & \textbf{Scenario 2}  & \textbf{Scenario 3} & \textbf{Scenario 4} & \textbf{Scenario 5} & \textbf{Scenario 6} & \textbf{Scenario 7} \\ \hline

\rowcolor{DarkGray}  \textit{One Expert} & \textit{Two Expert} & \textit{Three Expert} & \textit{Four Expert} & \textit{Five Expert} & \textit{Six Expert} & \textit{Seven Expert} \\ 

\rowcolor{DarkGray}\textit{Push-Button} & \textit{+Push}  & \textit{+Pick \& Insert} & \textit{+Pick \& Place} & \textit{+Rotate Open} & \textit{+Pick \& Drop} & \textit{+Pull-Open} \\ \hline

 2.64\% & 7.88\% & 1.36\% & 4.99\% & 53.23\% & 1.72\% & 1.08\% \\

\hline
\end{tabular}}
\end{footnotesize}
\label{table_sup:StandardDevRewardDemos}
\end{table}

\clearpage
\begin{table}[H]
        \centering 
        \caption{\textbf{The architectural details of ROMAN's hierarchical architecture in its preliminary 2D setting composed of five experts.} The table summarises the states and actions of each individual expert and the MN's of the ROMAN framework, including their individual dimensions.}
        \begin{small}
        \setlength\tabcolsep{3pt}
        \resizebox{\textwidth}{!}{\begin{tabular}{llllll} \bottomrule
        \rowcolor{DarkerGray} \multicolumn{6}{c}{{\textbf{Preliminary ROMAN stage in 2D Consisting of Five Experts}}} \\ \hline
        
        \rowcolor{Gray} \multicolumn{6}{c}{{\textbf{Network Architecture, Characteristics and Demonstration Settings}}} \\ \hline
        
        \multirow{2}{*}{\textbf{Master}} & \multicolumn{5}{c}{{\textbf{Experts NNs}}} \\ \cmidrule(lr){2-6}
        &{Push} & {Lift} & {Pick \& Place} & {Pick \& Drop} & {Pull} \\ 
        
        \hline
        \rowcolor{Gray} \multicolumn{6}{c}{{\textbf{State Space (Vector Size)}}} \\ \hline
        \scriptsize \textbf{\textit{Total: 29}} & \scriptsize \textbf{\textit{Total: 11}} & \scriptsize \textbf{\textit{Total: 9}} & \scriptsize \textbf{\textit{Total: 11}}  & \scriptsize \textbf{\textit{Total: 11}}  & \scriptsize \textbf{\textit{Total: 9}}\\ \hdashline
        
        \scriptsize Agent Position (2) & \scriptsize Agent Position (2) & \scriptsize Agent Position (2) & \scriptsize Agent Position (2) & \scriptsize Agent Position (2) & \scriptsize Agent Position (2) \\
        
        \scriptsize Agent Velocity (2) & \scriptsize Agent Velocity (2) & \scriptsize Agent Velocity (2) & \scriptsize Agent Velocity (2) & \scriptsize Agent Velocity (2) & \scriptsize Agent Velocity (2) \\
        
        \scriptsize Gripper Force (2) & \scriptsize Gripper Force (2) & \scriptsize Gripper Force (2) & \scriptsize Gripper Force (2) & \scriptsize Gripper Force (2) & \scriptsize Gripper Force (2)\\  

        \scriptsize Gripper State (1) & \scriptsize Gripper State (1) & \scriptsize Gripper State (1) & \scriptsize Gripper State (1) & \scriptsize Gripper State (1) & \scriptsize Gripper State (1)\\ 

        \scriptsize $\dagger$ Full Environment (14) & \scriptsize Rack Position (2) & \scriptsize Gate Position (2) & \scriptsize Rack Position (2) & \scriptsize Box Position (2) & \scriptsize Drawer Position (2)\\   
        
         & \scriptsize Rack Target Location (2) & & \scriptsize Vial Position (2) & \scriptsize Unbox Target Location (2) & \\   
        
        \hline \rowcolor{Gray} \multicolumn{6}{c}{{\textbf{Action Space (Vector Size)}}} \\ \hline
        \scriptsize \textbf{\textit{Total: 5}} & \scriptsize \textbf{\textit{Total: 3}} & \scriptsize \textbf{\textit{Total: 3}} & \scriptsize \textbf{\textit{Total: 3}}  & \scriptsize \textbf{\textit{Total: 3}}  & \scriptsize \textbf{\textit{Total: 3}}\\ \hdashline
        
        \scriptsize Agent Weights (5) & \scriptsize Agent Velocity (2) & \scriptsize Agent Velocity (2) & \scriptsize Agent Velocity (2) & \scriptsize Agent Velocity (2) & \scriptsize Agent Velocity (2)\\
        
        & \scriptsize Gripper State (1) & \scriptsize Gripper State (1) & \scriptsize Gripper State (1) & \scriptsize Gripper State (1) & \scriptsize Gripper State (1)\\     

        \hline
        \multicolumn{6}{c}{\scriptsize $\dagger$ Full Environment (14): Combined proprioceptive state space of all experts. } \\

        \bottomrule        
        \end{tabular}}
        \end{small}
        \label{table_sup:ROMAN_2D_DetailedNetworkCharacteristicsStructure}
\end{table}

\begin{table}[H]
\centering
\caption{\textbf{The neural network architectures of all incorporated NNs in the hierarchical architecture of ROMAN.} The table illustrates the NN architectures of the incorporated expert NNs and the MN's in the ROMAN framework. The NNs are referred to in this context as generators while also detailing the architecture of the respective discriminator by GAIL. The Multi-Layer Perceptron (MLP) for each are detailed in the table. As per the detailed expansion of the hybrid learning procedure in the main manuscript and in particular the modification to GAIL's discriminator to only differentiate between states ($s_{t}$) but not the actions ($a_{t}$) of the expert and generator's trajectories, the number of inputs for the generator and discriminator are similar. \textbf{Notation $\mathbf{\dagger}$:} Expert dependent, for more information consult the main manuscript and in particular Table 1 for the specific size of the state space for each expert NN. In particular, the state space of the expert NNs is decided on their individual sub-task goal, allowing these NNs, in turn, to solely focus on their respective end goal and omit non-relevant exteroceptive states. Due to the diverse nature of these expert NNs studied and concerned with different types of manipulation skills, this naturally drove the need for different training times for each individual expert. This was primarily dependent upon the higher-level task complexity the experts are concerned with. \textbf{Notation $\mathbf{\ddagger}$:}: The MN observed the combined state space of the incorporated experts in the hierarchical architecture. This ultimately allowed the MN to oversee the full environment state as to correctly infer the necessary orchestration of the incorporated experts as to achieve the long-horizon sequential end goal.}
\begin{tabular}{ll|ll}
\hline

\rowcolor{DarkerGray} \multicolumn{4}{c}{\underline{\textbf{ROMAN's Neural Network Architectures}}} \\
\rowcolor{DarkGray} \multicolumn{2}{c}{\textbf{Expert Networks}}
& \multicolumn{2}{c}{\textbf{Manipulation Network}}\\
\hline
\rowcolor{Gray} \multicolumn{2}{c}{{\textbf{State Space (Vector Size)}}} & \multicolumn{2}{c}{{\textbf{State Space (Vector Size)}}}\\
\hline
Agent Position & (3) & Agent Position & (3)  \\
Agent Velocity & (3) & Agent Velocity & (3)  \\
Gripper Force & (2) & Gripper Force & (2)  \\
Environment State & (3-6)$\dagger$ & Full Environment State & (21)$\ddagger$ \\
\hline

\rowcolor{Gray} \multicolumn{2}{c}{{\textbf{Action Space (Vector Size)}}} & \multicolumn{2}{c}{{\textbf{Action Space (Vector Size)}}}\\
\hline
Agent Velocity & (3) & Agent Weights & (7)  \\
Gripper State & (1) &   \\

\hline

\rowcolor{Gray} \multicolumn{2}{c}{\textbf{Demonstration Settings}} &
\multicolumn{2}{c}{\textbf{Demonstration Settings}} \\
\hline

Number of Demos & N = 20 & Number of Demos & N = 42 \\
Demo Time & $t \approx 6 \text{-} 12 \text{min}$$\dagger$ & Demo Time & $t \approx 42 \text{min}$ \\
\hline

\rowcolor{Gray} \multicolumn{2}{c}{\textbf{Generator (Expert Networks)}} &
\multicolumn{2}{c}{\textbf{Generator (Manipulation Network)}} \\
\hline
Number of Inputs & 11 to 14$\dagger$ & Number of Inputs & 29 \\ 
Number of Outputs & 4 & Number of Outputs & 7 \\ 
Number of Hidden Layers & 3  & Number of Hidden Layers & 3  \\
Hidden Units Per Layer & 128 - 256$\dagger$ & Hidden Units Per Layer & 256 \\\hline

\rowcolor{Gray} \multicolumn{2}{c}{\textbf{Discriminator (GAIL)}} & \multicolumn{2}{c}{\textbf{Discriminator (GAIL)}} \\
\hline
Number of Inputs &  11 to 14$\dagger$ & Number of Inputs & 29 \\ 
Number of Outputs & 1 & Number of Outputs & 1 \\ 
Number of Hidden Layers & 2  & Number of Hidden Layers & 2 \\
Hidden Units Per Layer & 128 & Hidden Units Per Layer & 128 \\ \hline
\end{tabular}
\label{table_sup:roman_nn_architectures}
\end{table}

\begin{table}[H]
\centering
\caption{\textbf{The hyperparameters for all the incorporated expert NNs in the ROMAN framework, including the gating network.} This table details the hyperparameter values used for all experts and that of the gating network. Additionally, the neural network structures (layers) are detailed for each. All hyperparameters were identical for all experts in the hierarchical architecture. As described in more detail in the manuscript, the hybrid learning procedure consists of initially using BC to warm-start the policy prior to using PPO and thereafter employing a GAIL reward for the PPO policy, in the form of an intrinsic reward $r_{I} = -log(1-D(s_t))$. For more details consult the main manuscript.}
\begin{tabular}{ll|ll}
\hline
\rowcolor{DarkGray} \multicolumn{4}{c}{\underline{\textbf{ROMAN's Hyperparameters Settings (PPO)}}} \\
\rowcolor{Gray} \multicolumn{2}{c}{\textbf{Expert Networks}}
& \multicolumn{2}{c}{\textbf{Manipulation Network}} \\
\hline
Minibatch Range                   & 1024 & Minibatch Range                   & 1024                            \\
GAE Parameter Lambda Range        & 0.95  & GAE Parameter Lambda Range        & 0.95                            \\
Entropy Coefficient Range         & 5.0e-3 & Entropy Coefficient Range         & 5.0e-3                           \\
Horizon Range                     & 1000 & Horizon Range                     & 1000                            \\
Number of Epochs                  & 3 & Number of Epochs                  & 3                               \\
Clipping Parameter Epsilon        & 0.2 & Clipping Parameter Epsilon        & 0.2                             \\
Discount Factor Gamma Range       & 0.99 & Discount Factor Gamma Range       & 0.99                            \\
Learning Rate                     & 3.0e-4 & Learning Rate                     & 3.0e-4                          \\
Replay Buffer Observation Size    & 10240  & Replay Buffer Observation Size    & 10240                           \\ \hline

\rowcolor{DarkGray} \multicolumn{4}{c}{\underline{\textbf{ROMAN's Hyperparameters Settings (GAIL - Discriminator)}}} \\

\rowcolor{Gray} \multicolumn{2}{c}{\textbf{Expert Networks}}
& \multicolumn{2}{c}{\textbf{Manipulation Network}} \\ \hline

Discount Factor Gamma Range       & 0.99 & Discount Factor Gamma Range       & 0.99                            \\
Learning Rate                     & 3.0e-4 & Learning Rate                     & 3.0e-4    \\
\hline

\rowcolor{DarkGray} \multicolumn{4}{c}{\underline{\textbf{ROMAN's Hyperparameters Settings (Behavioural Cloning)}}} \\

\rowcolor{Gray} \multicolumn{2}{c}{\textbf{Expert Networks}}
& \multicolumn{2}{c}{\textbf{Manipulation Network}} \\ \hline
 
Batch Size                    & 1024 & Batch Size                     & 1024    \\
Learning Rate                     & 3.0e-4 & Learning Rate                     & 3.0e-4    \\
\hline
\end{tabular}
\label{table_sup:Hyperparameters}
\end{table}

\begin{table}[H]
\centering
\caption{\textbf{The monolithic NN architecture settings for both the 2D and 3D cases.} The table details the architectural settings for the monolithic NNs for both 2D and 3D cases used as part of the baseline evaluation against the hierarchical architecture of ROMAN in its preliminary and final stages respectively. The state and action spaces are detailed, including the demonstrations provided to each. It is worthwhile to point out that the state space of the single NNs was of identical size to that of the MN of ROMAN's respective 2D and 3D cases. In contrast to ROMAN's hierarchical architecture whereby the MN controls the weights of incorporated expert NN, the action space of the monolithic NNs directly controls the velocity of the end-effector as well as the opening and closing of the gripper. The hyperparameter values employed for the 2D and 3D monolithic NN were identical to that of the MN for the 2D and 3D cases respectively to retain consistency. The same hybrid learning approach was used by ROMAN's experts and MN as well as for the monolithic NNs to further retain consistency.}
\setlength\tabcolsep{12pt}
\begin{small}
\begin{tabular}{ll|ll}
\hline
\rowcolor{DarkerGray} \multicolumn{4}{c}{\underline{\textbf{Monolithic NN Architectures}}} \\
\rowcolor{DarkGray} \multicolumn{2}{c}{\textbf{Single NN in 2D}}
& \multicolumn{2}{c}{\textbf{Single NN in 3D}}\\
\hline
\rowcolor{Gray} \multicolumn{2}{c}{{\textbf{State Space (Vector Size)}} $\dagger$} & \multicolumn{2}{c}{{\textbf{State Space (Vector Size)}} $\dagger$}\\
\hline
Agent Position & (2) & Agent Position & (3)  \\
Agent Velocity & (2) & Agent Velocity & (3)  \\
Gripper Force & (2) & Gripper Force & (2)  \\
Gripper State & (1) & Button Position & (3) \\
Rack Position & (2) & Rack Position & (3)  \\ 
Gate Position & (2) & Conveyor Position & (3)  \\ 
Vial Position & (2) & Vial Position & (3) \\ 
Box Position & (2) & Box Position & (3)  \\ 
Unbox Target Location & (2) & Drawer Position & (3) \\ 
Drawer Position & (2) & Cabinet Position & (3) \\ 
Rack Target Location & (2) & \\
\hline

\rowcolor{Gray} \multicolumn{2}{c}{{\textbf{Action Space (Vector Size)}} $\ddagger$} & \multicolumn{2}{c}{{\textbf{Action Space (Vector Size)}} $\ddagger$}\\
\hline
Agent Velocity & (2) & Agent Velocity & (3)  \\
Gripper State & (1) & Gripper State & (1)  \\

\hline

\rowcolor{Gray} \multicolumn{2}{c}{\textbf{Demonstration Settings}} &
\multicolumn{2}{c}{\textbf{Demonstration Settings}} \\
\hline

Number of Demos & N = 100 & Number of Demos & N = 140 \\
Demo Time & $t \approx 64 \text{min}$ & Demo Time & $t \approx 132 \text{min}$ \\
\hline

\rowcolor{Gray} \multicolumn{2}{c}{\textbf{Generator (Single NN in 2D)}} &
\multicolumn{2}{c}{\textbf{Generator (Single NN in 3D)}} \\
\hline
Number of Inputs & 21 & Number of Inputs & 29 \\ 
Number of Outputs & 3 & Number of Outputs & 4 \\ 
Number of Hidden Layers & 3  & Number of Hidden Layers & 3  \\
Hidden Units Per Layer & 128 & Hidden Units Per Layer & 256 \\ \hline

\rowcolor{Gray} \multicolumn{2}{c}{\textbf{Discriminator (GAIL)}} & \multicolumn{2}{c}{\textbf{Discriminator (GAIL)}} \\
\hline
Number of Inputs &  21 & Number of Inputs & 29 \\ 
Number of Outputs & 1 & Number of Outputs & 1 \\ 
Number of Hidden Layers & 2  & Number of Hidden Layers & 2 \\
Hidden Units Per Layer & 64 & Hidden Units Per Layer & 128 \\ \hline

\multicolumn{4}{c}{$\dagger$ Identical to ROMAN's MN state space for 2D/3D accordingly.} \\ 
\multicolumn{4}{c}{$\ddagger$ Identical to ROMAN's experts action space for 2D/3D accordingly.} \\ \hline
\end{tabular}
\end{small}
\label{table_sup:SingleNN_2D_3D}
\end{table}

\begin{table}[H]
\centering
\caption{\textbf{The hyperparameters for the monolithic NNs for both the 2D and 3D cases.} The table details the hyperparameter values for the monolithic NNs used as a baseline evaluation against ROMAN. As detailed in the main manuscript, an identical hybrid learning procedure was employed for the monolithic NNs as with ROMAN's hierarchical formation. To further retain consistency and conduct subsequent fair comparisons during the experimental evaluations, identical hyperparameters with ROMAN's MN were used for the corresponding 2D and 3D cases.}
\begin{tabular}{ll|ll}
\hline
\rowcolor{DarkGray} \multicolumn{4}{c}{\underline{\textbf{Monolithic NN Hyperparameters Settings (PPO)}}} \\
\rowcolor{Gray} \multicolumn{2}{c}{\textbf{Single NN (2D)}}
& \multicolumn{2}{c}{\textbf{Single NN (3D)}}\\
\hline
Minibatch Range                   & 1024 & Minibatch Range                   & 1024                            \\
GAE Parameter Lambda Range        & 0.95  & GAE Parameter Lambda Range        & 0.95                            \\
Entropy Coefficient Range         & 5.0e-3 & Entropy Coefficient Range         & 5.0e-3                           \\
Horizon Range                     & 1000 & Horizon Range                     & 1000                            \\
Number of Epochs                  & 3 & Number of Epochs                  & 3                               \\
Clipping Parameter Epsilon        & 0.2 & Clipping Parameter Epsilon        & 0.2                             \\
Discount Factor Gamma Range       & 0.99 & Discount Factor Gamma Range       & 0.99                            \\
Learning Rate                     & 3.0e-4 & Learning Rate                     & 3.0e-4                          \\
Replay Buffer Observation Size    & 10240  & Replay Buffer Observation Size    & 10240                           \\
\hline

\rowcolor{DarkGray} \multicolumn{4}{c}{\underline{\textbf{Monolithic NN Hyperparameters Settings (GAIL - Discriminator)}}} \\

\rowcolor{Gray} \multicolumn{2}{c}{\textbf{Single NN (2D)}}
& \multicolumn{2}{c}{\textbf{Single NN (3D)}}\\ \hline

Discount Factor Gamma Range       & 0.99 & Discount Factor Gamma Range       & 0.99                            \\
Learning Rate                     & 3.0e-4 & Learning Rate                     & 3.0e-4    \\
\hline

\rowcolor{DarkGray} \multicolumn{4}{c}{\underline{\textbf{Monolithic NN Hyperparameters Settings (Behavioural Cloning)}}} \\

\rowcolor{Gray} \multicolumn{2}{c}{\textbf{Single NN (2D)}}
& \multicolumn{2}{c}{\textbf{Single NN (3D)}}\\ \hline
 
Batch Size                    & 512 & Batch Size                     & 1024    \\
Learning Rate                     & 3.0e-4 & Learning Rate                     & 3.0e-4    \\
\hline
\end{tabular}
\label{table_sup:SingleNN_2D_3D_Hyperparameters}
\end{table}

\end{document}